\pdfoutput=1

\documentclass[11pt]{article}

\usepackage[preprint]{acl}

\usepackage{times}
\usepackage{latexsym}

\usepackage[T1]{fontenc}

\usepackage[utf8]{inputenc}

\usepackage{microtype}

\usepackage{inconsolata}

\usepackage{graphicx}

\usepackage{multicol}
\usepackage{multirow}
\usepackage{booktabs}
\usepackage{arydshln}
\usepackage{subcaption}
\usepackage{todonotes}

\usepackage{graphicx}
\usepackage{booktabs}

\usepackage{tikz}
\usepackage{comment}
\usepackage{amsmath,amssymb} 
\usepackage{color}
\usepackage{enumitem}
\usepackage{booktabs}
\usepackage{multirow}
\usepackage{bbm}
\usepackage{listings}
\usepackage{colortbl}
\usepackage{indentfirst}
\usepackage{soul}

\usepackage{wrapfig}

\title{SoupLM: Model Integration in Large Language
and Multi-Modal Models}

\author{
 \textbf{Yue Bai \textsuperscript{1,2}},
 \textbf{Zichen Zhang \textsuperscript{2}},
 \textbf{Jiasen Lu \textsuperscript{2}},
 \textbf{Yun Fu \textsuperscript{1}}
\\
\\
 \textsuperscript{1}Northeastern University,
 \textsuperscript{2}Allen Institute for AI
}

\begin{document}
\maketitle
\begin{abstract}
Training large language models (LLMs) and multimodal LLMs necessitates significant computing resources, and existing publicly available LLMs are typically pre-trained on diverse, privately curated datasets spanning various tasks.
For instance, LLaMA, Vicuna, and LLaVA are three LLM variants trained with LLaMA base models using very different training recipes, tasks, and data modalities.
The training cost and complexity for such LLM variants grow rapidly.
In this study, we propose to use a soup strategy to assemble these LLM variants into a single well-generalized multimodal LLM (SoupLM) in a cost-efficient manner.
Assembling these LLM variants efficiently brings knowledge and specialities trained from different domains and data modalities into an integrated one (e.g., chatbot speciality from user-shared conversations for Vicuna, and visual capacity from vision-language data for LLaVA), therefore, to avoid computing costs of repetitive training on several different domains.
We propose series of soup strategies to systematically benchmark  performance gains across various configurations, and probe the soup behavior across base models in the interpolation space.

\end{abstract}

\section{Introduction}
Training large language models (LLMs)~\cite{brown2020language,achiam2023gpt,devlin2018bert} presents several significant challenges, such as how to deploy immense size models on infrastructures and make large-scale optimization~\cite{xie2024doremi,narayanan2021efficient}, and how to collect and prepare massive training data to match the model size~\cite{swayamdipta2020dataset,wang2022self}.
As a result, the computational cost and other efforts of training such networks is rapidly growing.
For example, training a model like LLaMA3-7B~\cite{touvron2023llama} requires an extensive amount of computation with carefully defined data and training recipe, not to mention a 70B model demands even more resources and training complexity, measured in thousands of H100 hours~\cite{choquette2023nvidia}.
Constraints caused by these substantial computational costs mean that research into new large language models is often restricted to a limited number of teams with extensive resources, which may hinder the community development.

Moreover, while extending the model capacities for multiple domains by transitioning LLMs into large multi-modal models (LMMs), additional challenges arise~\cite{liu2024visual,zhu2023minigpt,yan2021videogpt}.
Training LMMs typically follows the post-training approach, which involves finetuning the base model with a multi-modal instructional tuning dataset~\cite{liu2024improved,li2024llava}.
For example, LLaVA~\cite{liu2024visual} enable its base Vicuna~\cite{zheng2023judging} model to understand visual input by finetuning it with vision-language instruction data.
In addition, extending the model with new architecture, such as branch mixing and training~\cite{sukhbaatar2024branch} under Mixture-of-Experts (MoE) design~\cite{shazeer2017outrageously}, further complicates the process.
Overall, as models become more unified and integrate diverse modalities, they face new issues like data and modality drift.
Such issues require even more complicated data and optimization recipes, which are more complex than traditional challenges and further increase the multi-modal training costs.

In this context, the concept of model soup emerges as an effective strategy to merge the base model and its finetuned variants.
It initially focuses on image classification task~\cite{wortsman2022model}.
Instead of picking the model with highest validation accuracy, model soup combines tuned models of different hyperparameter configurations, where all variants are trained from the same random initialized model that seen as the base model.
The soup strategy obtains a robust model with the highest performance, which can be generalized to several visual backbones like CLIP~\cite{radford2021learning} and ViT~\cite{dosovitskiy2020image}.
Unlike typical ensemble, the model soup directly merges weights of model variants, resulting in no additional inference and memory costs.

Motivated by the challenges above with model soup inspiration, in this work, we systematically study how to merge the model variants of different domains in the context of the large language model.
More specifically, we focus on language (LLMs) and vision-language (LMMs) domains upon the autoregressive architecture~\cite{radford2019language}.
We take Vicuna, and its variant LLaVA as two base models for a study case to explore the model integration in LLMs and LMMs, namely, SoupLM.
We propose series of soup strategies from naive weight average into finegrained learnable soup, and find SoupLM improves both language and multi-modal task performances as an integrated well-generalized model.
Such process has no additional inference cost and requires almost ignorable extra training cost, where naive soup has no training cost and learnable soup has tiny effort to adjust the soup weight.
We systematically benchmark extensive evaluations across different soup configurations to fully explore its improvement potential, statistically providing intuitions to find a better soup setting.

We are also curious about the finegrained soup behavior across base models.
For example, if the base models are given, what is the learned $\alpha$ distributions under different tuning conditions?
Correspondingly, we make detailed analysis upon different settings and further use a simple regularized soup strategy, to initially probe the soup dynamics.
To summarize our effort of this paper:
\begin{itemize}
    \item We propose SoupLM to first investigate the model soup strategy in the context of the autoregressive architecture.
    SoupLM integrates base models of different domains as a well-generalized multi-modal model, introducing ignorable training and no inference cost.
    
    \item We systematically benchmark the learnable soup strategy across various configurations to test the potential performance gain.
    It observes statistical patterns under the hyperparameter space, and inspires a principle design to derive better soup settings.
    
    \item Finegrained soup behaviors are initially probed under learnable and regularized soup, and we find the interpolation distributions are stable under training constraints and certain finetuning supervisions.
    It is expected to inspire more mechanism studies to probe its soup behaviors in an interpretable way.
\end{itemize}

\section{Method}\label{sec:method}

This section introduces \textit{vanilla}, \textit{learnable}, and \textit{regularized} soup strategies for our SoupLM exploration, where \textit{vanilla} initially explores the effectiveness of soup, \textit{learnable} serves as our central method and \textit{regularized} mainly for soup behavior analysis to validate our hypothesis.
Given a set of base models with isomorphic model structures $M=\{f(\theta^1), f(\theta^2), ..., f(\theta^n)\}$, where $n$ is the number of base models.
Here, the model $f(\cdot)$ generally represents network module at different granularities (e.g., each weight, each MLP block, and the whole model), which varies according to different soup strategies.
We keep the model structure ${f(\cdot)}$ fixed and merge $\theta^{*}$ to obtain a souped model $f(\theta^s)$.
The merging also keeps the weight $\theta^{*}$ fixed and only assign a bunch of $\alpha$ to bridge base models.
Then, the integrated one is given by

\begin{equation}
    f(\theta^s) = \sum_{i=1}^n \alpha^i \theta^i,
\end{equation}
where $\alpha$ is the critical factor of our study and explored by following soup strategies.
In this study, we specifically consider two autoregressive Transformer~\cite{vaswani2017attention} base models, Vicuna and LLaVA, for the following soup strategies and the number of base models can be easily enlarged.
And we ensure $\sum_{i=1}^n \alpha^i = 1$ to interpolate weight in linear model space.

\subsection{Vanilla Soup}\label{sec:method_vanilla_soup}

We use vanilla soup as a simple baseline to initially explore if directly combining weights of two base models improves the performance.
Herein, $f(\cdot)$ represents the whole model, which is the largest granularity.
We manually set different ratios $\alpha^1$ (e.g., 0.5) for the first base model and use $\alpha^2 = 1 - \alpha^1$ for the second.
The vanilla souped model is given by
\begin{equation}\label{eq:vanilla_soup}
    f(\theta^s) = \alpha^1 \theta^1 + (1-\alpha^1) \theta^2,
\end{equation}
where we use $\alpha^1=\{0.1, 0.2,... ,0.9\}$ in our experiments (see Sec.~\ref{sec:vanilla_soup})

\begin{table*}[htbp]
  \centering
  \caption{Summary of five meta sets from language and vision-language domains.
  }\label{tab:meta_set}
  \resizebox{0.83\textwidth}{!}{
    \begin{tabular}{cccccc}
      \toprule
      Meta Set & MMMU & LLaVA665K & MMLU & GSM8k & Hellaswag \\
      \midrule
      Number of validation& 150 & 665K & 99.8K & 7.47K & 39.9K \\
      Number of test& 900 & 60 (LLaVA-Bench) & 14K & 1.32K & 10K \\
      \bottomrule
    \end{tabular}
    }
\end{table*}

\subsection{Learnable Soup}\label{sec:method_learnable_soup}
Instead of merging base models using model-level granularity as vanilla soup, we propose to refine the process by decreasing the soup granularity to bridge base models in a fine-grained way, which is the central method in this paper.
Concretely, we choose each module in Transformer block as a smaller soup unit $f(\cdot)$, such as the \textit{Q, K, V, O} mappings in attention block and \textit{up, down} mappings in MLP block.
In addition, we also include all normalization layers, the very first embedding layer, and the last LM head mapping as units for soup.
Basically, this process can be seen as a finegrained soup at \textit{per-mapping} granularity.

Rather than manually assignment, we propose to optimize the finegrained $\alpha$ using a tiny development set $D$.
The optimization follows the typical finetuning protocol of autoregressive model to minimize the next token prediction loss, but only tuning the $\alpha_{[*,*]}$ while fixing both base models ($\theta^1_{[*,*]}$, $\theta^2_{[*,*]}$).
It integrates the weights in the model space spanned by two base models, which is formally given by:
\begin{equation}\label{eq:learnable_soup}
    \alpha_{[s,l]} = \arg \min_{\alpha} \mathcal{L}(\alpha_{[s,l]}; \theta^1_{[s,l]}, \theta^2_{[s,l]}, \mathcal{D}),
\end{equation}
where $s$ represents different soup units (e.g., Q/up project in attention/MLP) and $l$ means different Transformer layer indices.
$\mathcal{L}(\cdot;\cdot)$ is the autoregressive loss.
It elaborates the merging process by delicately tuning the soup weights following the data supervision to better take advantages of both base models.
Such refinement with smaller soup granularity firstly leads to a more flexible model interpolation space to benefit further performance gain.
Furthermore, it provides an access to investigate the functional mechanism of each soup unit by analyzing their merging behaviors.
Please note that the learnable soup can be further elaborated by reducing the soup granularity such as neuron or other self-defined units and we keep the per-mapping soup units for this study.

\subsection{Regularized Soup}\label{sec:method_regularized_soup}
Learnable soup picks smaller granularity and merges base models by fixing the original ones.
It also provides an intuitive way to investigate the model merging behaviors in the model space.
To do so, we involve a regularization term to elaborate the soup process and point out the merging behavior for analysis.
We use L1 normalization on the soup $\alpha$ and augment Eq.~\ref{eq:learnable_soup} as

\begin{equation}\label{eq:reg_train}
    \mathcal{L}_{reg}(\alpha) = \mathcal{L}(\alpha; \theta^1, \theta^2, D) + \lambda\|\alpha\|_1,
\end{equation}
where we omit the subscript of $[s,l]$ for $\alpha$.
$\lambda$ is the regularization strength parameter and $\mathcal{L}_{reg}$ is the final regularized training objective.
Other regularization formats (e.g., L2) can be easily extended and we simply consider L1 here.
Through adding regularization on the $\alpha$, its optimized values are constrainted close to its initializations.
In this way, we set increasing regularization magnitudes to observe the changes of soup distribution, and validate the hypothesis that model soup performs stable behavior according to the given base models.
Different from learnable soup above aiming to exhaust the soup potential, regularized soup is mainly to provide further intuitions of model soup behavior among base models during finetuning.

\begin{figure*}[htbp]
  \centering
  \begin{subfigure}[b]{0.19\textwidth}
    \includegraphics[width=\textwidth]{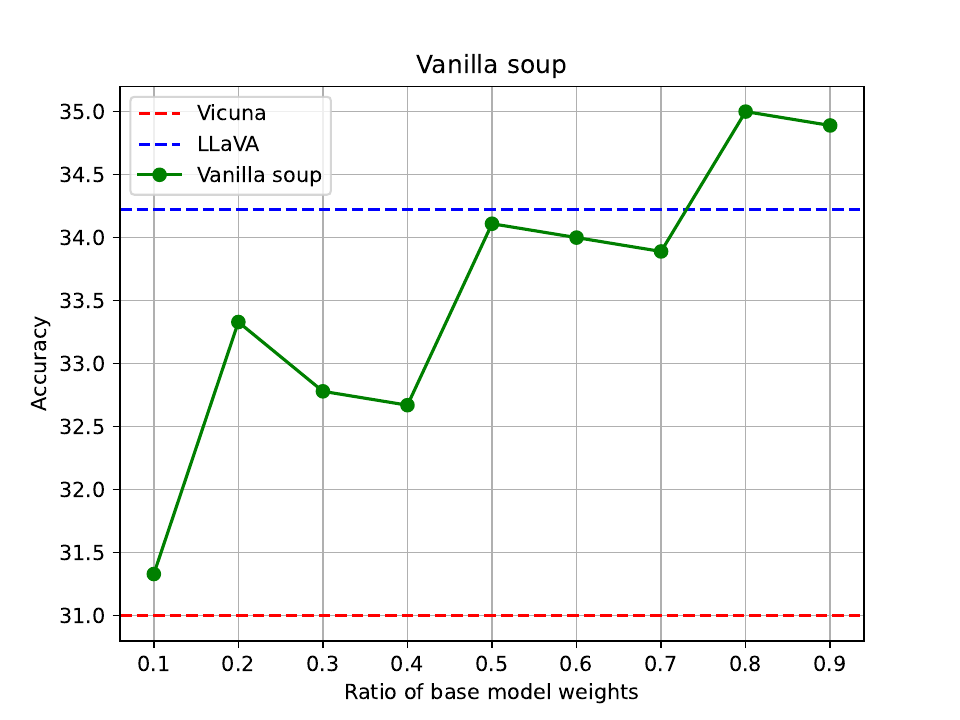}
    \caption{MMMU}
  \end{subfigure}
  \hfill 
  \begin{subfigure}[b]{0.19\textwidth}
    \includegraphics[width=\textwidth]{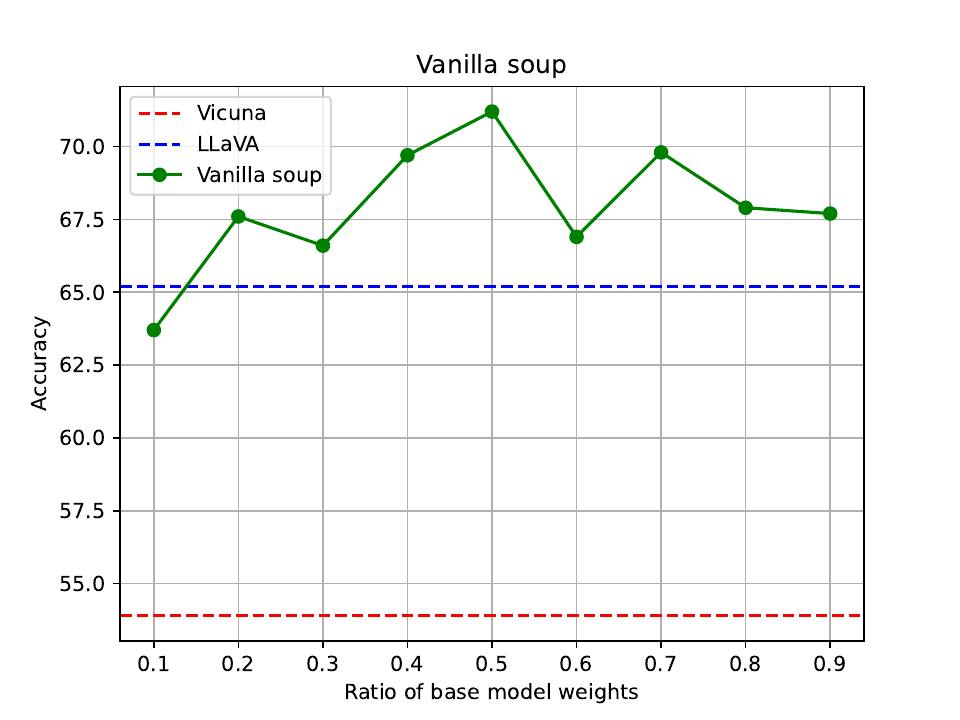}
    \caption{LLaVA-Bench}
  \end{subfigure}
  \hfill 
  \begin{subfigure}[b]{0.19\textwidth}
    \includegraphics[width=\textwidth]{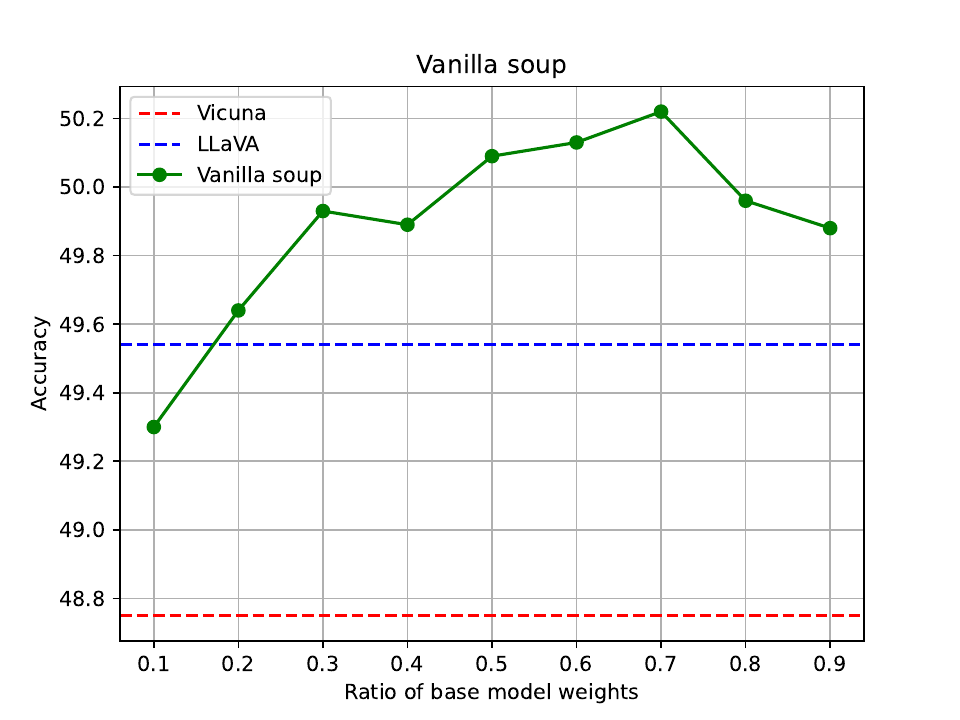}
    \caption{MMLU}
  \end{subfigure}
  \hfill 
  \begin{subfigure}[b]{0.19\textwidth}
    \includegraphics[width=\textwidth]{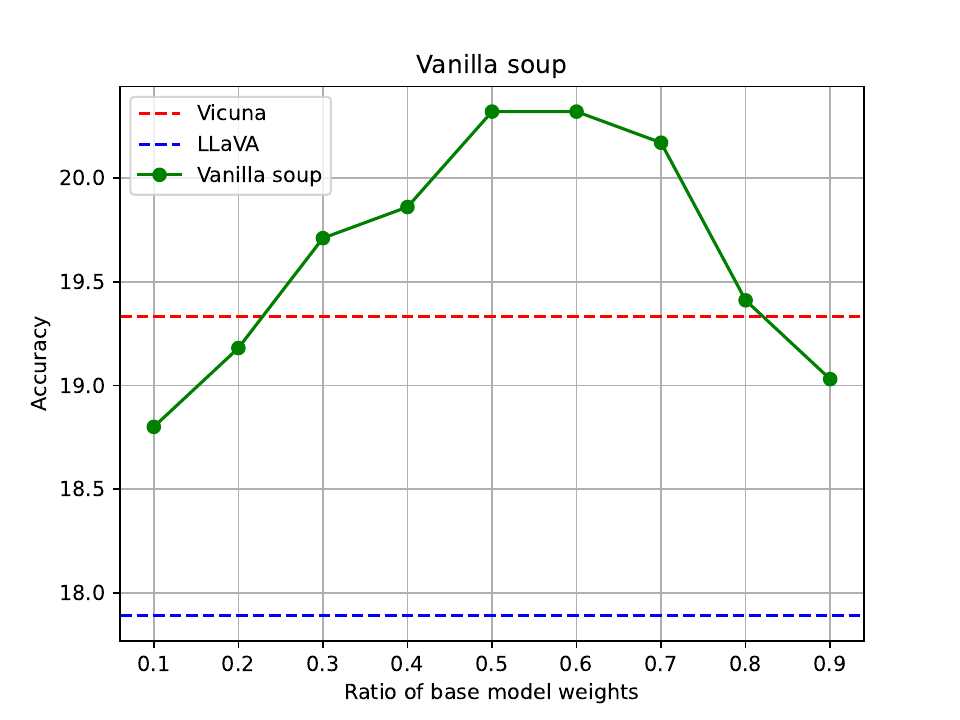}
    \caption{GSM8K}
  \end{subfigure}
  \hfill 
  \begin{subfigure}[b]{0.19\textwidth}
    \includegraphics[width=\textwidth]{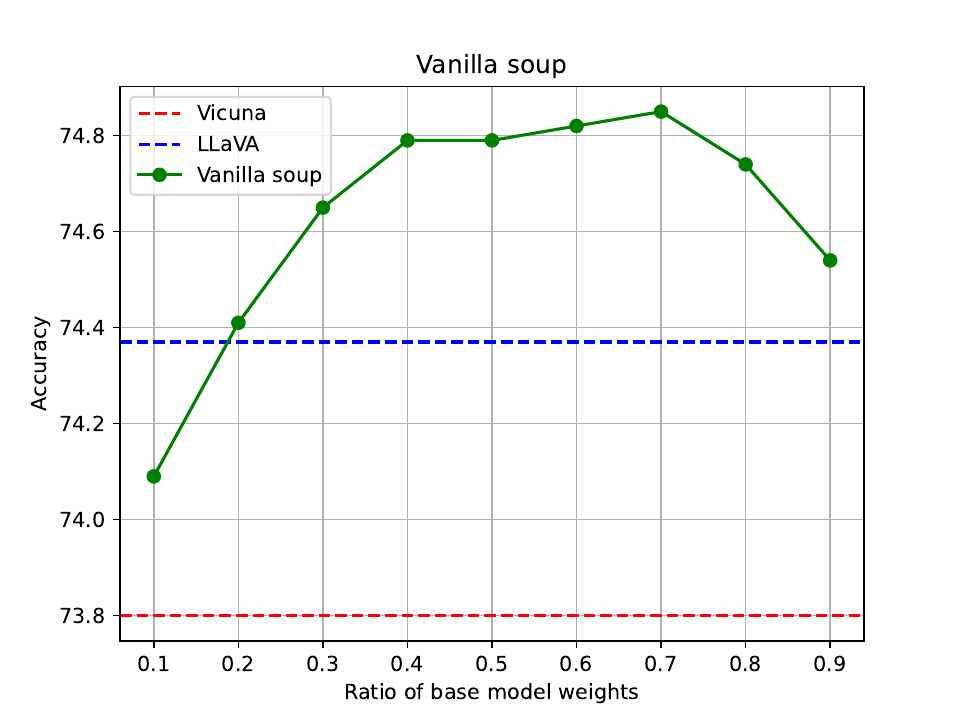}
    \caption{Hellaswag}
  \end{subfigure}

  \caption{Vanilla soup evaluations on five meta sets, including MMMU, LLaVA-Bench for multi-modality, and MMLU, GSM8K, Hellaswag for language.
  The x-axis shows increasing soup ratio from 0.1 to 0.9 of ($\alpha^1$) of LLaVA.
  The y-axis means the evaluation performance.
  Green dots serve as soup performances.
  Two base models are shown in blue and red lines. We find vanilla soup generally outperforms baselines, and direct average with $\alpha^1 = 0.5$ often obtains better results except for the MMMU dataset.
  }\label{fig:vanilla_soup}
\end{figure*}

\section{Experiments}\label{sec:experiments}

\subsection{Principle Design}
Since we study series finegrained soup strategies based on multi-modal models with massive parameters, it is critical to propose a feasible path to manage the hyperparameter spaces for a reasonable exploration pipeline.
Therefore, we briefly introduce \textit{base models}, \textit{meta sets}, and \textit{soup strategies}, then elaborate them in the following sections.

\noindent \textbf{Base Models}

We specifically consider vision-language domains and choose representative Vicuna~\cite{zheng2024judging} and its visual variants LLaVA~\cite{liu2024visual} as two base models.
Vicuna is finetuned from LLaMA~\cite{touvron2023llama} with human conversation instruction, which enable it with chatbot function.
LLaVA is further finetuned from Vicuna using vision-language instructions, therefore, the model can understand visual input and interact with users by language.
Basically, they are both variants from original LLaMA, sharing the isomorphical structures on language decoder, and their weights are consistently optimized step-by-step.
Such consistencies benefits to further explore model interpolation upon these two models.
Specifically, we use their 7B and V1.5 version to represent language and multi-modal domains.
Among our experiments, we fix two base models and only investigate the interpolation weight $\alpha$ based on different soup strategies.
We also fix the visual encoder and alignment MLP of LLaVA for both training and test.
Please note the base model candidates can be easily generalized into other domains (e.g., audio and video) and multiple (>2) base models, but we only take language and vision-language ones in our study.

\noindent \textbf{Meta Sets}

Various evaluation benchmarks are designed for both language and vision-language models from different purposes, we choose a few representative ones as our meta (development) sets for benchmarking.
Such meta sets fulfil: 1) they are well-prepared and robust evaluation datasets for certain general purposes, 2) they cover both language and vision-language multi-modal domains, 3) they contain training and corresponding test set.
In this study, we choose MMMU~\cite{yue2023mmmu}, LLaVA665K~\cite{liu2023improvedllava} for vision-language domain; MMLU~\cite{hendryckstest2021}, GSM8K~\cite{cobbe2021gsm8k}, and Hellaswag~\cite{zellers2019hellaswag} for language domain.
We use their given training set for finetuning and test set for evaluation\footnote{LLaVA665K is the instruction finetuning data for LLaVA without corresponding test set, we regard the LLaVA-Bench~\cite{liu2024visual} as its in-domain test set.}.
The meta sets information is summarized in Tab.~\ref{tab:meta_set}

\noindent \textbf{Soup Strategies}

We study a series of soup strategies that interpolate two base models while fixing their original weights based on five meta sets.
At first, we simply use vanilla soup as a initial baseline (Sec.~\ref{sec:method_vanilla_soup}) to test if such a naive method improves performance on 5 meta sets without complicated experimental designs.
Then, we expound learnable soup (Sec.~\ref{sec:method_learnable_soup}) as the central role in our experiments to 1) fully explore the soup potential for performance gain, 2) statistically depict the soup performance patterns under multiple hyperparameter dimensions.
Finally, other than pursuing better performance, we deploy regularized soup (Sec.~\ref{sec:method_regularized_soup}) to intuitively probe the stability of soup behavior under various regularized training scenarios.



\begin{table*}[htbp]
  \centering
  \caption{Performance summary of different soup strategies on five meta sets.
  It includes two base model baselines and records the best performance of three soup strategies among various configurations.
  }\label{tab:meta_set_comp}
  \resizebox{0.8\textwidth}{!}{
    \begin{tabular}{cccccc}
      \toprule
      Model & MMMU & LLaVA-Bench & MMLU & GSM8k & Hellaswag \\
      \midrule
      Vicuna-7B-v1.5 & 31.00 & 53.90 & 48.75 & 19.33 & 73.80 \\
      LLaVA-7B-v1.5 & 34.22 & 65.20 & 49.54 & 17.89 & 74.37 \\
      Vanilla Soup* & 34.89 & 71.20 & 50.22 & 20.32 & 74.85 \\
      Single Meta-Set* & \textbf{35.78} & \textbf{72.10} & 51.24 & 21.15 & \textbf{74.86} \\
      Pair Meta-Set* & 35.11 & - & \textbf{51.65} & \textbf{21.38} & 74.82 \\
      \bottomrule
    \end{tabular}
    }
\end{table*}

\subsection{Vanilla Soup}\label{sec:vanilla_soup}
Our exploration begins with the simplest vanilla soup.
Given Vicuna and LLaVA as base models, we set $\alpha^1 = \{0.1, 0.2, ..., 0.9\}$ ($\alpha^2$ correspondingly obtained by Eq.~\ref{eq:vanilla_soup}) to merge them and test on meta sets.
Fig.~\ref{fig:vanilla_soup} shows the soup performance (green dots) and two base models as baselines (blue and red lines).
We conclude 1) LLaVA naturally improves vision-language tasks (MMMU and LLaVA-Bench), as it is visually finetuned.
Further, since the visual finetuning also contain language partition, it also enhances two general language-only tasks (MMLU and Hellaswag), but not for GSM8K which is more specific in math.
2) Vanilla soup performs generally better than two baselines proving the soup strategy effectiveness.
3) For 4 out of 5 meta sets (except MMMU), the trending of vanilla soup performance shows half-half average of base models obtains better results compared with other ratios, especially certain extreme cases (e.g., $\alpha^1=0.1, 0.9$).
However, this is not for MMMU which highly relies on the visual finetuning for improvement.
We track the performance comparison in Tab.~\ref{tab:meta_set_comp}

\subsection{Learnable Soup}
After vanilla soup as a simple proof-of-concept validation, we then go into details of learnable soup method, where we elaborate extensive ablation study.
This ablation aims to firstly find if such fine-grained soup can 1) further obtain performance gain compared with vanilla soup, and 2) find statistical soup patterns across several hyperparameter dimensions, helping to understand the soup sensitivity under different settings.
Specifically, given five meta sets for finetuning and evaluation, we cover 1) datasets, 2) epoch, 3) learning rate, 4) sample number, 5) sample ratio, and 6) activation aspects for ablations.
It is hard to systematically discover the global oracle setting, as all dimensions are entangled together.
Therefore, we heuristically design a path to search for the best combination from several rounds of ablation study.
Along with them, we summarize the soup performance patterns in a statistical way.

\noindent \textbf{First Round}

\begin{figure*}[htbp]
  \centering
  \begin{subfigure}[b]{0.19\textwidth}
    \includegraphics[width=\textwidth]{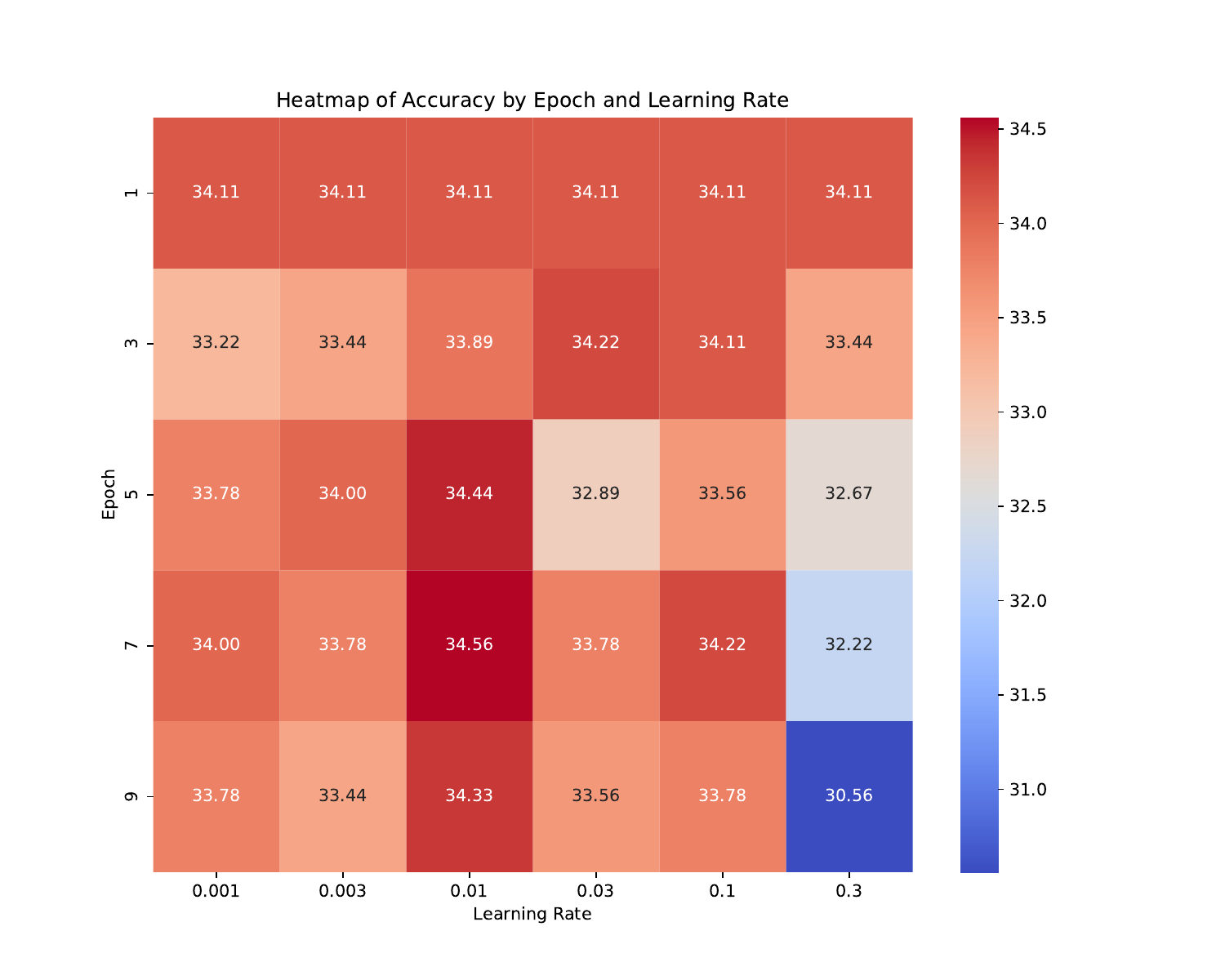}
    \caption{MM-MM}
  \end{subfigure}
  \hfill 
  \begin{subfigure}[b]{0.19\textwidth}
    \includegraphics[width=\textwidth]{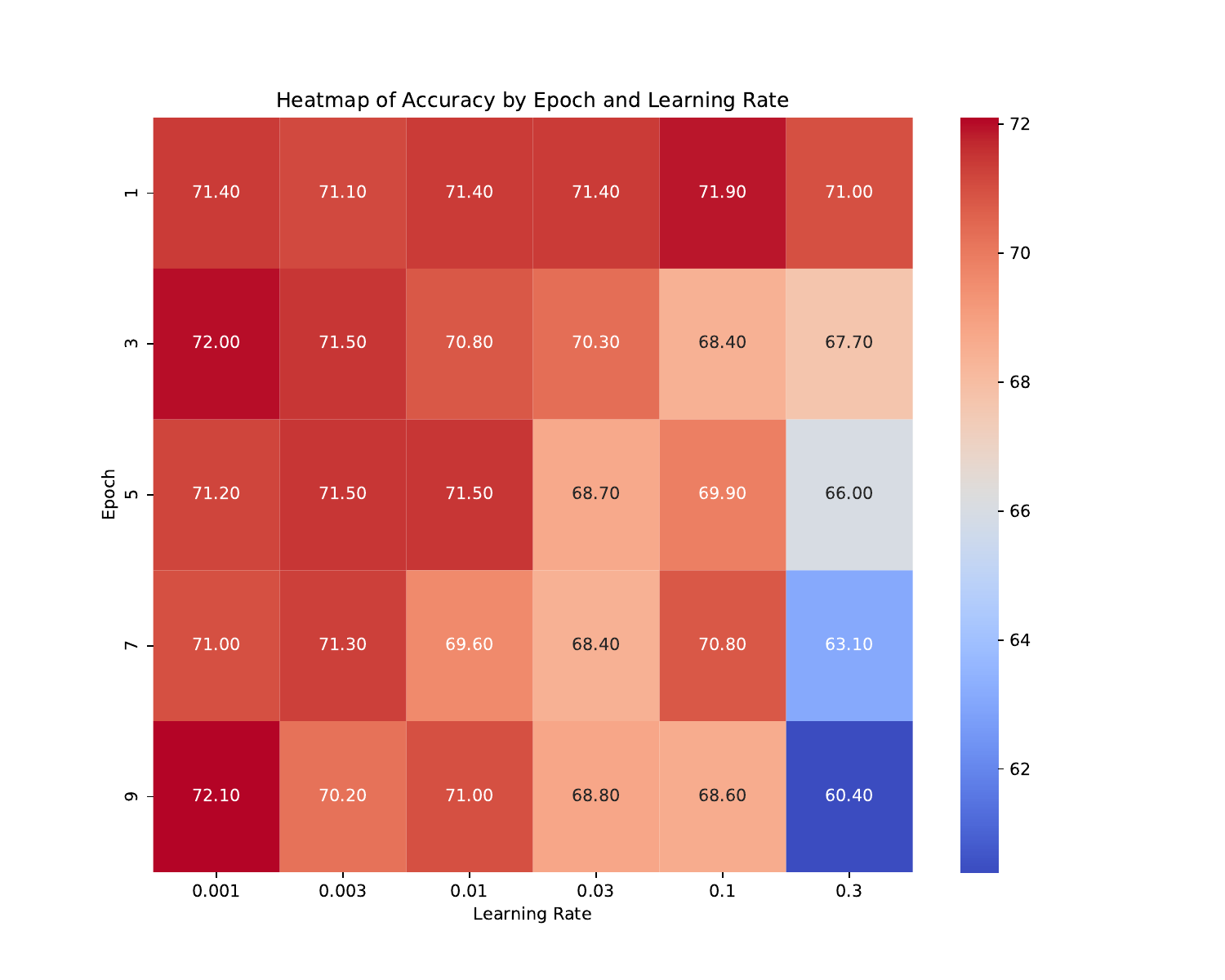}
    \caption{MM-L}
  \end{subfigure}
  \hfill 
  \begin{subfigure}[b]{0.19\textwidth}
    \includegraphics[width=\textwidth]{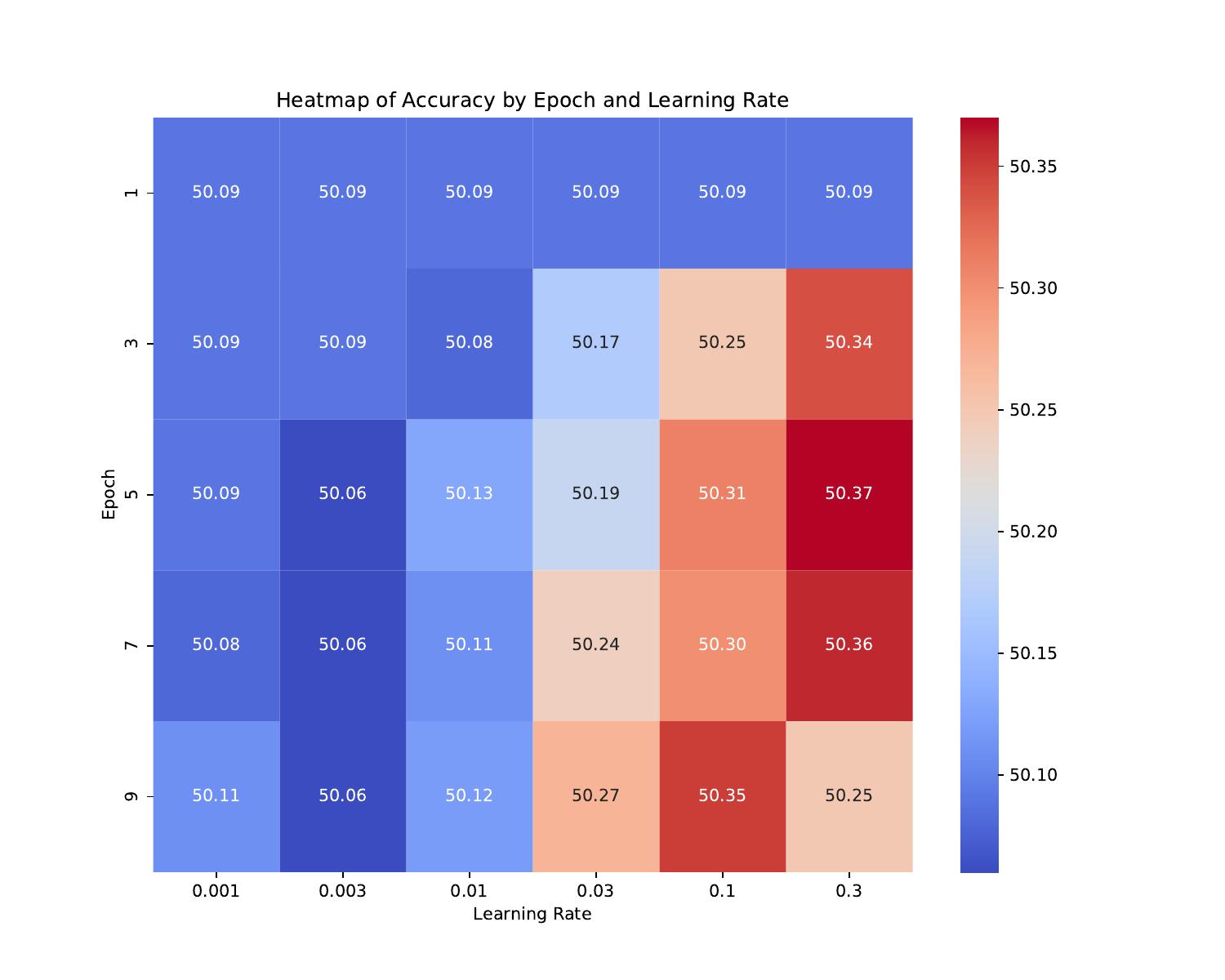}
    \caption{MM-ML}
  \end{subfigure}
  \hfill 
  \begin{subfigure}[b]{0.19\textwidth}
    \includegraphics[width=\textwidth]{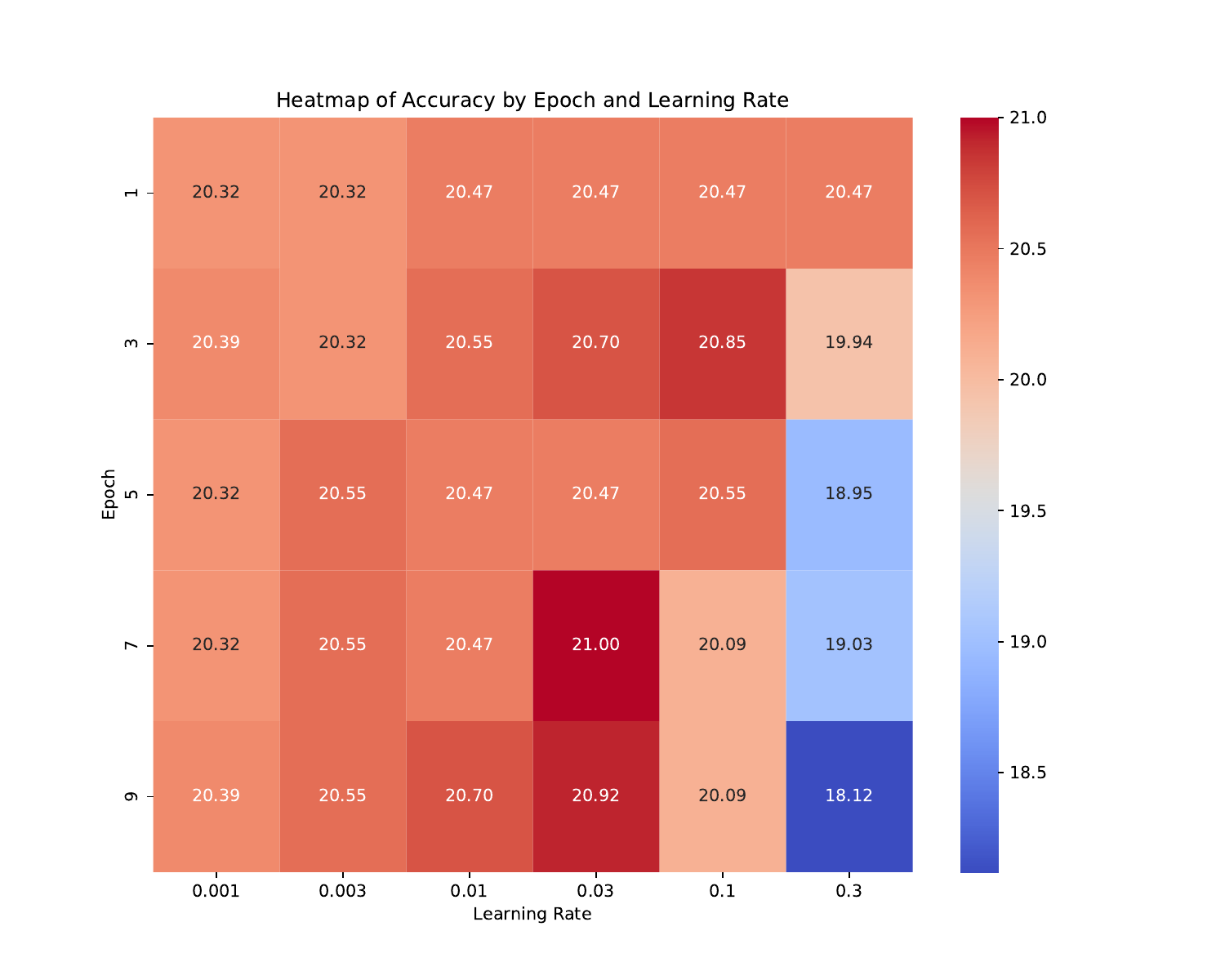}
    \caption{MM-G}
  \end{subfigure}
  \hfill 
  \begin{subfigure}[b]{0.19\textwidth}
    \includegraphics[width=\textwidth]{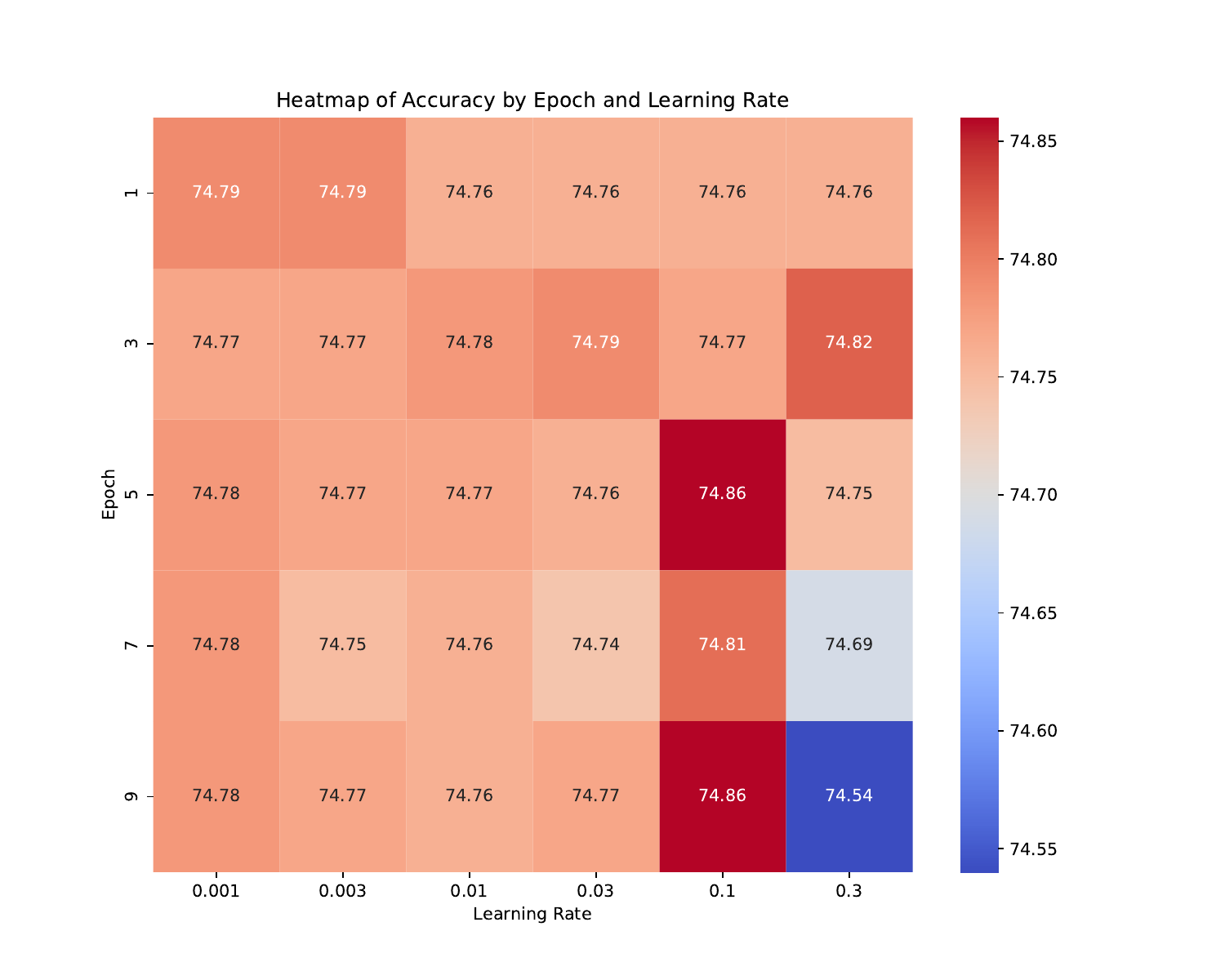}
    \caption{MM-H}
  \end{subfigure}
  \hfill  
  \caption{
  Representative MMMU single set evaluation.
  MM, L, ML, G, and H represent MMMU, LLaVA-Bench, MMLU, GSK8K, and Hellaswag, respectively.
  For each heatmap, x/y axis means ablated learning rates and epochs.
  Different colors show the performance variances on evaluation sets.
  }\label{fig:1_set_ablation_main}
  \vspace{+2mm}
\end{figure*}

We begin with searching for the best meta sets combination by: 1) using each individual meta set to finetune, 2) fixing the total sample number as 1000, 3) ablating the epoch from 1 to 9, 4) ablating the learning rate from 0.001 to 0.3, 5) evaluating on 5 meta sets.
We representatively show a bunch of visualization in Fig.~\ref{fig:1_set_ablation_main}, which uses MMMU as finetuning set.
The rest visualizations are supplemented in Fig.~\ref{fig:1_set_ablation_supp} in appendix due to the limited space.
Corresponding performances are also tracked in Tab.~\ref{tab:meta_set_comp}.
To summarize all visualizations, we calculate the mean and maximum performance of 5 meta sets across epochs and learning rates in Tab.~\ref{tab:1_set_stats}.
We conclude 1) finegrained learnable soup outperforms vanilla soup for each evaluation task, obtaining further performance gain compared with two baselines.
However, the best results of each meta set are based on different hyperparameter settings.
Due to the different properties of training and evaluation sets, the soup performance varies significantly among them.
2) There are clear trends of performance changes with ablated learning rates and epochs (color changes in heatmap plots), indicating a clear hyperparameter patterns at least within one meta set, but may change across meta sets.
3) The soup patterns dramatically differs across different training-evaluation sets combination.
For example, MM-MM observes the best combination in the middle with the worst at bottom right corner, but MM-ML shows completely different clues.
4) Based on the results in Tab.~\ref{tab:1_set_stats}, we find LLaVA665K is better than MMMU to be chosen in multi-modal domain.
MMLU and Hellaswag show their advantages in language-only domain.
Considering, MMLU follows the multiple-choice task instead of typical natural language, thus we choose MMLU instead of Hellaswag.

As a summary, the first round ablation results in 1) learnable soup further improves the evaluation performance, 2) soup performance patterns change dramatically across different finetuning and test set combinations, but show clear pattern given a fixed training and test pair, and
3) overall, we use LLaVA665K and MMLU as training sets for following ablation rounds.

\begin{table*}[htbp]
  \centering
  \caption{Statistical summary of first round ablation: mean/max accuracy across epochs and learning rates}\label{tab:1_set_stats}
  \resizebox{0.98\textwidth}{!}{
    \begin{tabular}{ccccccc}
      \toprule
      Meta\textbackslash Eval & MMMU & LLaVA-Bench & MMLU & GSM8k & Hellaswag & Sum\\
      \midrule
      MMMU & 33.68/34.56 & 69.77/72.10 & 50.17/50.37 & 20.29/21.00 & 74.77/74.86 & 248.68/252.89 \\
      LLaVA665k & 34.63/35.78 & 69.55/72.10 & 49.99/50.20 & 20.13/21.08 & 74.72/74.84 & \textbf{249.02}/\textbf{254.00} \\
      \midrule
      MMLU & 32.48/34.89 & 64.56/71.30 & 50.53/51.13 & 19.46/21.15 & 74.58/74.81 & \textbf{241.61}/253.28 \\
      GSM8K & 31.65/34.33 & 60.97/71.80 & 50.37/51.24 & 19.24/21.00 & 74.51/74.86 & 236.74/253.23 \\
      Hellaswag & 31.64/35.11 & 60.02/71.80 & 50.21/51.03 & 18.98/21.00 & 74.35/74.84 & 235.20/\textbf{253.78} \\
      \bottomrule
    \end{tabular}
    }
    \vspace{+1mm}
\end{table*}

\noindent \textbf{Second Round}

Using LLaVA665K and MMLU as meta sets, we conduct the second round ablation study.
It aims to find the best hyperparameter setting including 1) learning rate, 2) epoch, 3) sample number, and 4) activation.
Concretely, we 1) fixing the training data as LLaVA665K and MMLU, 2) ablating sample numbers from 10 to 1000, 2) ablating learning rate from 0.001 to 0.3, 3) ablating epoch from 1 to 9, 4) ablating activation using \textit{sigmoid}, \textit{linear}, \textit{clamp}, and \textit{softmax} options\footnote{The implementation details of activation: We initialize the $\alpha$ as 0, 0.5, 0.5, and (0.5, 0.5) for sigmoid, linear, clamp, and softmax, respectively, where we finetune $\alpha^1$ and $\alpha^2$ for softmax and only learn $\alpha^1$ and $\alpha^2 = 1- \alpha^1$ for the rest.

For sigmoid, linear, and clamp, we apply sigmoid, keep it the same, or clamp (from 0 to 1) operation on $\alpha^1$, then, obtain $\alpha^2=1-\alpha^1$.
For softmax, we directly apply softmax operation on $\alpha^1$ and $\alpha^2$.
}.
Please note, from this round, we only evaluate four meta sets except for LLaVA-Bench here due to its massive request of OpenAI API.
We show the representative visualization in Fig.~\ref{fig:2_set_ablation_main} and the rest visualizations are supplemented in the appendix (Fig.~\ref{fig:2_set_ablation_supp}) due to the limited space.
We conclude 1) using LLaVA665K and MMLU as paired meta sets further improve the performance but not significantly.
Similarly, the best setting for each evaluation task varies, indicating the soup process is sensitive to specific test set.
2) The performance changes are still clear given a fixed finetuning and test combination across learning rate, epoch, and activation, however, not consistent while varying the number of samples.
Especially for MMLU task, the trend changes reversely as the number of sample increases.
3) The activation choice affects performances by a large margin such as the linear activation dramatically affect the performance, and overall the other options perform better than linear.
We track the pair meta sets results in Tab.~\ref{tab:meta_set_comp} and we search the best setting based on overall performance on meta sets.
The statistical summary is given by Fig.~\ref{fig:2nd_round_stats} in appendix and we choose the best setting with 3 epoch, 50 sample, 0.1 learning rate, and softmax activation.

As a summary, given LLaVA665K and MMLU as meta sets, the second round ablation search the epoch, sample number, learning rate, and activations.
We find the little performance gain compared with the first round and the soup performances vary across differen settings.
The best overall setting is picked for the next round ablation.

\noindent \textbf{Third Round}

\begin{figure*}[htbp]
  \centering
  \begin{subfigure}[b]{0.24\textwidth}
    \includegraphics[width=\textwidth]{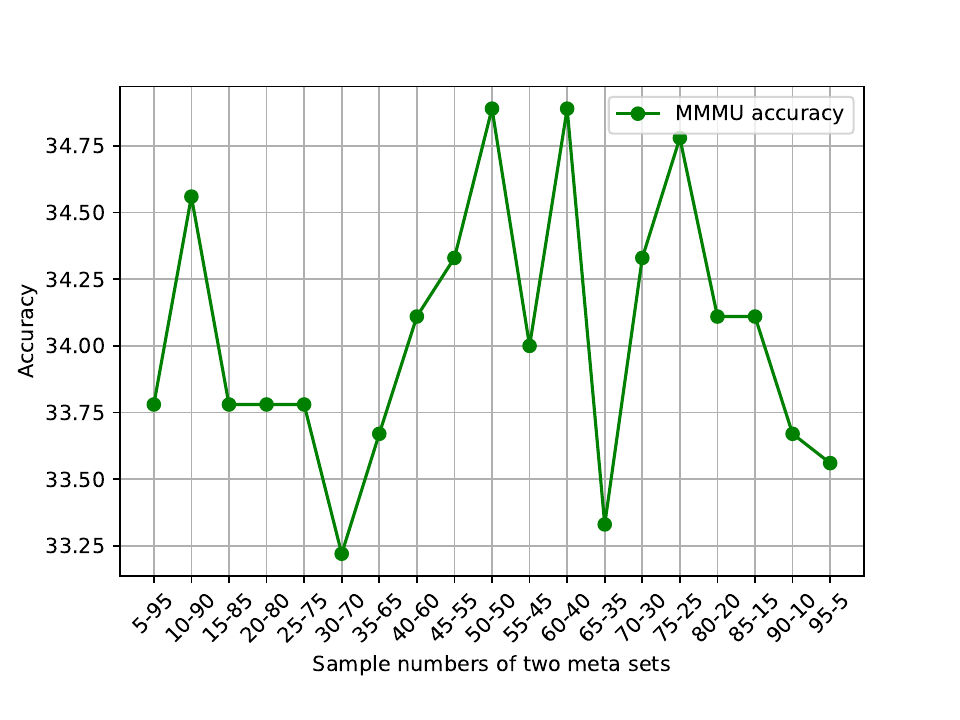}
    \caption{MMMU}
  \end{subfigure}
  \hfill 
  \begin{subfigure}[b]{0.24\textwidth}
    \includegraphics[width=\textwidth]{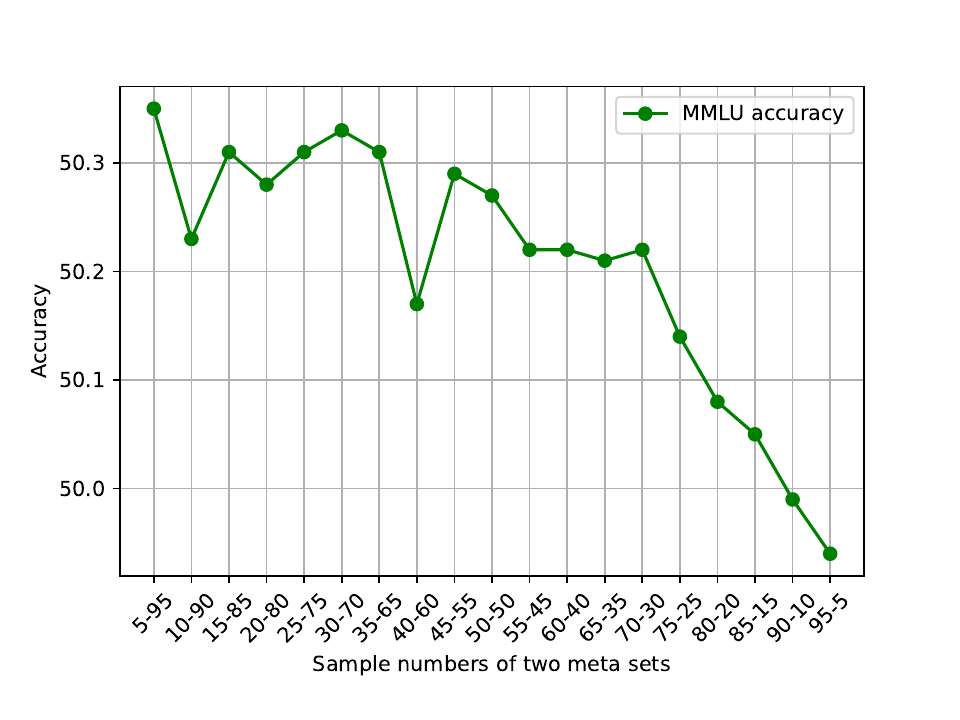}
    \caption{MMLU}
  \end{subfigure}
  \hfill 
  \begin{subfigure}[b]{0.24\textwidth}
    \includegraphics[width=\textwidth]{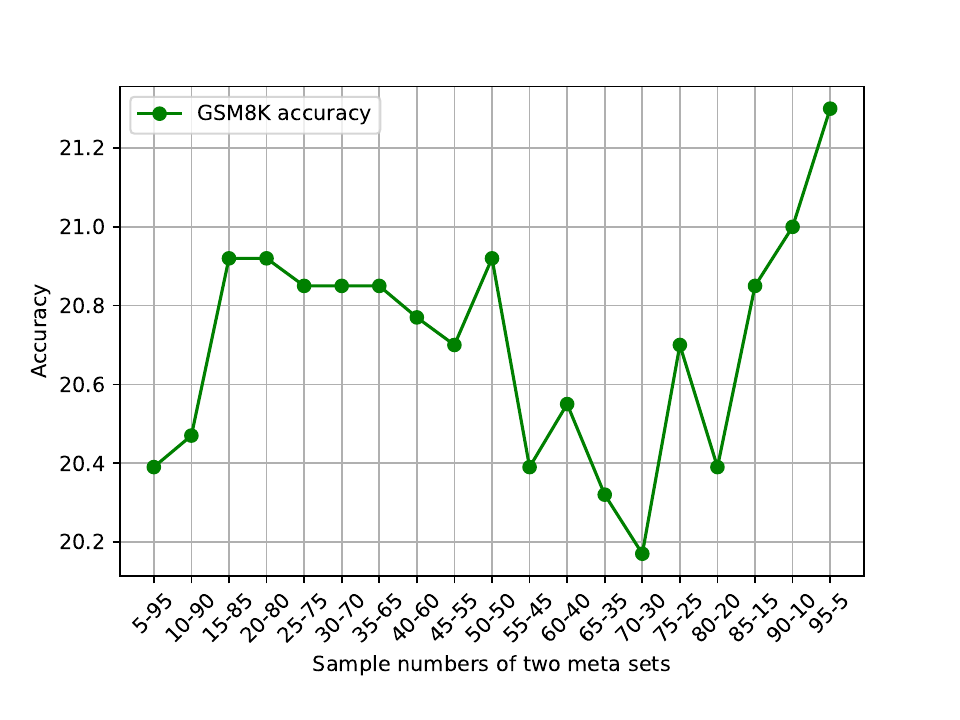}
    \caption{GSM8K}
  \end{subfigure}
  \hfill 
  \begin{subfigure}[b]{0.24\textwidth}
    \includegraphics[width=\textwidth]{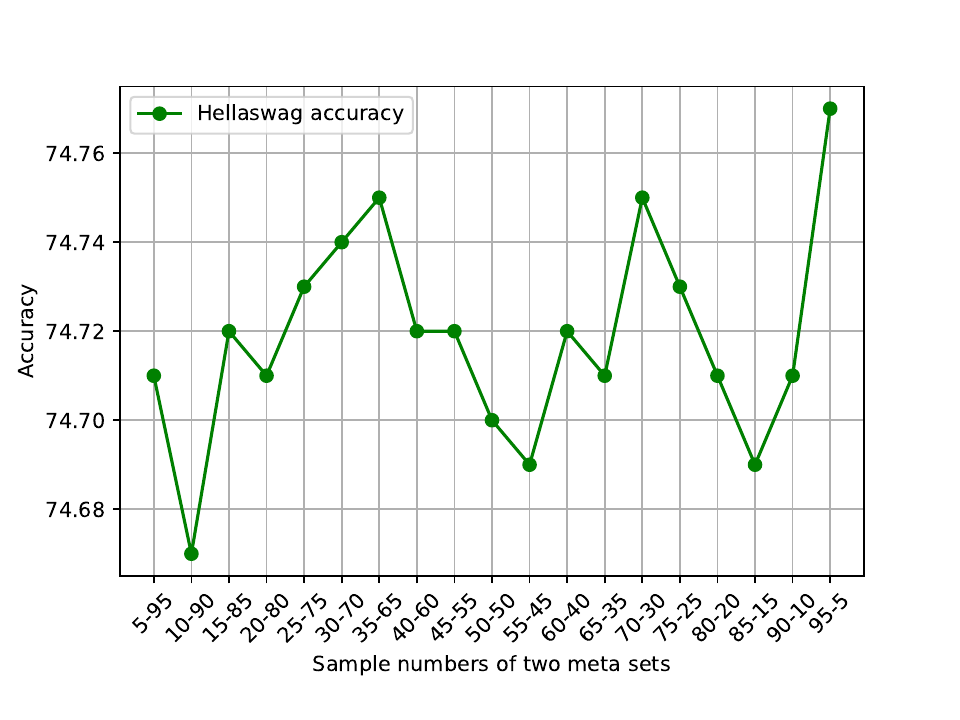}
    \caption{Hellaswag}
  \end{subfigure}

  \caption{Ratio ablation on MMMU, MMLU, GSM8K, and Hellaswag on LLaVA665K and MMLU meta sets.}\label{fig:ratio}
\end{figure*}

We finally make ablation on the ratio of given meta sets as the last round.
Given the setting from first and second round, we adjust the sample ratio from LLaVA665K and MMLU from 5-95 to 95-5 to test if the ratio is a sensitive factor for evaluation.
Performance variances are shown in Fig.~\ref{fig:ratio}.
We conclude there are no clear trend according to the sample ratio based on the given setting, except for the MMLU task.
Overall, the 50-50 ratio achieves the averagely better results than others.
Through the three rounds heuristic ablations, we benchmark the soup performance on 5 meta sets, covering several hyperparameter configurations and fully exploring the model soup potential.
Statistically, we find the better configurations and provide intuitions of the hyperparameter properties for SoupLM.

\begin{figure*}[htbp]
  \centering
  \begin{subfigure}[b]{0.24\textwidth}
    \includegraphics[width=\textwidth]{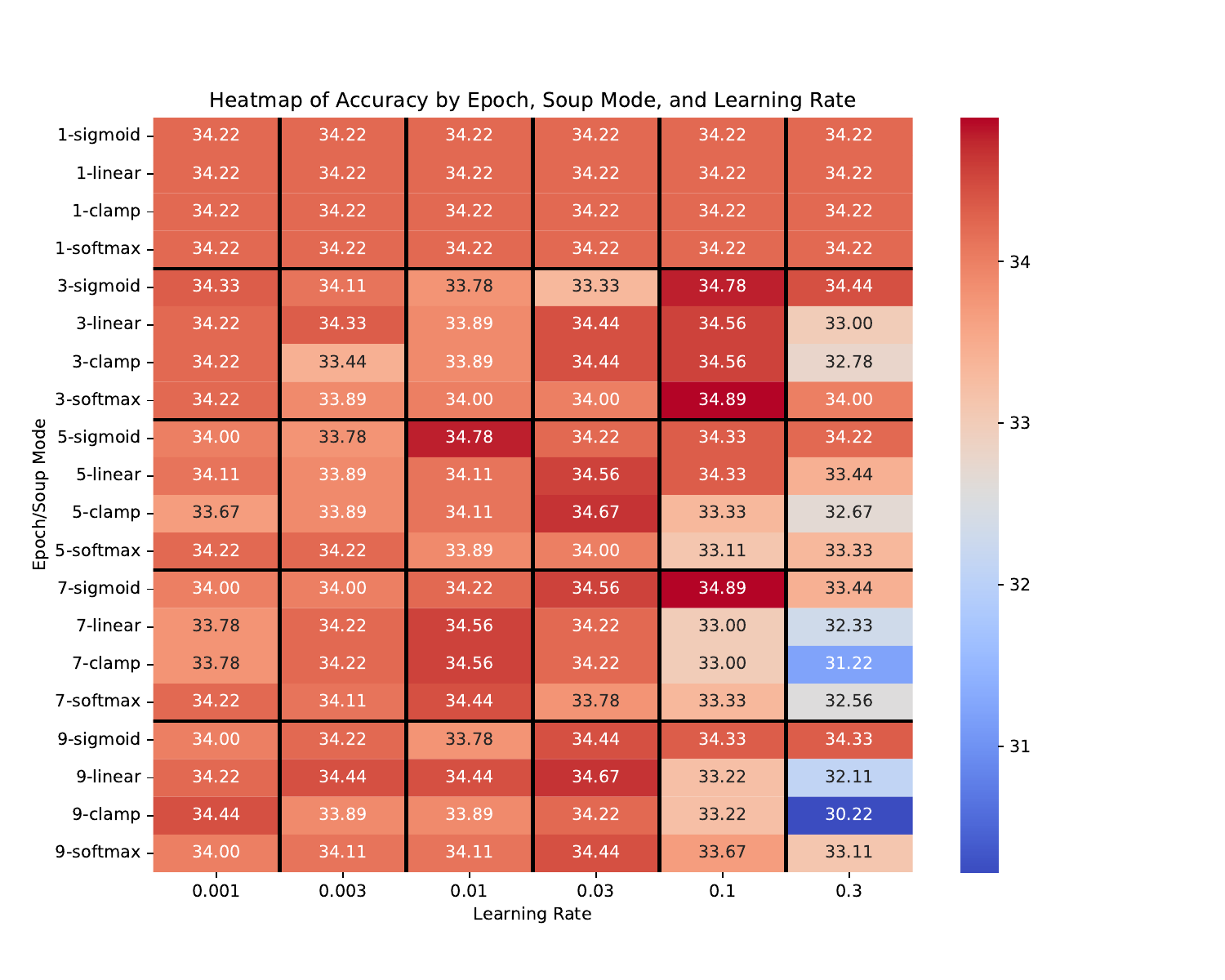}
    \caption{50:MM}
  \end{subfigure}
  \hfill 
  \begin{subfigure}[b]{0.24\textwidth}
    \includegraphics[width=\textwidth]{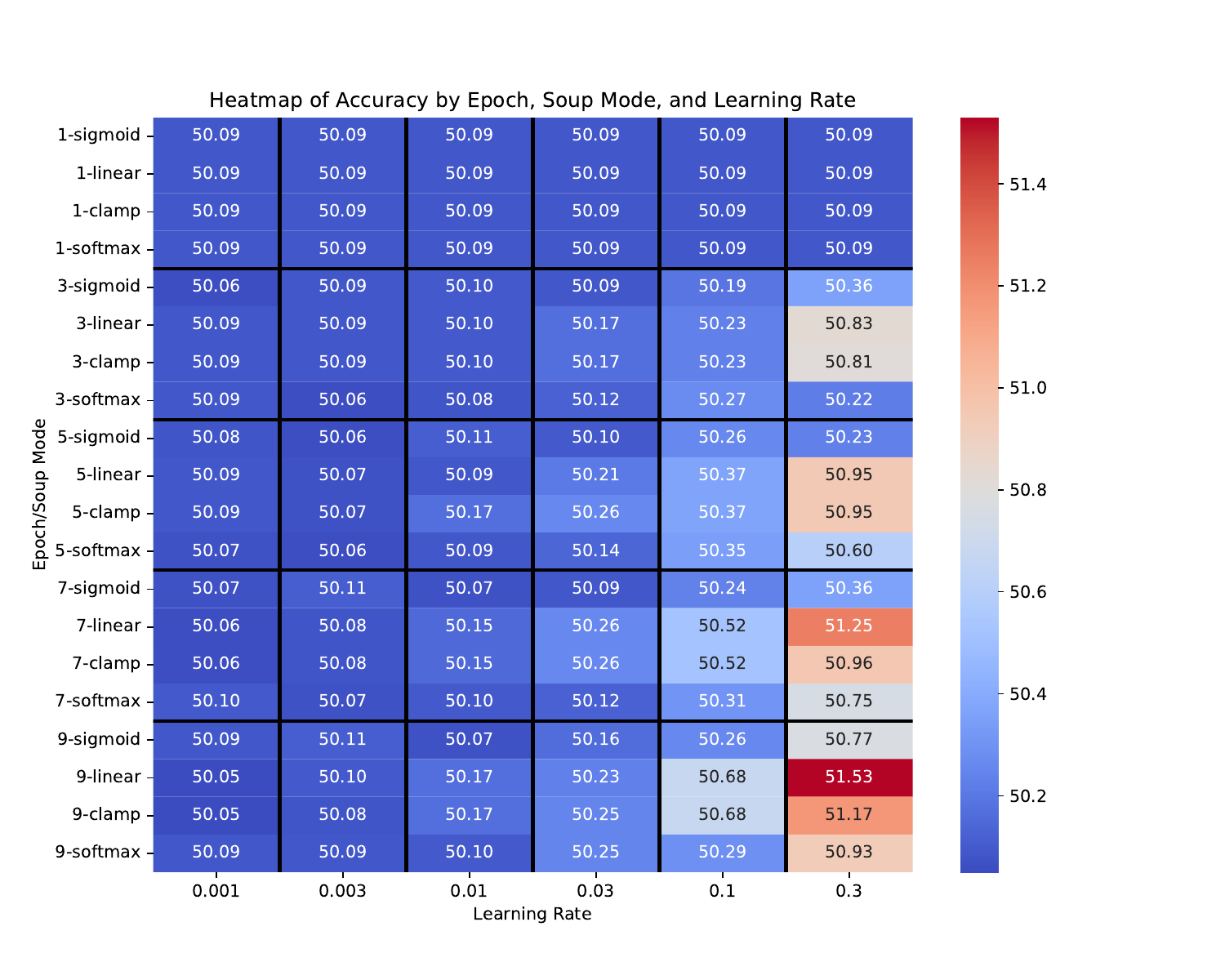}
    \caption{50:ML}
  \end{subfigure}
  \hfill 
  \begin{subfigure}[b]{0.24\textwidth}
    \includegraphics[width=\textwidth]{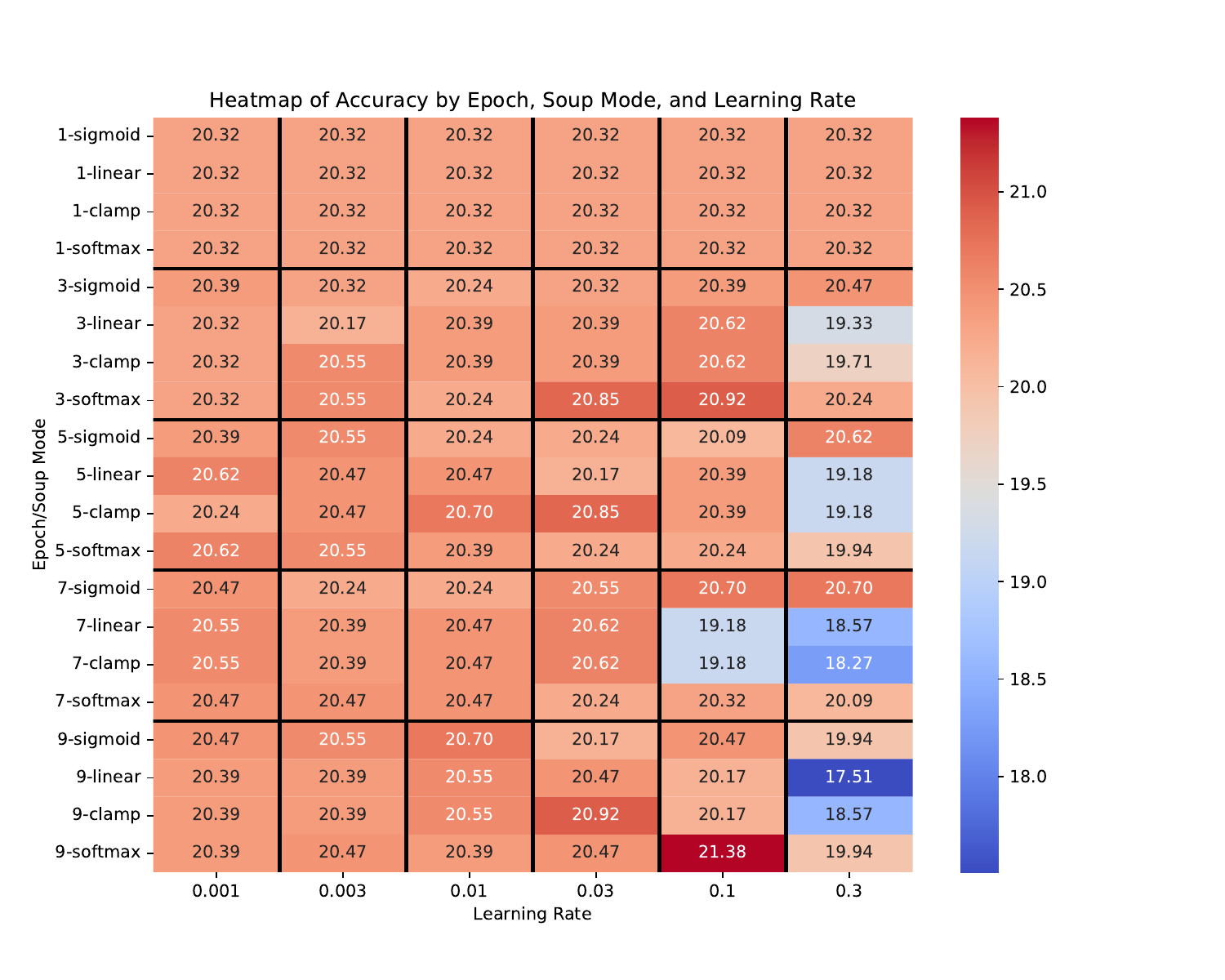}
    \caption{50:G}
  \end{subfigure}
  \hfill 
  \begin{subfigure}[b]{0.24\textwidth}
    \includegraphics[width=\textwidth]{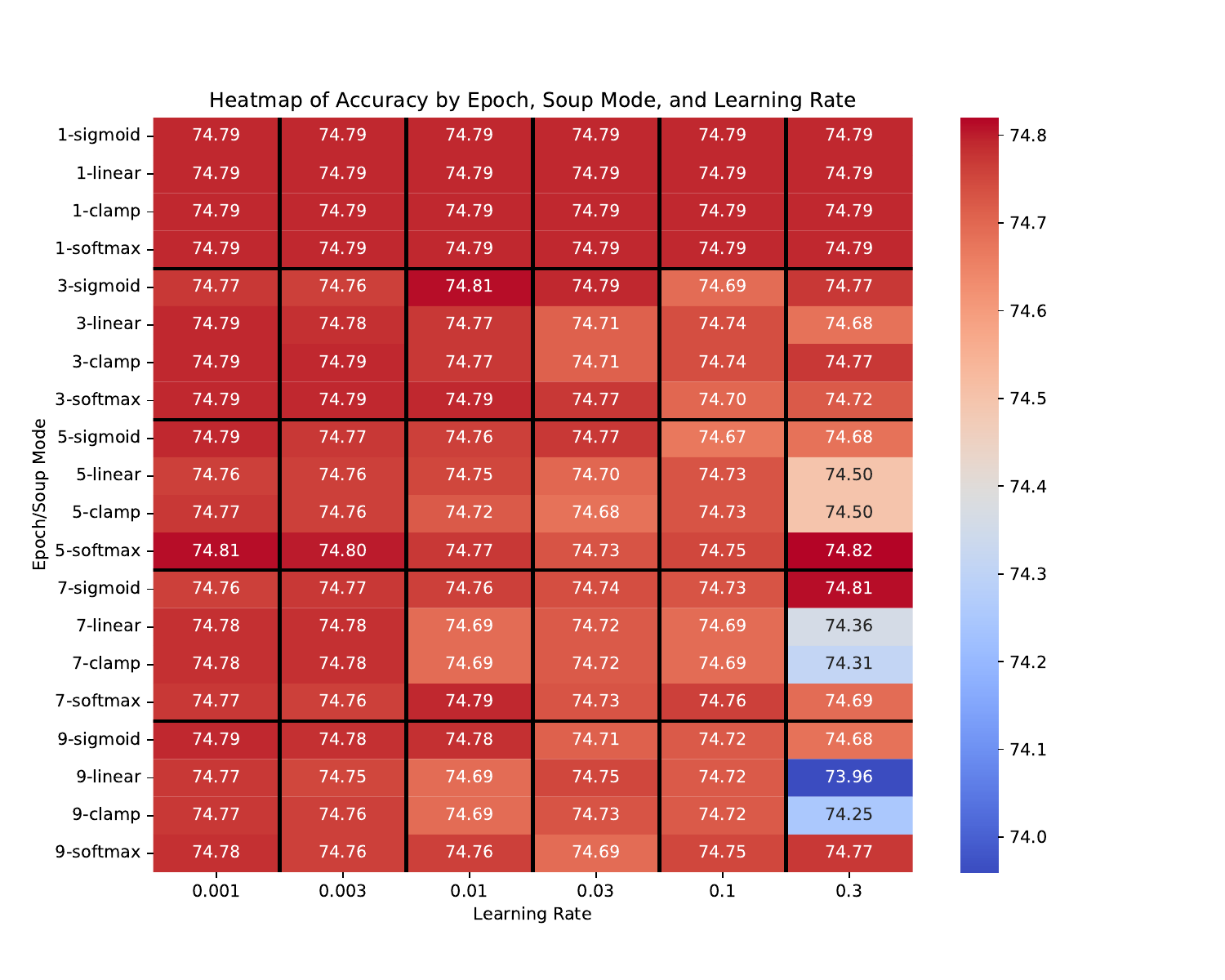}
    \caption{50:H}
  \end{subfigure}
  \caption{Second round ablation for epoch, sample number, learning rate, and activation.
  MM, ML, G, H are for MMMU, MMLU, GSM8K, Hellaswag.
  Colors show performance changes.
  X-axis is learning rate.
  Y-axis is number of epoch and activation function.
  Here, we use 50 samples for LLaVA665K and 50 samples for MMLU.
  }\label{fig:2_set_ablation_main}
\end{figure*}

\noindent \textbf{More Evaluations}

Using the best soup setting from three rounds ablation, we evaluate its soup performance on more diverse evaluation tasks other than given five meta sets.
We choose Winoground~\cite{thrush2022winoground}, PiQA~\cite{bisk2020piqa}, MathQA~\cite{amini2019mathqa}, BoolQA~\cite{clark2019boolq}, and BBH~\cite{suzgun2022challenging} for language and POPE~\cite{li2023evaluating} and MM-Bench~\cite{liu2023mmbench} for vision-language domains.
Tab.~\ref{tab:more_eval} in appendix shows model soup generally outperforms baselines, but may also drop the performance for certain tasks, which may due to severe domain drift such as MathQA.

\section{Soup Behavior}\label{sec:soup_behavior}
Beside of discussing performance gain, we initially study the soup behavior based on empirical results (Sec.~\ref{sec:experiments}) and regularized soup (Sec.~\ref{sec:method_regularized_soup}).
We are curious if the soup dynamics follow certain patterns under different training constraints and supervisions.
We first probe such behavior through visualizing the learned $\alpha$ from different meta sets.
Since we are only curious about its distribution, we tune the $\alpha$
with 0.3 learning rate, 9 epochs, and 1000 samples to ensure it is fully optimized.
We visualize an exemplar case of key mapping across the language decoder layers (Fig.~\ref{fig:set_dist_main}).
We visualize the rest of visualizations in the appendix (Sec.~\ref{sec:alpha_dist_supp}) including other mappings, normalization layers, etc.
Furthermore, we set series of regularization magnitudes to observe if the soup behavior varies under training constraints.
We visualize the regularized soup of key mapping with 0.0001 magnitude in Fig.~\ref{fig:reg_dist_main} and leave the rest magnitudes in appendix (Sec.~\ref{sec:reg_alpha_dist_supp}).
Figures show how the two base models are integrated into the souped model.
Through the x-axis, they show different meta sets across different layers.
Y-axis indicates the learned ratios between Vicuna and LLaVA.
If the ratio is more than 0.5, meaning the corresponging base model dominates the soup process for this mapping, we color it as green, otherwise, as red.
According to these figures, we observe 1) for some layers, the color distributions are very neat across different meta set, while for some others, these consistencies are not stable.
2) For the $\alpha$ under regularized soup, we find the soup trends are not vulnerable, only generally close to the initial value 0.5 as constrained by the regularization.
3) Please note different mappings may show varied distributions and see more cases in the appendix.
Overall, we draw the conclusions that the soup behaviors are not vulnerable under regularized constraints, and show consistency across certain layers but may vary different layers and mappings.
In this study, we initially probe the soup behavior to provide intuitions by visualizations, and hope it inspires more model interpolation mechanism explorations.

\begin{figure*}[htbp]
  \centering
  \begin{subfigure}[b]{0.98\textwidth}
    \includegraphics[width=\textwidth]{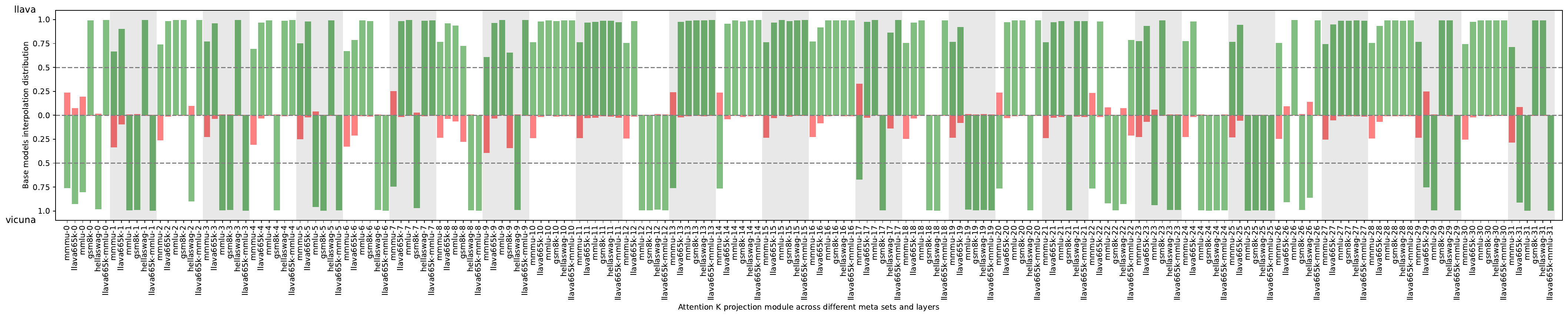}
  \end{subfigure}
  \caption{Learned alpha distribution
on LLaVA-Vicuna model space of key mapping across different meta sets and Transformer layers.
This set of $\alpha$ is tuned on 9 epochs, 0.3 learning rate, and 1000 samples.
Certain layers show stable consistency across different meta sets.
}\label{fig:set_dist_main}
\end{figure*}

\begin{figure*}[htbp]
  \centering
  \begin{subfigure}[b]{0.98\textwidth}
    \includegraphics[width=\textwidth]{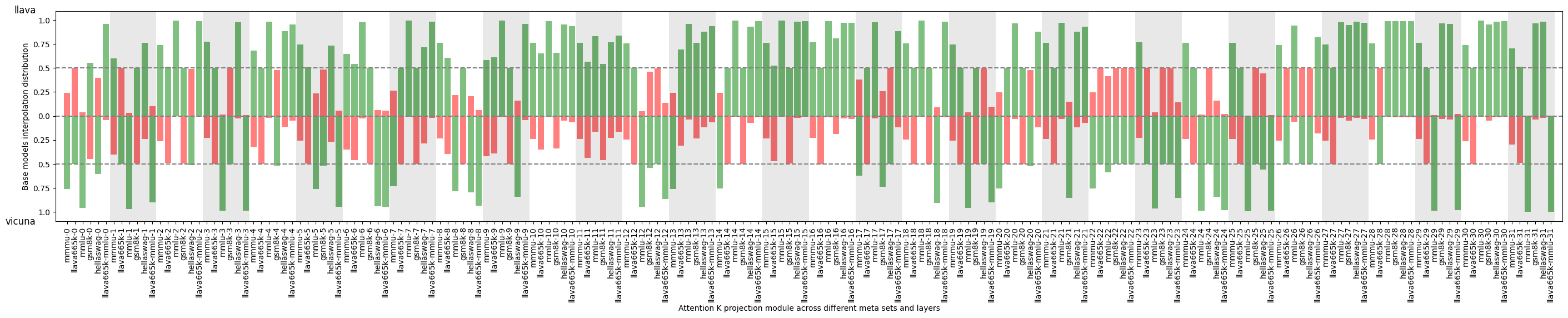}
  \end{subfigure}
  \caption{Learned alpha distribution with 0.0001 regularization.
  It follows the same finetuning settings as figure above.
  The regularization limits the $\alpha$ values close to the initial 0.5 but shows the same $\alpha$ distribution with the unregularized one.
  }\label{fig:reg_dist_main}
\end{figure*}



\section{Related Work}
\subsection{Large Language and Multi-Modal Models}
Large-scale language models (LLMs) show that large-scale pretraining enables model with strong language capacity with massive knowledge~\cite{radford2018improving, radford2019language, brown2020language, devlin2018bert, liu2019roberta, touvron2023llama}.
Downstream finetuning improves task performances and aligns the model behavior with human preference~\cite{ouyang2022training, zheng2024judging, zhang2023llama, taori2023alpaca, wang2022self, ziegler2019fine, stiennon2020learning}.
Centered around pretrained LLMs, their model variants are widely extended to other domains by finetuning with instruction datasets~\cite{achiam2023gpt, reid2024gemini, huang2024audiogpt,xu2023pointllm,liu2024visual,lin2023video}.
Instead of finetuning a pretrained LM, multi-modal capacity can be also obtained simultaneously by training a unified model from scratch~\cite{lu2022unified,lu2023unified,luo2020univl, tang2024any, pan2023kosmos, jin2023unified, koh2024generating}.
SoupLM proposes to efficiently assemble model variants to deliver a well-generalized one without extra training cost.

\subsection{Model Soup}
Model soup (weight averaging) is widely used to study optimization process~\cite{ahmadianfar2022info, bansal2011inertia}.
Many works study how it works on improving neural network capacity or analyze the model behavior~\cite{nowlan2018simplifying, blundell2015weight}.
For large-scale networks, model soup is firstly studied by ~\cite{wortsman2022model}.
It benchmarks the soup method on image classification with different backbones, and obtain free performance gain with no inference cost, which is critical for large-scale models.
Soup strategy also benefits to enhance adapter structure~\cite{chronopoulou2023adaptersoup}, personalized finetuning~\cite{jang2023personalized}, continue training~\cite{akiba2024evolutionary}, etc, for language models.
Different from existing works, our work explores model soup for large language and vision-language models in a cross-domain fashion with more general purposes.


\section{Conclusion}
We propose SoupLM to first explore the model soup strategy in autoregressive large language models (LLMs) and large multi-modal models (LMMs).
This study takes Vicuna and LLaVA as a study case to 1) propose series soup strategies to fully explore the model soup potential pursuing performance gain, 2) statistically benchmark learnable soup capacity across systematically designed configuration space and observe comprehensive hyperparameter patterns, 3) initially probe the soup behavior to observe its consistent property across configurations and regularizations.
SoupLM efficiently assembles isomorphical model variants into a well-generalized one that handles multiple domains, with no inference and ignorable training costs.
It inspires to fast integrate and iterate large-scale models with multiple domain capacities while avoiding costly additional training efforts.

\section{Limitations}
We propose SoupLM to merge LLM and LMM into a well-generalized model that handles both language and vision-language domains.
However, due to the massive computational requirements to benchmark the model soup for large-scale models, 1) we only take two base models with 7B model size as a study case, which can be easily extended into more general cases, 2) we only provide a heuristic design to benchmark the soup performance on base models, since it is almost not feasible to find the oracle setting among several configuration dimensions.
We leave more general studies in our future work.
\bibliography{arxiv}

\begin{thebibliography}{62}
\providecommand{\natexlab}[1]{#1}

\bibitem[{Achiam et~al.(2023)Achiam, Adler, Agarwal, Ahmad, Akkaya, Aleman, Almeida, Altenschmidt, Altman, Anadkat et~al.}]{achiam2023gpt}
Josh Achiam, Steven Adler, Sandhini Agarwal, Lama Ahmad, Ilge Akkaya, Florencia~Leoni Aleman, Diogo Almeida, Janko Altenschmidt, Sam Altman, Shyamal Anadkat, et~al. 2023.
\newblock Gpt-4 technical report.
\newblock \emph{arXiv preprint arXiv:2303.08774}.

\bibitem[{Ahmadianfar et~al.(2022)Ahmadianfar, Heidari, Noshadian, Chen, and Gandomi}]{ahmadianfar2022info}
Iman Ahmadianfar, Ali~Asghar Heidari, Saeed Noshadian, Huiling Chen, and Amir~H Gandomi. 2022.
\newblock Info: An efficient optimization algorithm based on weighted mean of vectors.
\newblock \emph{Expert Systems with Applications}, 195:116516.

\bibitem[{Akiba et~al.(2024)Akiba, Shing, Tang, Sun, and Ha}]{akiba2024evolutionary}
Takuya Akiba, Makoto Shing, Yujin Tang, Qi~Sun, and David Ha. 2024.
\newblock Evolutionary optimization of model merging recipes.
\newblock \emph{arXiv preprint arXiv:2403.13187}.

\bibitem[{Amini et~al.(2019)Amini, Gabriel, Lin, Koncel-Kedziorski, Choi, and Hajishirzi}]{amini2019mathqa}
Aida Amini, Saadia Gabriel, Peter Lin, Rik Koncel-Kedziorski, Yejin Choi, and Hannaneh Hajishirzi. 2019.
\newblock Mathqa: Towards interpretable math word problem solving with operation-based formalisms.
\newblock \emph{arXiv preprint arXiv:1905.13319}.

\bibitem[{Bansal et~al.(2011)Bansal, Singh, Saraswat, Verma, Jadon, and Abraham}]{bansal2011inertia}
Jagdish~Chand Bansal, PK~Singh, Mukesh Saraswat, Abhishek Verma, Shimpi~Singh Jadon, and Ajith Abraham. 2011.
\newblock Inertia weight strategies in particle swarm optimization.
\newblock In \emph{2011 Third world congress on nature and biologically inspired computing}, pages 633--640. IEEE.

\bibitem[{Bisk et~al.(2020)Bisk, Zellers, Gao, Choi et~al.}]{bisk2020piqa}
Yonatan Bisk, Rowan Zellers, Jianfeng Gao, Yejin Choi, et~al. 2020.
\newblock Piqa: Reasoning about physical commonsense in natural language.
\newblock In \emph{Proceedings of the AAAI conference on artificial intelligence}, volume~34, pages 7432--7439.

\bibitem[{Blundell et~al.(2015)Blundell, Cornebise, Kavukcuoglu, and Wierstra}]{blundell2015weight}
Charles Blundell, Julien Cornebise, Koray Kavukcuoglu, and Daan Wierstra. 2015.
\newblock Weight uncertainty in neural network.
\newblock In \emph{International conference on machine learning}, pages 1613--1622. PMLR.

\bibitem[{Brown et~al.(2020)Brown, Mann, Ryder, Subbiah, Kaplan, Dhariwal, Neelakantan, Shyam, Sastry, Askell et~al.}]{brown2020language}
Tom Brown, Benjamin Mann, Nick Ryder, Melanie Subbiah, Jared~D Kaplan, Prafulla Dhariwal, Arvind Neelakantan, Pranav Shyam, Girish Sastry, Amanda Askell, et~al. 2020.
\newblock Language models are few-shot learners.
\newblock \emph{Advances in neural information processing systems}, 33:1877--1901.

\bibitem[{Choquette(2023)}]{choquette2023nvidia}
Jack Choquette. 2023.
\newblock Nvidia hopper h100 gpu: Scaling performance.
\newblock \emph{IEEE Micro}.

\bibitem[{Chronopoulou et~al.(2023)Chronopoulou, Peters, Fraser, and Dodge}]{chronopoulou2023adaptersoup}
Alexandra Chronopoulou, Matthew~E Peters, Alexander Fraser, and Jesse Dodge. 2023.
\newblock Adaptersoup: Weight averaging to improve generalization of pretrained language models.
\newblock \emph{arXiv preprint arXiv:2302.07027}.

\bibitem[{Clark et~al.(2019)Clark, Lee, Chang, Kwiatkowski, Collins, and Toutanova}]{clark2019boolq}
Christopher Clark, Kenton Lee, Ming-Wei Chang, Tom Kwiatkowski, Michael Collins, and Kristina Toutanova. 2019.
\newblock Boolq: Exploring the surprising difficulty of natural yes/no questions.
\newblock \emph{arXiv preprint arXiv:1905.10044}.

\bibitem[{Cobbe et~al.(2021)Cobbe, Kosaraju, Bavarian, Chen, Jun, Kaiser, Plappert, Tworek, Hilton, Nakano, Hesse, and Schulman}]{cobbe2021gsm8k}
Karl Cobbe, Vineet Kosaraju, Mohammad Bavarian, Mark Chen, Heewoo Jun, Lukasz Kaiser, Matthias Plappert, Jerry Tworek, Jacob Hilton, Reiichiro Nakano, Christopher Hesse, and John Schulman. 2021.
\newblock Training verifiers to solve math word problems.
\newblock \emph{arXiv preprint arXiv:2110.14168}.

\bibitem[{Devlin et~al.(2018)Devlin, Chang, Lee, and Toutanova}]{devlin2018bert}
Jacob Devlin, Ming-Wei Chang, Kenton Lee, and Kristina Toutanova. 2018.
\newblock Bert: Pre-training of deep bidirectional transformers for language understanding.
\newblock \emph{arXiv preprint arXiv:1810.04805}.

\bibitem[{Dosovitskiy et~al.(2020)Dosovitskiy, Beyer, Kolesnikov, Weissenborn, Zhai, Unterthiner, Dehghani, Minderer, Heigold, Gelly et~al.}]{dosovitskiy2020image}
Alexey Dosovitskiy, Lucas Beyer, Alexander Kolesnikov, Dirk Weissenborn, Xiaohua Zhai, Thomas Unterthiner, Mostafa Dehghani, Matthias Minderer, Georg Heigold, Sylvain Gelly, et~al. 2020.
\newblock An image is worth 16x16 words: Transformers for image recognition at scale.
\newblock \emph{arXiv preprint arXiv:2010.11929}.

\bibitem[{Gao et~al.(2023)Gao, Tow, Abbasi, Biderman, Black, DiPofi, Foster, Golding, Hsu, Le~Noac'h, Li, McDonell, Muennighoff, Ociepa, Phang, Reynolds, Schoelkopf, Skowron, Sutawika, Tang, Thite, Wang, Wang, and Zou}]{eval-harness}
Leo Gao, Jonathan Tow, Baber Abbasi, Stella Biderman, Sid Black, Anthony DiPofi, Charles Foster, Laurence Golding, Jeffrey Hsu, Alain Le~Noac'h, Haonan Li, Kyle McDonell, Niklas Muennighoff, Chris Ociepa, Jason Phang, Laria Reynolds, Hailey Schoelkopf, Aviya Skowron, Lintang Sutawika, Eric Tang, Anish Thite, Ben Wang, Kevin Wang, and Andy Zou. 2023.
\newblock \href {https://doi.org/10.5281/zenodo.10256836} {A framework for few-shot language model evaluation}.

\bibitem[{Hendrycks et~al.(2021)Hendrycks, Burns, Basart, Zou, Mazeika, Song, and Steinhardt}]{hendryckstest2021}
Dan Hendrycks, Collin Burns, Steven Basart, Andy Zou, Mantas Mazeika, Dawn Song, and Jacob Steinhardt. 2021.
\newblock Measuring massive multitask language understanding.
\newblock \emph{Proceedings of the International Conference on Learning Representations (ICLR)}.

\bibitem[{Huang et~al.(2024)Huang, Li, Yang, Shi, Chang, Ye, Wu, Hong, Huang, Liu et~al.}]{huang2024audiogpt}
Rongjie Huang, Mingze Li, Dongchao Yang, Jiatong Shi, Xuankai Chang, Zhenhui Ye, Yuning Wu, Zhiqing Hong, Jiawei Huang, Jinglin Liu, et~al. 2024.
\newblock Audiogpt: Understanding and generating speech, music, sound, and talking head.
\newblock In \emph{Proceedings of the AAAI Conference on Artificial Intelligence}, volume~38, pages 23802--23804.

\bibitem[{Jang et~al.(2023)Jang, Kim, Lin, Wang, Hessel, Zettlemoyer, Hajishirzi, Choi, and Ammanabrolu}]{jang2023personalized}
Joel Jang, Seungone Kim, Bill~Yuchen Lin, Yizhong Wang, Jack Hessel, Luke Zettlemoyer, Hannaneh Hajishirzi, Yejin Choi, and Prithviraj Ammanabrolu. 2023.
\newblock Personalized soups: Personalized large language model alignment via post-hoc parameter merging.
\newblock \emph{arXiv preprint arXiv:2310.11564}.

\bibitem[{Jin et~al.(2023)Jin, Xu, Chen, Liao, Tan, Chen, Lei, Liu, Song, Lei et~al.}]{jin2023unified}
Yang Jin, Kun Xu, Liwei Chen, Chao Liao, Jianchao Tan, Bin Chen, Chenyi Lei, An~Liu, Chengru Song, Xiaoqiang Lei, et~al. 2023.
\newblock Unified language-vision pretraining with dynamic discrete visual tokenization.
\newblock \emph{arXiv preprint arXiv:2309.04669}.

\bibitem[{Koh et~al.(2024)Koh, Fried, and Salakhutdinov}]{koh2024generating}
Jing~Yu Koh, Daniel Fried, and Russ~R Salakhutdinov. 2024.
\newblock Generating images with multimodal language models.
\newblock \emph{Advances in Neural Information Processing Systems}, 36.

\bibitem[{Li et~al.(2024)Li, Wong, Zhang, Usuyama, Liu, Yang, Naumann, Poon, and Gao}]{li2024llava}
Chunyuan Li, Cliff Wong, Sheng Zhang, Naoto Usuyama, Haotian Liu, Jianwei Yang, Tristan Naumann, Hoifung Poon, and Jianfeng Gao. 2024.
\newblock Llava-med: Training a large language-and-vision assistant for biomedicine in one day.
\newblock \emph{Advances in Neural Information Processing Systems}, 36.

\bibitem[{Li et~al.(2023)Li, Du, Zhou, Wang, Zhao, and Wen}]{li2023evaluating}
Yifan Li, Yifan Du, Kun Zhou, Jinpeng Wang, Wayne~Xin Zhao, and Ji-Rong Wen. 2023.
\newblock Evaluating object hallucination in large vision-language models.
\newblock \emph{arXiv preprint arXiv:2305.10355}.

\bibitem[{Lin et~al.(2023)Lin, Zhu, Ye, Ning, Jin, and Yuan}]{lin2023video}
Bin Lin, Bin Zhu, Yang Ye, Munan Ning, Peng Jin, and Li~Yuan. 2023.
\newblock Video-llava: Learning united visual representation by alignment before projection.
\newblock \emph{arXiv preprint arXiv:2311.10122}.

\bibitem[{Liu et~al.(2023{\natexlab{a}})Liu, Li, Li, and Lee}]{liu2023improvedllava}
Haotian Liu, Chunyuan Li, Yuheng Li, and Yong~Jae Lee. 2023{\natexlab{a}}.
\newblock Improved baselines with visual instruction tuning.

\bibitem[{Liu et~al.(2024{\natexlab{a}})Liu, Li, Li, and Lee}]{liu2024improved}
Haotian Liu, Chunyuan Li, Yuheng Li, and Yong~Jae Lee. 2024{\natexlab{a}}.
\newblock Improved baselines with visual instruction tuning.
\newblock In \emph{Proceedings of the IEEE/CVF Conference on Computer Vision and Pattern Recognition}, pages 26296--26306.

\bibitem[{Liu et~al.(2024{\natexlab{b}})Liu, Li, Wu, and Lee}]{liu2024visual}
Haotian Liu, Chunyuan Li, Qingyang Wu, and Yong~Jae Lee. 2024{\natexlab{b}}.
\newblock Visual instruction tuning.
\newblock \emph{Advances in neural information processing systems}, 36.

\bibitem[{Liu et~al.(2019)Liu, Ott, Goyal, Du, Joshi, Chen, Levy, Lewis, Zettlemoyer, and Stoyanov}]{liu2019roberta}
Yinhan Liu, Myle Ott, Naman Goyal, Jingfei Du, Mandar Joshi, Danqi Chen, Omer Levy, Mike Lewis, Luke Zettlemoyer, and Veselin Stoyanov. 2019.
\newblock Roberta: A robustly optimized bert pretraining approach.
\newblock \emph{arXiv preprint arXiv:1907.11692}.

\bibitem[{Liu et~al.(2023{\natexlab{b}})Liu, Duan, Zhang, Li, Zhang, Zhao, Yuan, Wang, He, Liu et~al.}]{liu2023mmbench}
Yuan Liu, Haodong Duan, Yuanhan Zhang, Bo~Li, Songyang Zhang, Wangbo Zhao, Yike Yuan, Jiaqi Wang, Conghui He, Ziwei Liu, et~al. 2023{\natexlab{b}}.
\newblock Mmbench: Is your multi-modal model an all-around player?
\newblock \emph{arXiv preprint arXiv:2307.06281}.

\bibitem[{Lu et~al.(2023)Lu, Clark, Lee, Zhang, Khosla, Marten, Hoiem, and Kembhavi}]{lu2023unified}
Jiasen Lu, Christopher Clark, Sangho Lee, Zichen Zhang, Savya Khosla, Ryan Marten, Derek Hoiem, and Aniruddha Kembhavi. 2023.
\newblock Unified-io 2: Scaling autoregressive multimodal models with vision, language, audio, and action.
\newblock \emph{arXiv preprint arXiv:2312.17172}.

\bibitem[{Lu et~al.(2022)Lu, Clark, Zellers, Mottaghi, and Kembhavi}]{lu2022unified}
Jiasen Lu, Christopher Clark, Rowan Zellers, Roozbeh Mottaghi, and Aniruddha Kembhavi. 2022.
\newblock Unified-io: A unified model for vision, language, and multi-modal tasks.
\newblock In \emph{The Eleventh International Conference on Learning Representations}.

\bibitem[{Luo et~al.(2020)Luo, Ji, Shi, Huang, Duan, Li, Li, Bharti, and Zhou}]{luo2020univl}
Huaishao Luo, Lei Ji, Botian Shi, Haoyang Huang, Nan Duan, Tianrui Li, Jason Li, Taroon Bharti, and Ming Zhou. 2020.
\newblock Univl: A unified video and language pre-training model for multimodal understanding and generation.
\newblock \emph{arXiv preprint arXiv:2002.06353}.

\bibitem[{Narayanan et~al.(2021)Narayanan, Shoeybi, Casper, LeGresley, Patwary, Korthikanti, Vainbrand, Kashinkunti, Bernauer, Catanzaro et~al.}]{narayanan2021efficient}
Deepak Narayanan, Mohammad Shoeybi, Jared Casper, Patrick LeGresley, Mostofa Patwary, Vijay Korthikanti, Dmitri Vainbrand, Prethvi Kashinkunti, Julie Bernauer, Bryan Catanzaro, et~al. 2021.
\newblock Efficient large-scale language model training on gpu clusters using megatron-lm.
\newblock In \emph{Proceedings of the International Conference for High Performance Computing, Networking, Storage and Analysis}, pages 1--15.

\bibitem[{Nowlan and Hinton(2018)}]{nowlan2018simplifying}
Steven~J Nowlan and Geoffrey~E Hinton. 2018.
\newblock Simplifying neural networks by soft weight sharing.
\newblock In \emph{The Mathematics of Generalization}, pages 373--394. CRC Press.

\bibitem[{Ouyang et~al.(2022)Ouyang, Wu, Jiang, Almeida, Wainwright, Mishkin, Zhang, Agarwal, Slama, Ray et~al.}]{ouyang2022training}
Long Ouyang, Jeffrey Wu, Xu~Jiang, Diogo Almeida, Carroll Wainwright, Pamela Mishkin, Chong Zhang, Sandhini Agarwal, Katarina Slama, Alex Ray, et~al. 2022.
\newblock Training language models to follow instructions with human feedback.
\newblock \emph{Advances in neural information processing systems}, 35:27730--27744.

\bibitem[{Pan et~al.(2023)Pan, Dong, Huang, Peng, Chen, and Wei}]{pan2023kosmos}
Xichen Pan, Li~Dong, Shaohan Huang, Zhiliang Peng, Wenhu Chen, and Furu Wei. 2023.
\newblock Kosmos-g: Generating images in context with multimodal large language models.
\newblock \emph{arXiv preprint arXiv:2310.02992}.

\bibitem[{Radford et~al.(2021)Radford, Kim, Hallacy, Ramesh, Goh, Agarwal, Sastry, Askell, Mishkin, Clark et~al.}]{radford2021learning}
Alec Radford, Jong~Wook Kim, Chris Hallacy, Aditya Ramesh, Gabriel Goh, Sandhini Agarwal, Girish Sastry, Amanda Askell, Pamela Mishkin, Jack Clark, et~al. 2021.
\newblock Learning transferable visual models from natural language supervision.
\newblock In \emph{International conference on machine learning}, pages 8748--8763. PMLR.

\bibitem[{Radford et~al.(2018)Radford, Narasimhan, Salimans, Sutskever et~al.}]{radford2018improving}
Alec Radford, Karthik Narasimhan, Tim Salimans, Ilya Sutskever, et~al. 2018.
\newblock Improving language understanding by generative pre-training.

\bibitem[{Radford et~al.(2019)Radford, Wu, Child, Luan, Amodei, Sutskever et~al.}]{radford2019language}
Alec Radford, Jeffrey Wu, Rewon Child, David Luan, Dario Amodei, Ilya Sutskever, et~al. 2019.
\newblock Language models are unsupervised multitask learners.
\newblock \emph{OpenAI blog}, 1(8):9.

\bibitem[{Reid et~al.(2024)Reid, Savinov, Teplyashin, Lepikhin, Lillicrap, Alayrac, Soricut, Lazaridou, Firat, Schrittwieser et~al.}]{reid2024gemini}
Machel Reid, Nikolay Savinov, Denis Teplyashin, Dmitry Lepikhin, Timothy Lillicrap, Jean-baptiste Alayrac, Radu Soricut, Angeliki Lazaridou, Orhan Firat, Julian Schrittwieser, et~al. 2024.
\newblock Gemini 1.5: Unlocking multimodal understanding across millions of tokens of context.
\newblock \emph{arXiv preprint arXiv:2403.05530}.

\bibitem[{Shazeer et~al.(2017)Shazeer, Mirhoseini, Maziarz, Davis, Le, Hinton, and Dean}]{shazeer2017outrageously}
Noam Shazeer, Azalia Mirhoseini, Krzysztof Maziarz, Andy Davis, Quoc Le, Geoffrey Hinton, and Jeff Dean. 2017.
\newblock Outrageously large neural networks: The sparsely-gated mixture-of-experts layer.
\newblock \emph{arXiv preprint arXiv:1701.06538}.

\bibitem[{Stiennon et~al.(2020)Stiennon, Ouyang, Wu, Ziegler, Lowe, Voss, Radford, Amodei, and Christiano}]{stiennon2020learning}
Nisan Stiennon, Long Ouyang, Jeffrey Wu, Daniel Ziegler, Ryan Lowe, Chelsea Voss, Alec Radford, Dario Amodei, and Paul~F Christiano. 2020.
\newblock Learning to summarize with human feedback.
\newblock \emph{Advances in Neural Information Processing Systems}, 33:3008--3021.

\bibitem[{Sukhbaatar et~al.(2024)Sukhbaatar, Golovneva, Sharma, Xu, Lin, Rozi{\`e}re, Kahn, Li, Yih, Weston et~al.}]{sukhbaatar2024branch}
Sainbayar Sukhbaatar, Olga Golovneva, Vasu Sharma, Hu~Xu, Xi~Victoria Lin, Baptiste Rozi{\`e}re, Jacob Kahn, Daniel Li, Wen-tau Yih, Jason Weston, et~al. 2024.
\newblock Branch-train-mix: Mixing expert llms into a mixture-of-experts llm.
\newblock \emph{arXiv preprint arXiv:2403.07816}.

\bibitem[{Suzgun et~al.(2022)Suzgun, Scales, Sch{\"a}rli, Gehrmann, Tay, Chung, Chowdhery, Le, Chi, Zhou et~al.}]{suzgun2022challenging}
Mirac Suzgun, Nathan Scales, Nathanael Sch{\"a}rli, Sebastian Gehrmann, Yi~Tay, Hyung~Won Chung, Aakanksha Chowdhery, Quoc~V Le, Ed~H Chi, Denny Zhou, et~al. 2022.
\newblock Challenging big-bench tasks and whether chain-of-thought can solve them.
\newblock \emph{arXiv preprint arXiv:2210.09261}.

\bibitem[{Swayamdipta et~al.(2020)Swayamdipta, Schwartz, Lourie, Wang, Hajishirzi, Smith, and Choi}]{swayamdipta2020dataset}
Swabha Swayamdipta, Roy Schwartz, Nicholas Lourie, Yizhong Wang, Hannaneh Hajishirzi, Noah~A Smith, and Yejin Choi. 2020.
\newblock Dataset cartography: Mapping and diagnosing datasets with training dynamics.
\newblock \emph{arXiv preprint arXiv:2009.10795}.

\bibitem[{Tang et~al.(2024)Tang, Yang, Zhu, Zeng, and Bansal}]{tang2024any}
Zineng Tang, Ziyi Yang, Chenguang Zhu, Michael Zeng, and Mohit Bansal. 2024.
\newblock Any-to-any generation via composable diffusion.
\newblock \emph{Advances in Neural Information Processing Systems}, 36.

\bibitem[{Taori et~al.(2023)Taori, Gulrajani, Zhang, Dubois, Li, Guestrin, Liang, and Hashimoto}]{taori2023alpaca}
Rohan Taori, Ishaan Gulrajani, Tianyi Zhang, Yann Dubois, Xuechen Li, Carlos Guestrin, Percy Liang, and Tatsunori~B Hashimoto. 2023.
\newblock Alpaca: A strong, replicable instruction-following model.
\newblock \emph{Stanford Center for Research on Foundation Models. https://crfm. stanford. edu/2023/03/13/alpaca. html}, 3(6):7.

\bibitem[{Thrush et~al.(2022)Thrush, Jiang, Bartolo, Singh, Williams, Kiela, and Ross}]{thrush2022winoground}
Tristan Thrush, Ryan Jiang, Max Bartolo, Amanpreet Singh, Adina Williams, Douwe Kiela, and Candace Ross. 2022.
\newblock Winoground: Probing vision and language models for visio-linguistic compositionality.
\newblock In \emph{Proceedings of the IEEE/CVF Conference on Computer Vision and Pattern Recognition}, pages 5238--5248.

\bibitem[{Touvron et~al.(2023)Touvron, Lavril, Izacard, Martinet, Lachaux, Lacroix, Rozi{\`e}re, Goyal, Hambro, Azhar et~al.}]{touvron2023llama}
Hugo Touvron, Thibaut Lavril, Gautier Izacard, Xavier Martinet, Marie-Anne Lachaux, Timoth{\'e}e Lacroix, Baptiste Rozi{\`e}re, Naman Goyal, Eric Hambro, Faisal Azhar, et~al. 2023.
\newblock Llama: Open and efficient foundation language models.
\newblock \emph{arXiv preprint arXiv:2302.13971}.

\bibitem[{Vaswani et~al.(2017)Vaswani, Shazeer, Parmar, Uszkoreit, Jones, Gomez, Kaiser, and Polosukhin}]{vaswani2017attention}
Ashish Vaswani, Noam Shazeer, Niki Parmar, Jakob Uszkoreit, Llion Jones, Aidan~N Gomez, {\L}ukasz Kaiser, and Illia Polosukhin. 2017.
\newblock Attention is all you need.
\newblock \emph{Advances in neural information processing systems}, 30.

\bibitem[{Wang et~al.(2022)Wang, Kordi, Mishra, Liu, Smith, Khashabi, and Hajishirzi}]{wang2022self}
Yizhong Wang, Yeganeh Kordi, Swaroop Mishra, Alisa Liu, Noah~A Smith, Daniel Khashabi, and Hannaneh Hajishirzi. 2022.
\newblock Self-instruct: Aligning language models with self-generated instructions.
\newblock \emph{arXiv preprint arXiv:2212.10560}.

\bibitem[{Wolf et~al.(2020)Wolf, Debut, Sanh, Chaumond, Delangue, Moi, Cistac, Rault, Louf, Funtowicz, Davison, Shleifer, von Platen, Ma, Jernite, Plu, Xu, Scao, Gugger, Drame, Lhoest, and Rush}]{wolf-etal-2020-transformers}
Thomas Wolf, Lysandre Debut, Victor Sanh, Julien Chaumond, Clement Delangue, Anthony Moi, Pierric Cistac, Tim Rault, Rémi Louf, Morgan Funtowicz, Joe Davison, Sam Shleifer, Patrick von Platen, Clara Ma, Yacine Jernite, Julien Plu, Canwen Xu, Teven~Le Scao, Sylvain Gugger, Mariama Drame, Quentin Lhoest, and Alexander~M. Rush. 2020.
\newblock \href {https://www.aclweb.org/anthology/2020.emnlp-demos.6} {Transformers: State-of-the-art natural language processing}.
\newblock In \emph{Proceedings of the 2020 Conference on Empirical Methods in Natural Language Processing: System Demonstrations}, pages 38--45, Online. Association for Computational Linguistics.

\bibitem[{Wortsman et~al.(2022)Wortsman, Ilharco, Gadre, Roelofs, Gontijo-Lopes, Morcos, Namkoong, Farhadi, Carmon, Kornblith et~al.}]{wortsman2022model}
Mitchell Wortsman, Gabriel Ilharco, Samir~Ya Gadre, Rebecca Roelofs, Raphael Gontijo-Lopes, Ari~S Morcos, Hongseok Namkoong, Ali Farhadi, Yair Carmon, Simon Kornblith, et~al. 2022.
\newblock Model soups: averaging weights of multiple fine-tuned models improves accuracy without increasing inference time.
\newblock In \emph{International conference on machine learning}, pages 23965--23998. PMLR.

\bibitem[{Xie et~al.(2024)Xie, Pham, Dong, Du, Liu, Lu, Liang, Le, Ma, and Yu}]{xie2024doremi}
Sang~Michael Xie, Hieu Pham, Xuanyi Dong, Nan Du, Hanxiao Liu, Yifeng Lu, Percy~S Liang, Quoc~V Le, Tengyu Ma, and Adams~Wei Yu. 2024.
\newblock Doremi: Optimizing data mixtures speeds up language model pretraining.
\newblock \emph{Advances in Neural Information Processing Systems}, 36.

\bibitem[{Xu et~al.(2023)Xu, Wang, Wang, Chen, Pang, and Lin}]{xu2023pointllm}
Runsen Xu, Xiaolong Wang, Tai Wang, Yilun Chen, Jiangmiao Pang, and Dahua Lin. 2023.
\newblock Pointllm: Empowering large language models to understand point clouds.
\newblock \emph{arXiv preprint arXiv:2308.16911}.

\bibitem[{Yan et~al.(2021)Yan, Zhang, Abbeel, and Srinivas}]{yan2021videogpt}
Wilson Yan, Yunzhi Zhang, Pieter Abbeel, and Aravind Srinivas. 2021.
\newblock Videogpt: Video generation using vq-vae and transformers.
\newblock \emph{arXiv preprint arXiv:2104.10157}.

\bibitem[{Yue et~al.(2023)Yue, Ni, Zhang, Zheng, Liu, Zhang, Stevens, Jiang, Ren, Sun et~al.}]{yue2023mmmu}
Xiang Yue, Yuansheng Ni, Kai Zhang, Tianyu Zheng, Ruoqi Liu, Ge~Zhang, Samuel Stevens, Dongfu Jiang, Weiming Ren, Yuxuan Sun, et~al. 2023.
\newblock Mmmu: A massive multi-discipline multimodal understanding and reasoning benchmark for expert agi.
\newblock \emph{arXiv preprint arXiv:2311.16502}.

\bibitem[{Zellers et~al.(2019)Zellers, Holtzman, Bisk, Farhadi, and Choi}]{zellers2019hellaswag}
Rowan Zellers, Ari Holtzman, Yonatan Bisk, Ali Farhadi, and Yejin Choi. 2019.
\newblock Hellaswag: Can a machine really finish your sentence?
\newblock In \emph{Proceedings of the 57th Annual Meeting of the Association for Computational Linguistics}.

\bibitem[{Zhang et~al.(2023)Zhang, Han, Liu, Gao, Zhou, Hu, Yan, Lu, Li, and Qiao}]{zhang2023llama}
Renrui Zhang, Jiaming Han, Chris Liu, Peng Gao, Aojun Zhou, Xiangfei Hu, Shilin Yan, Pan Lu, Hongsheng Li, and Yu~Qiao. 2023.
\newblock Llama-adapter: Efficient fine-tuning of language models with zero-init attention.
\newblock \emph{arXiv preprint arXiv:2303.16199}.

\bibitem[{Zheng et~al.(2024)Zheng, Chiang, Sheng, Zhuang, Wu, Zhuang, Lin, Li, Li, Xing et~al.}]{zheng2024judging}
Lianmin Zheng, Wei-Lin Chiang, Ying Sheng, Siyuan Zhuang, Zhanghao Wu, Yonghao Zhuang, Zi~Lin, Zhuohan Li, Dacheng Li, Eric Xing, et~al. 2024.
\newblock Judging llm-as-a-judge with mt-bench and chatbot arena.
\newblock \emph{Advances in Neural Information Processing Systems}, 36.

\bibitem[{Zheng et~al.(2023)Zheng, Chiang, Sheng, Zhuang, Wu, Zhuang, Lin, Li, Li, Xing, Zhang, Gonzalez, and Stoica}]{zheng2023judging}
Lianmin Zheng, Wei-Lin Chiang, Ying Sheng, Siyuan Zhuang, Zhanghao Wu, Yonghao Zhuang, Zi~Lin, Zhuohan Li, Dacheng Li, Eric.~P Xing, Hao Zhang, Joseph~E. Gonzalez, and Ion Stoica. 2023.
\newblock \href {https://arxiv.org/abs/2306.05685} {Judging llm-as-a-judge with mt-bench and chatbot arena}.
\newblock \emph{Preprint}, arXiv:2306.05685.

\bibitem[{Zhu et~al.(2023)Zhu, Chen, Shen, Li, and Elhoseiny}]{zhu2023minigpt}
Deyao Zhu, Jun Chen, Xiaoqian Shen, Xiang Li, and Mohamed Elhoseiny. 2023.
\newblock Minigpt-4: Enhancing vision-language understanding with advanced large language models.
\newblock \emph{arXiv preprint arXiv:2304.10592}.

\bibitem[{Ziegler et~al.(2019)Ziegler, Stiennon, Wu, Brown, Radford, Amodei, Christiano, and Irving}]{ziegler2019fine}
Daniel~M Ziegler, Nisan Stiennon, Jeffrey Wu, Tom~B Brown, Alec Radford, Dario Amodei, Paul Christiano, and Geoffrey Irving. 2019.
\newblock Fine-tuning language models from human preferences.
\newblock \emph{arXiv preprint arXiv:1909.08593}.

\end{thebibliography}

\clearpage

\appendix

\section{Supplementary Material}\label{sec:appendix}
\subsection{More Implementation Details}
Our experiments are conducted on A6000 GPUs.
We borrow the code of LLaVA and use its provided model checkpoints for Vicuna and LLaVA base models, and the LLaVA665K instruction dataset.
We directly use the training split of meta sets from Huggingface~\cite{wolf-etal-2020-transformers}.
For language evaluation, we leverage on the organized lm-evaluation-harness~\cite{eval-harness} codebase, and for vision-language tasks, we follow the evaluation instruction from LLaVA or use their official evaluation protocols.
Our exploration is mainly based on 7B model with their V1.5 version, but it can easily extended to larger model size and other versions of models.

\subsection{More Evaluation Performances}
Due to the limited space in the main draft, we provide more evaluation performances on language and vision-language domains (\textbf{More Evaluations} section in Sec.~\ref{sec:experiments}) in Tab.~\ref{tab:more_eval}

\subsection{Complete First Round Ablation Visualizations}
\begin{figure*}[htbp]
  \centering
  \begin{subfigure}[b]{0.19\textwidth}
    \includegraphics[width=\textwidth]{figs/1_set_ablation/mmmu_mmmu.pdf}
    \caption{MM-MM}
  \end{subfigure}
  \hfill 
  \begin{subfigure}[b]{0.19\textwidth}
    \includegraphics[width=\textwidth]{figs/1_set_ablation/mmmu_llavabench.pdf}
    \caption{MM-L}
  \end{subfigure}
  \hfill 
  \begin{subfigure}[b]{0.19\textwidth}
    \includegraphics[width=\textwidth]{figs/1_set_ablation/mmmu_mmlu.pdf}
    \caption{MM-ML}
  \end{subfigure}
  \hfill 
  \begin{subfigure}[b]{0.19\textwidth}
    \includegraphics[width=\textwidth]{figs/1_set_ablation/mmmu_gsm8k.pdf}
    \caption{MM-G}
  \end{subfigure}
  \hfill 
  \begin{subfigure}[b]{0.19\textwidth}
    \includegraphics[width=\textwidth]{figs/1_set_ablation/mmmu_hellaswag.pdf}
    \caption{MM-H}
  \end{subfigure}
  \hfill

  \begin{subfigure}[b]{0.19\textwidth}
    \includegraphics[width=\textwidth]{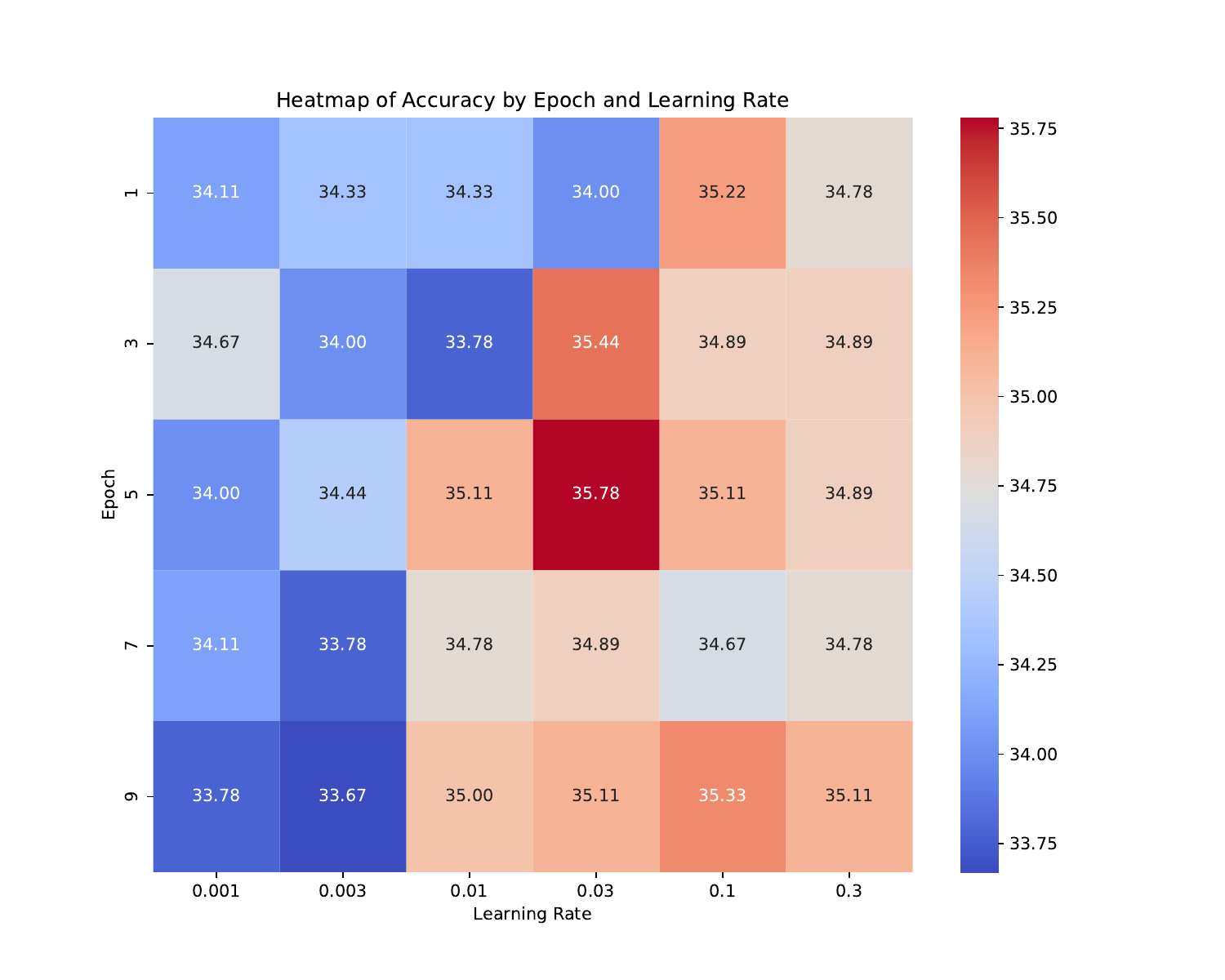}
    \caption{L-MM}
  \end{subfigure}
  \hfill 
  \begin{subfigure}[b]{0.19\textwidth}
    \includegraphics[width=\textwidth]{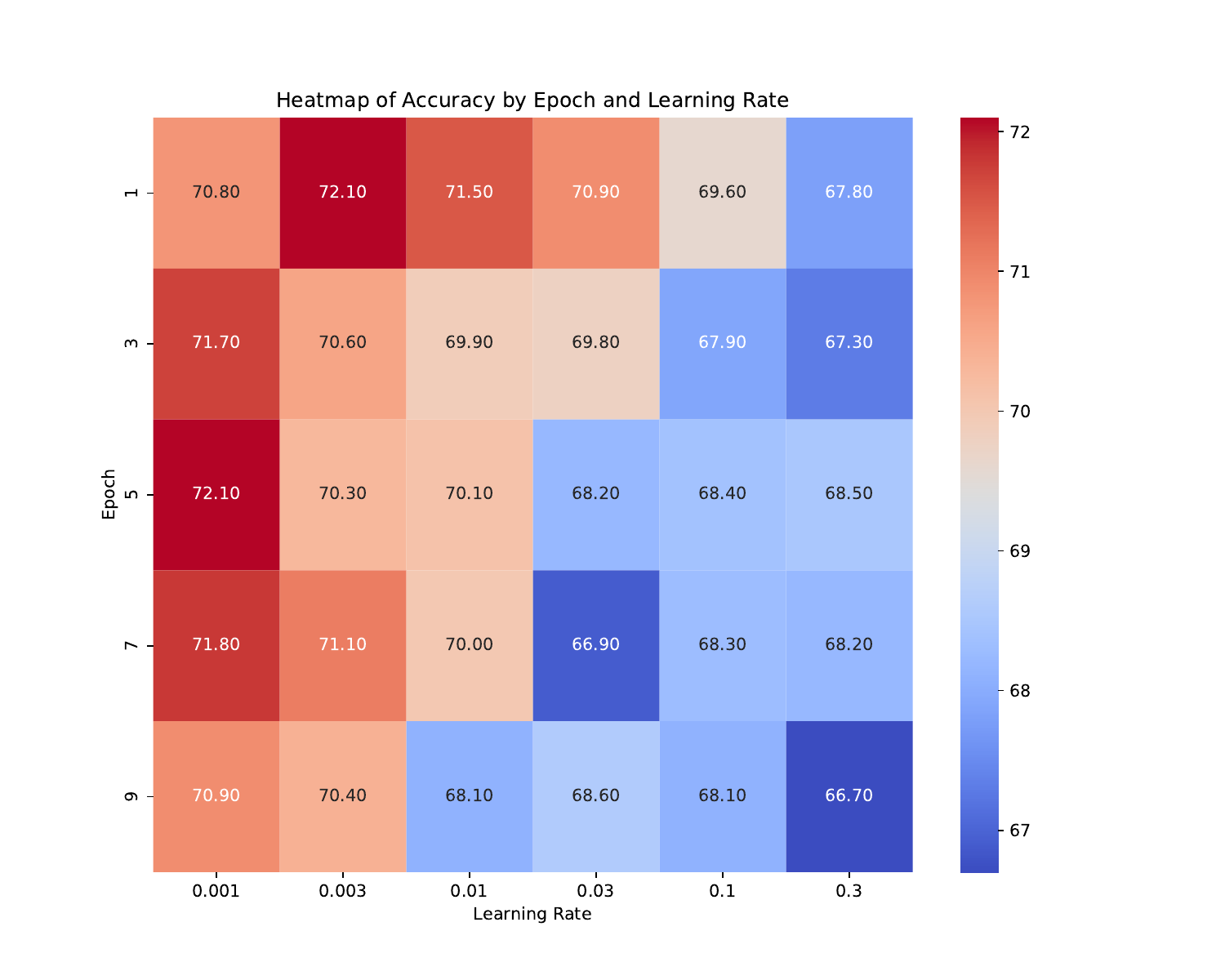}
    \caption{L-L}
  \end{subfigure}
  \hfill 
  \begin{subfigure}[b]{0.19\textwidth}
    \includegraphics[width=\textwidth]{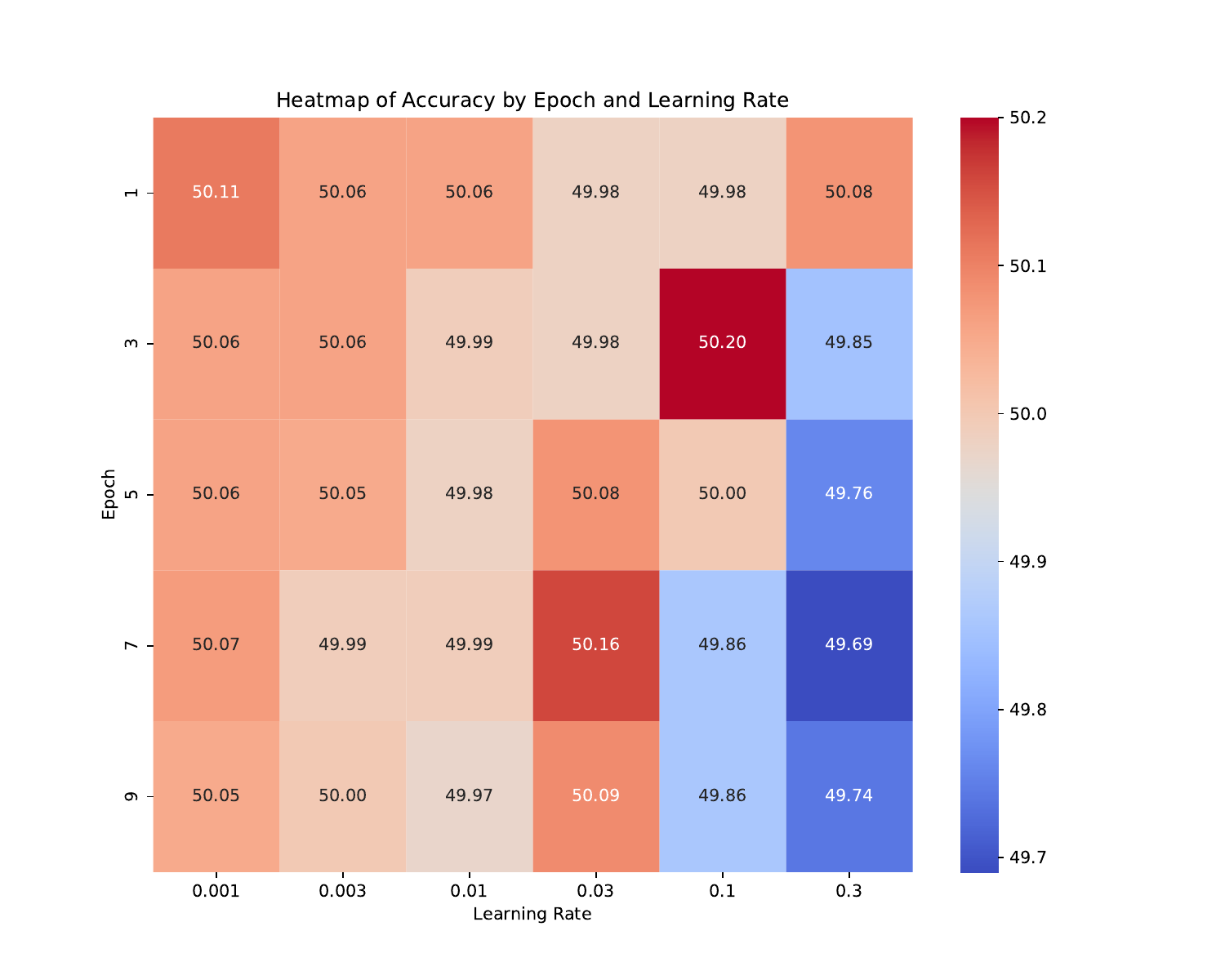}
    \caption{L-ML}
  \end{subfigure}
  \hfill 
  \begin{subfigure}[b]{0.19\textwidth}
    \includegraphics[width=\textwidth]{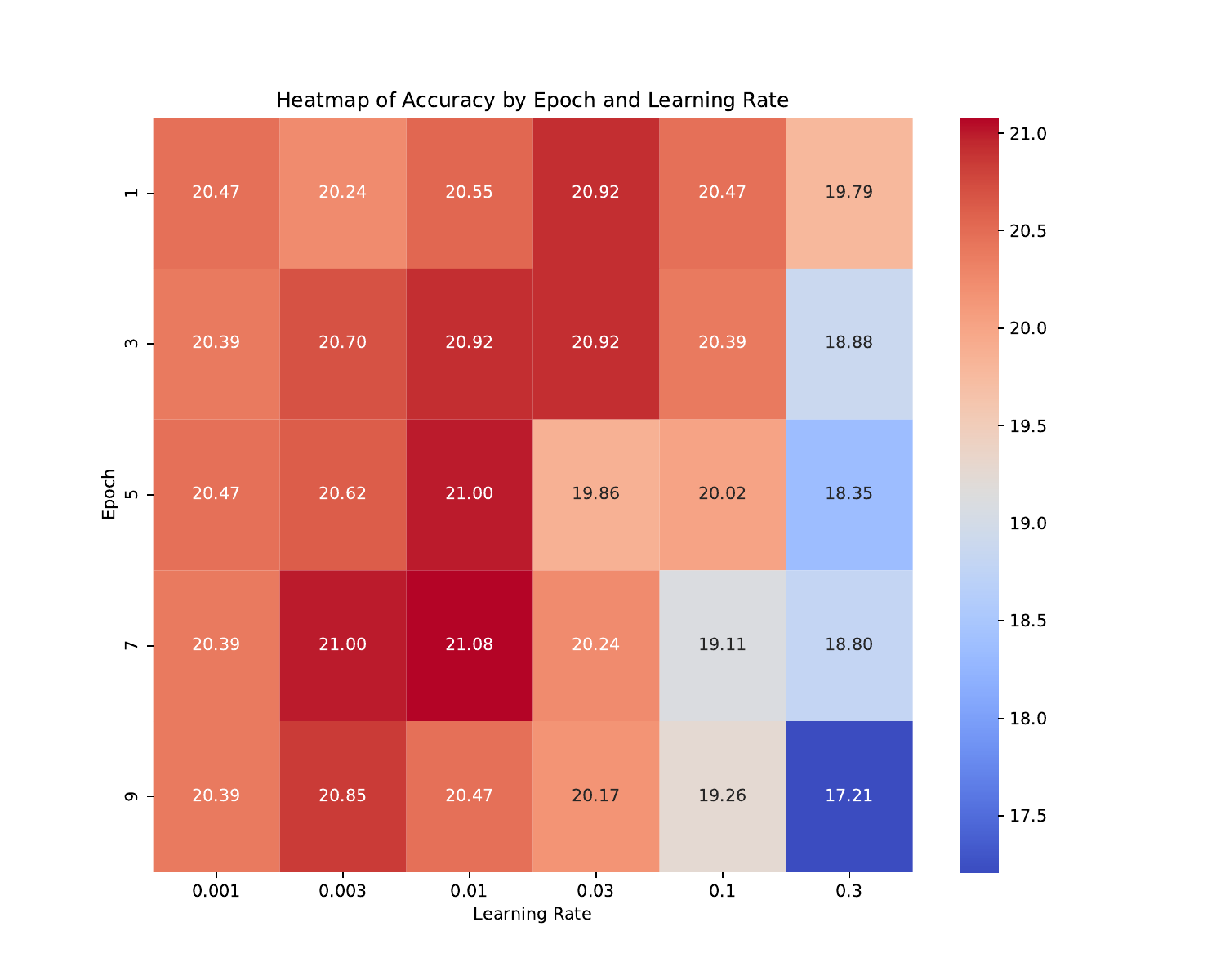}
    \caption{L-G}
  \end{subfigure}
  \hfill 
  \begin{subfigure}[b]{0.19\textwidth}
    \includegraphics[width=\textwidth]{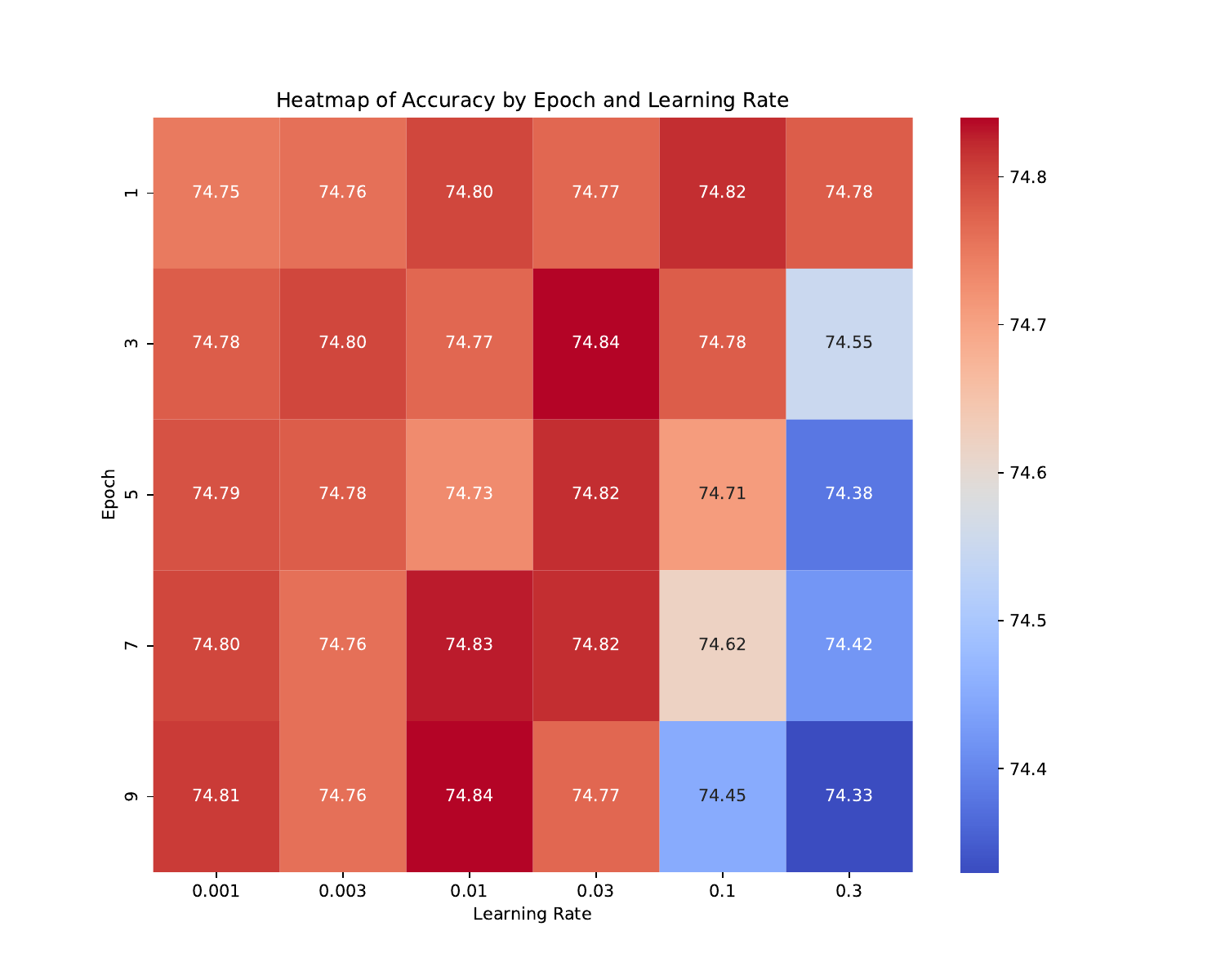}
    \caption{L-H}
  \end{subfigure}
  \hfill

  \begin{subfigure}[b]{0.19\textwidth}
    \includegraphics[width=\textwidth]{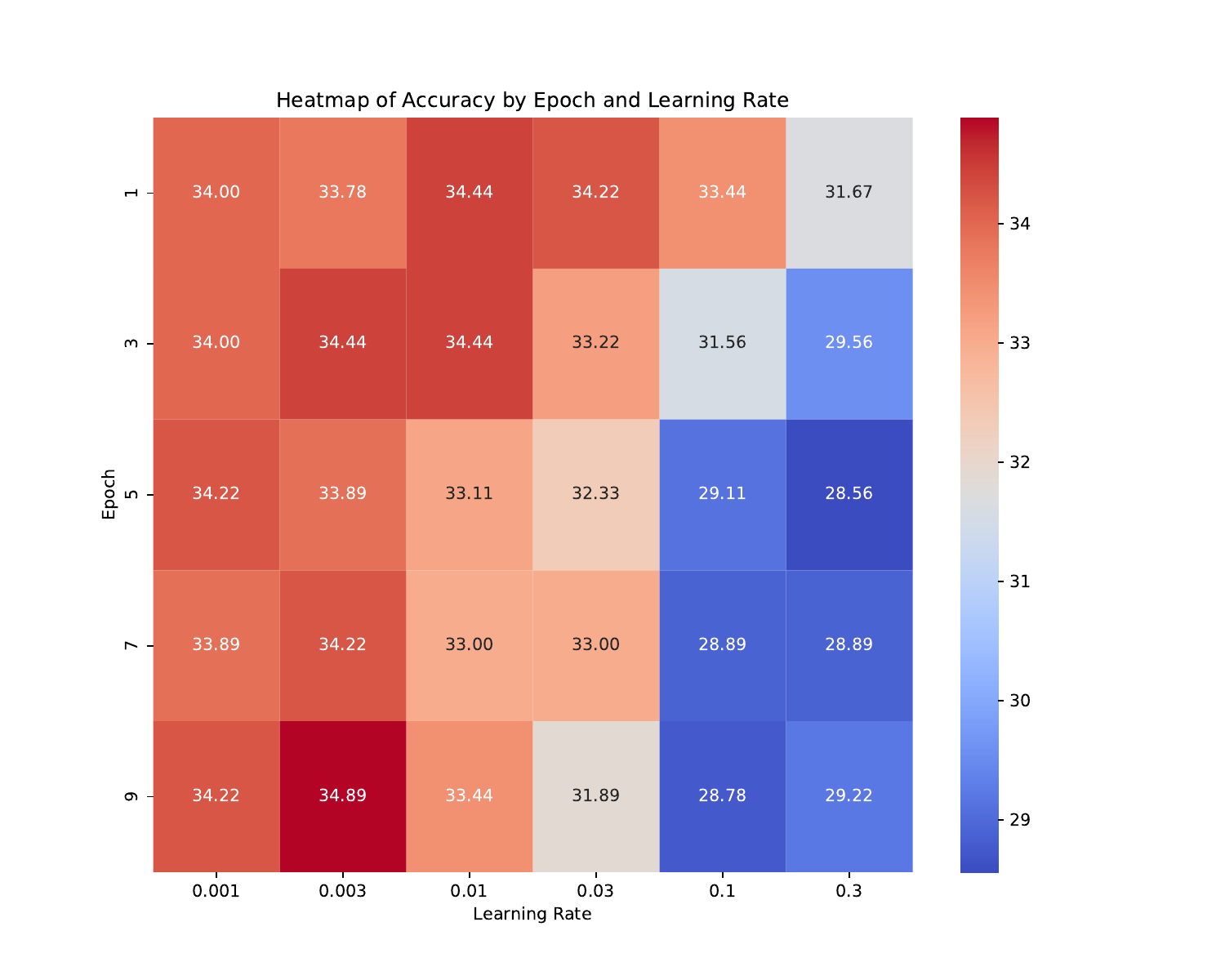}
    \caption{ML-MM}
  \end{subfigure}
  \hfill
  \begin{subfigure}[b]{0.19\textwidth}
    \includegraphics[width=\textwidth]{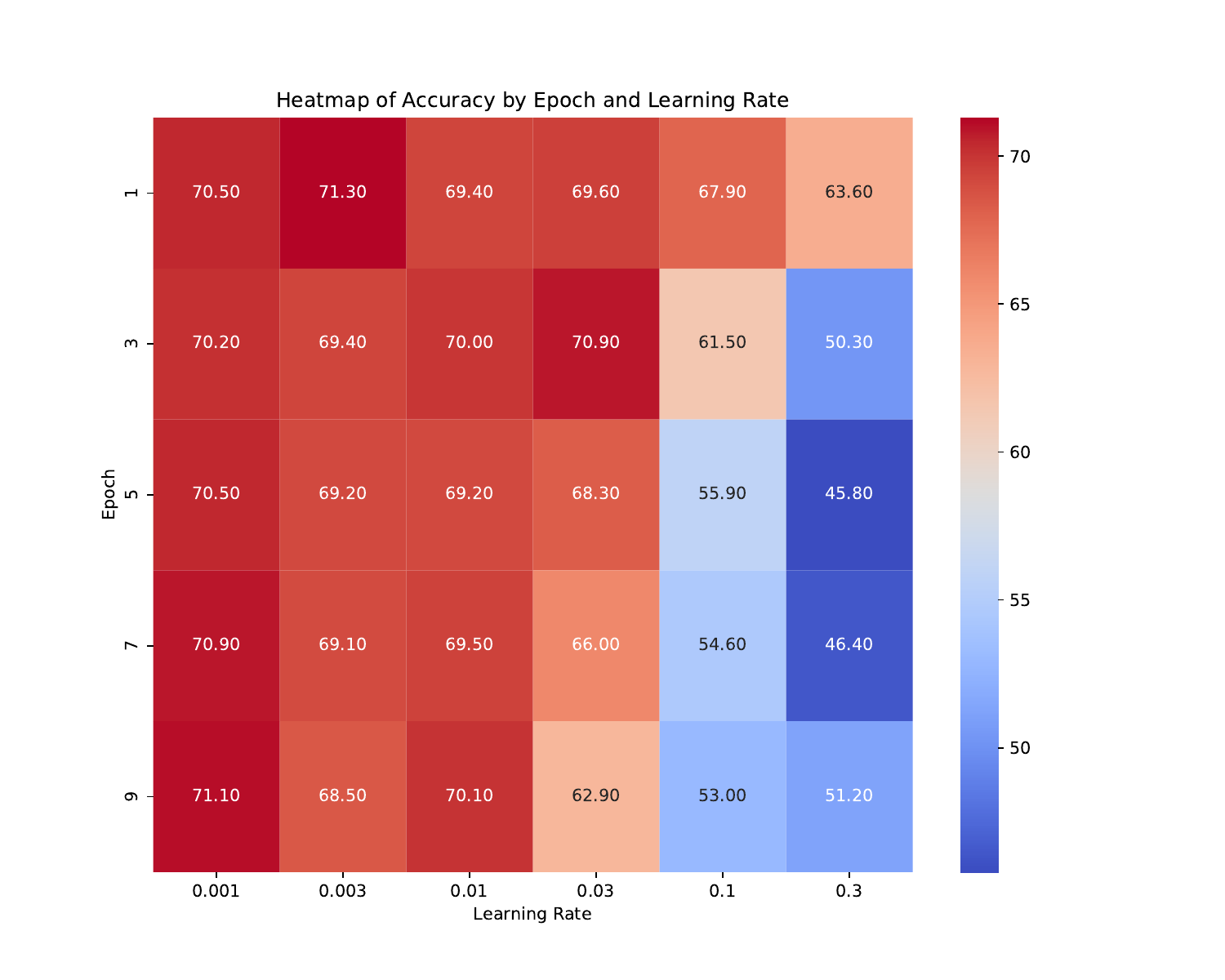}
    \caption{ML-L}
  \end{subfigure}
  \hfill
  \begin{subfigure}[b]{0.19\textwidth}
    \includegraphics[width=\textwidth]{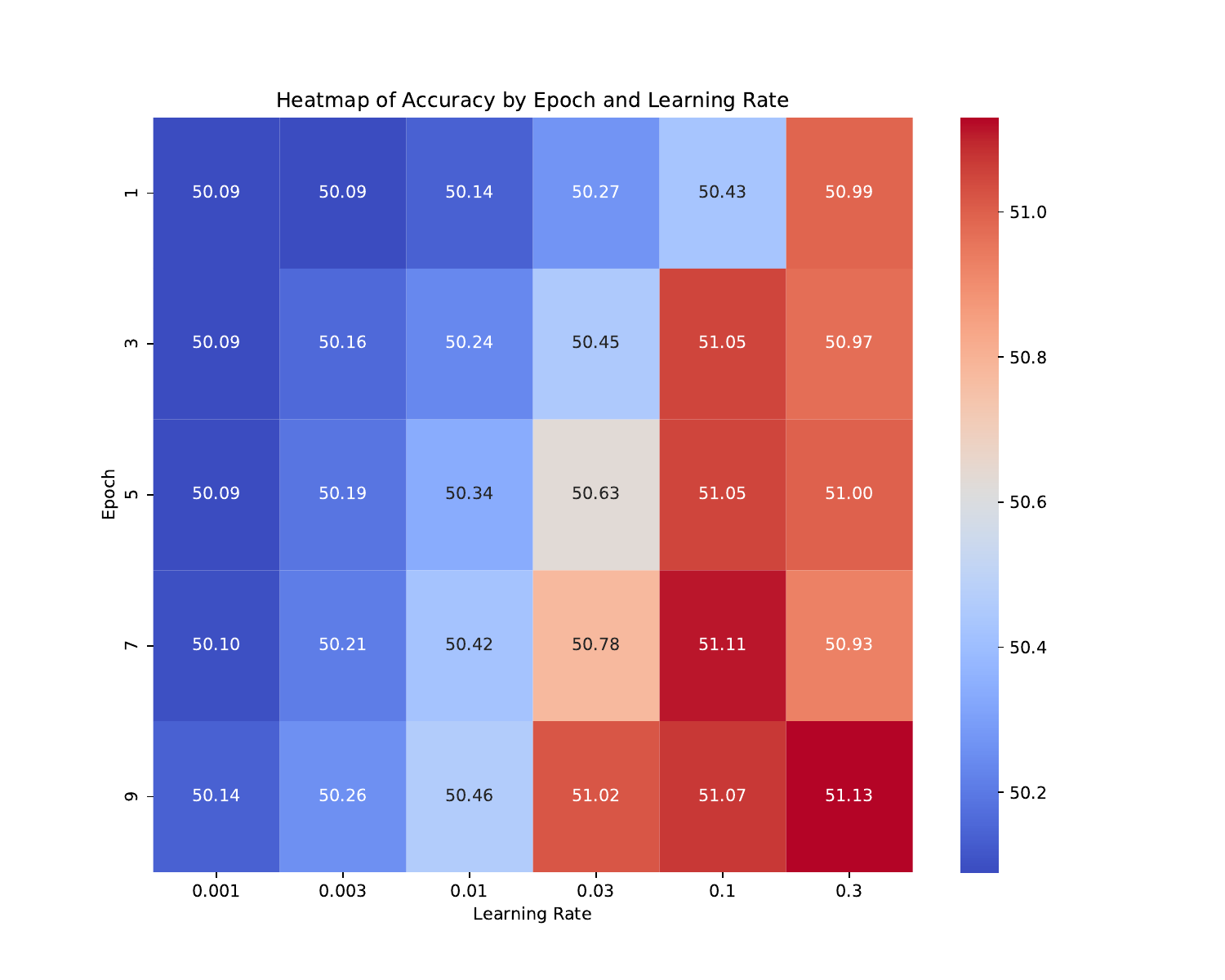}
    \caption{ML-ML}
  \end{subfigure}
  \hfill
  \begin{subfigure}[b]{0.19\textwidth}
    \includegraphics[width=\textwidth]{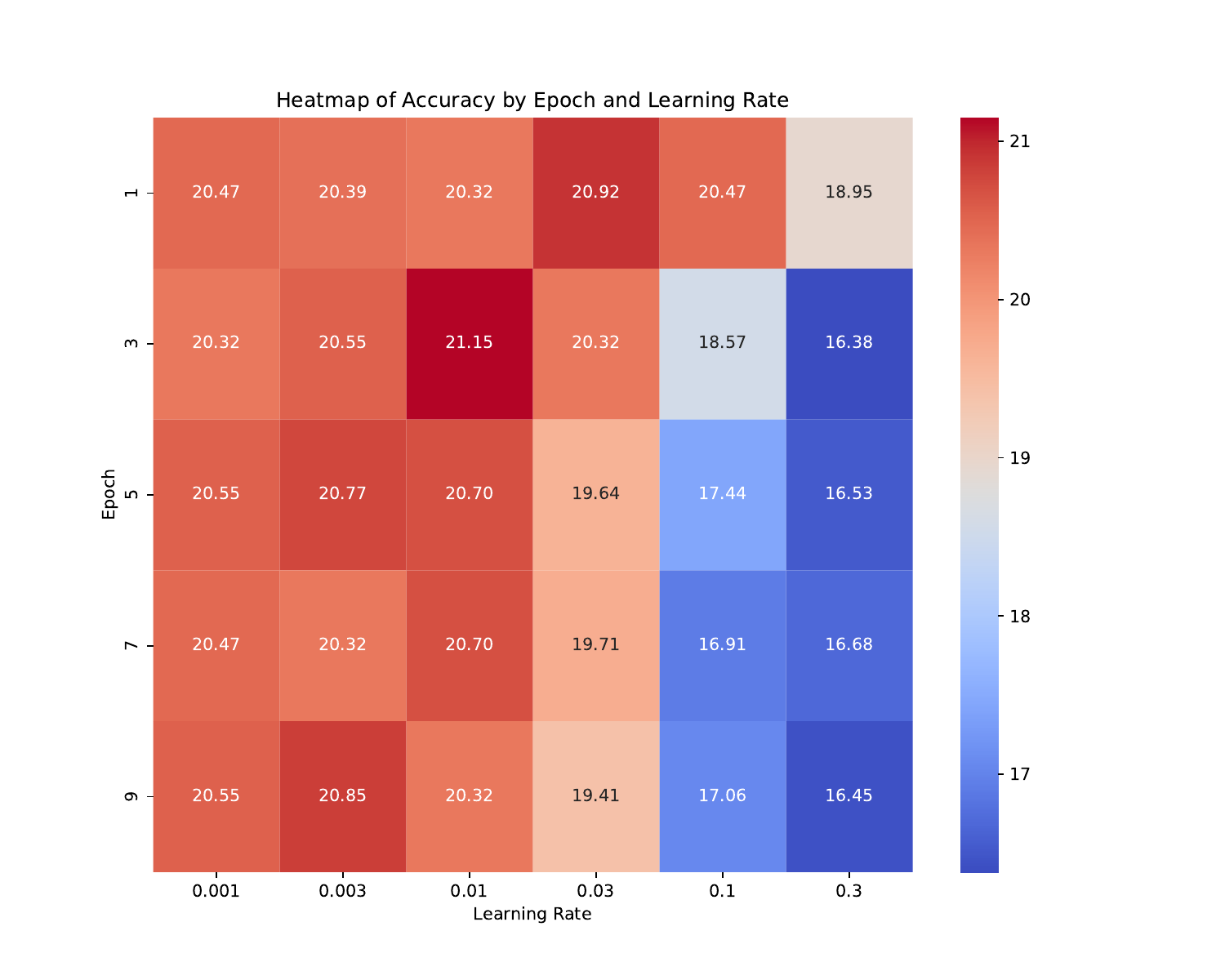}
    \caption{ML-G}
  \end{subfigure}
  \hfill
  \begin{subfigure}[b]{0.19\textwidth}
    \includegraphics[width=\textwidth]{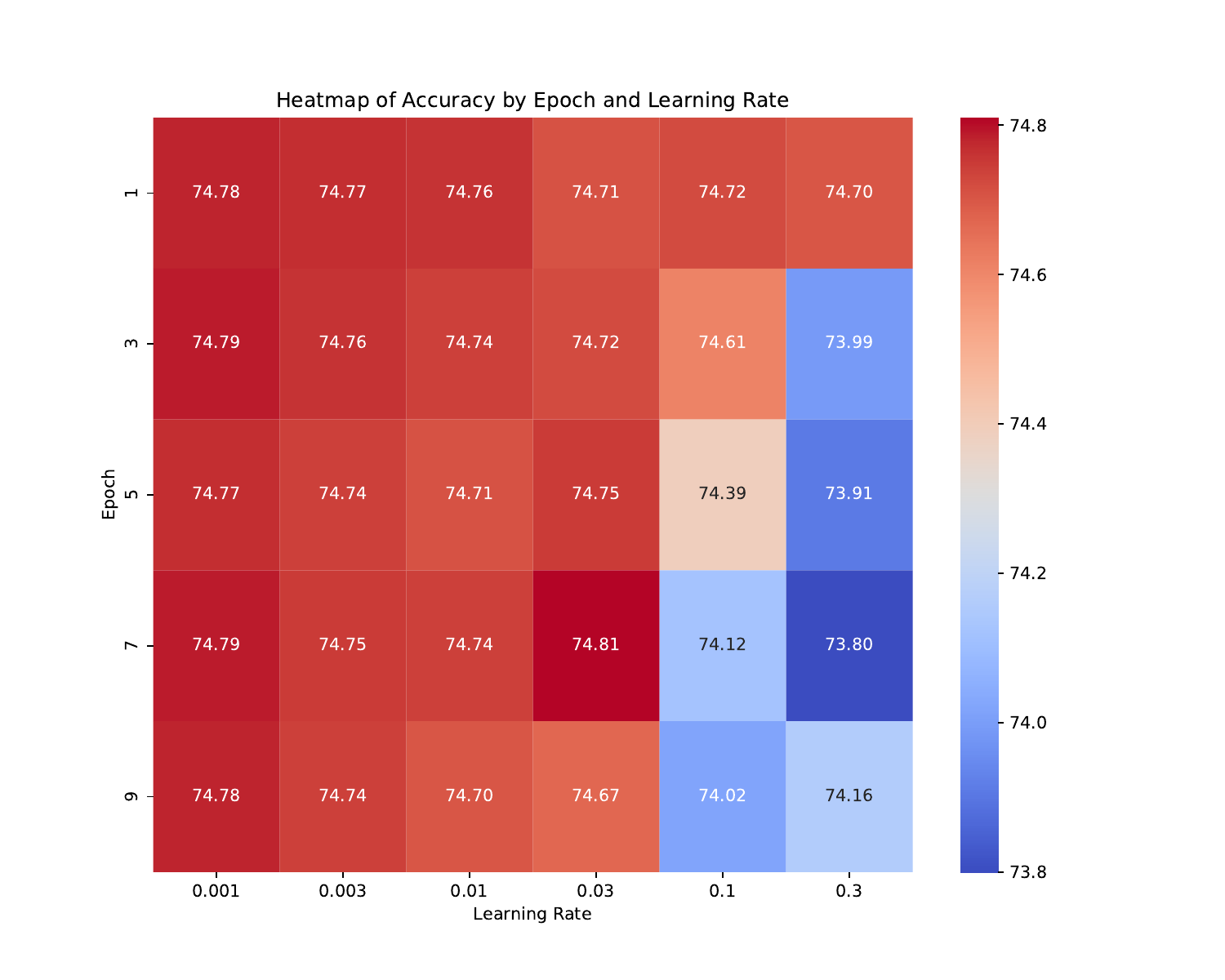}
    \caption{ML-H}
  \end{subfigure}
  \hfill

  \begin{subfigure}[b]{0.19\textwidth}
    \includegraphics[width=\textwidth]{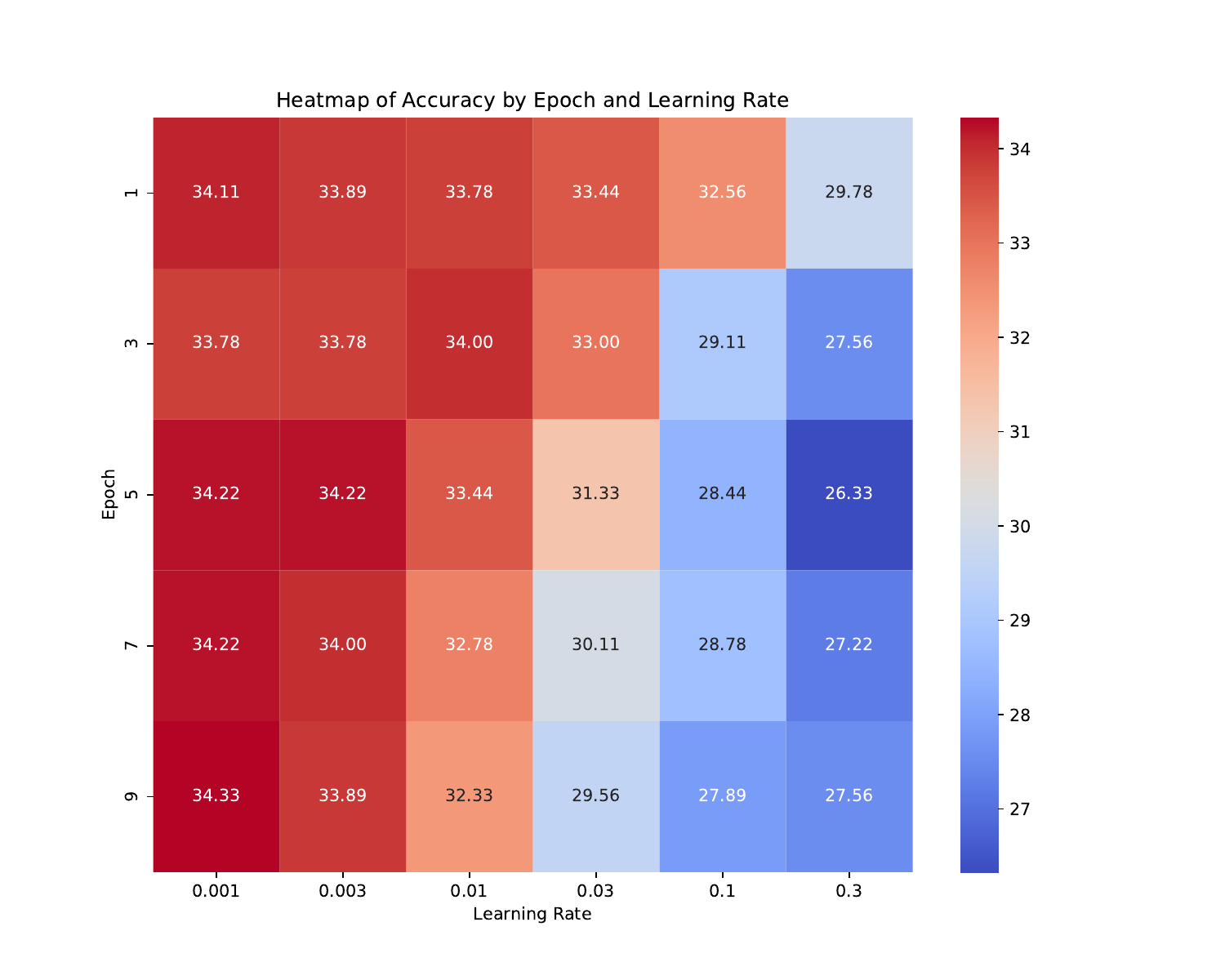}
    \caption{G-MM}
  \end{subfigure}
  \hfill
  \begin{subfigure}[b]{0.19\textwidth}
    \includegraphics[width=\textwidth]{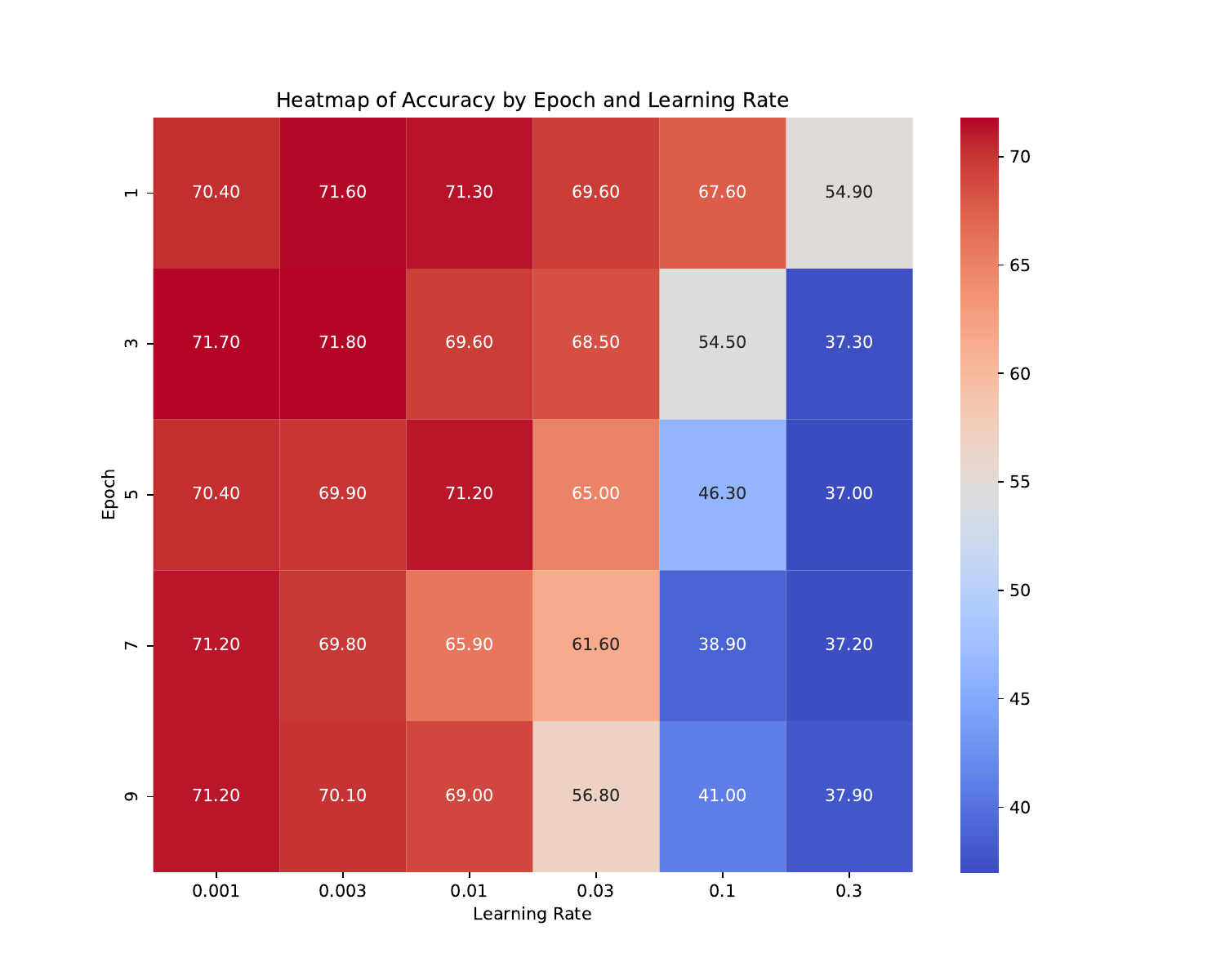}
    \caption{G-L}
  \end{subfigure}
  \hfill
  \begin{subfigure}[b]{0.19\textwidth}
    \includegraphics[width=\textwidth]{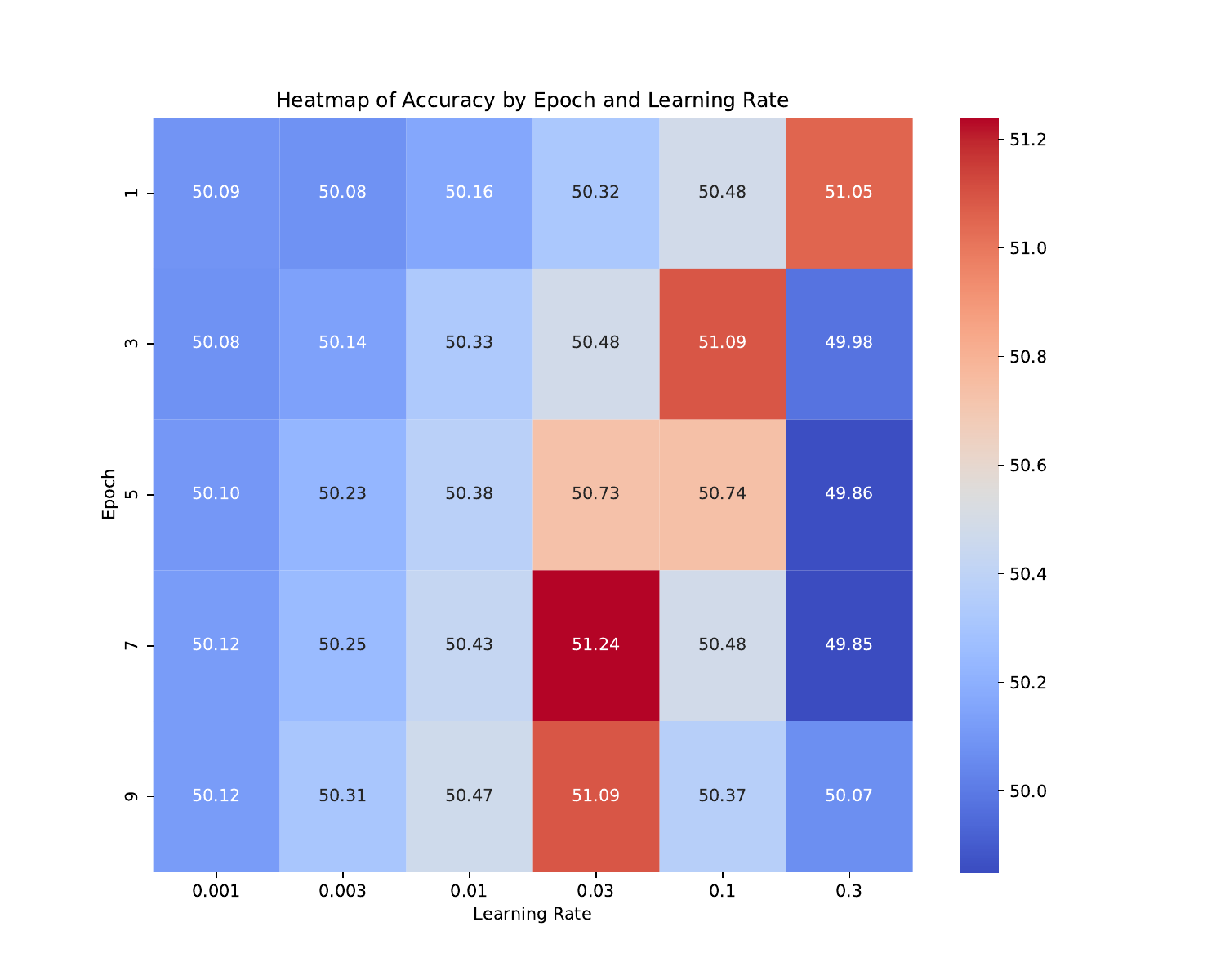}
    \caption{G-ML}
  \end{subfigure}
  \hfill
  \begin{subfigure}[b]{0.19\textwidth}
    \includegraphics[width=\textwidth]{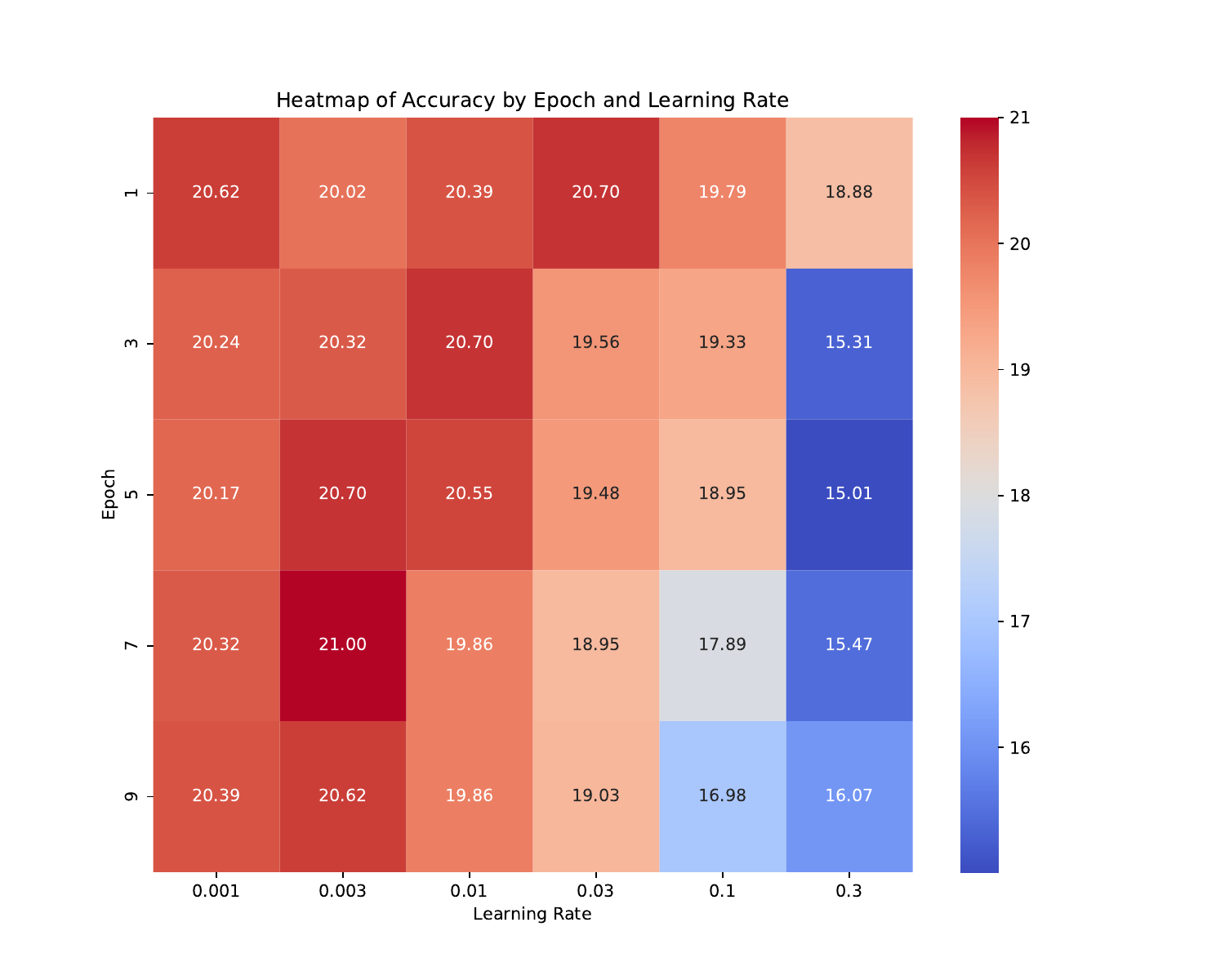}
    \caption{G-G}
  \end{subfigure}
  \hfill
  \begin{subfigure}[b]{0.19\textwidth}
    \includegraphics[width=\textwidth]{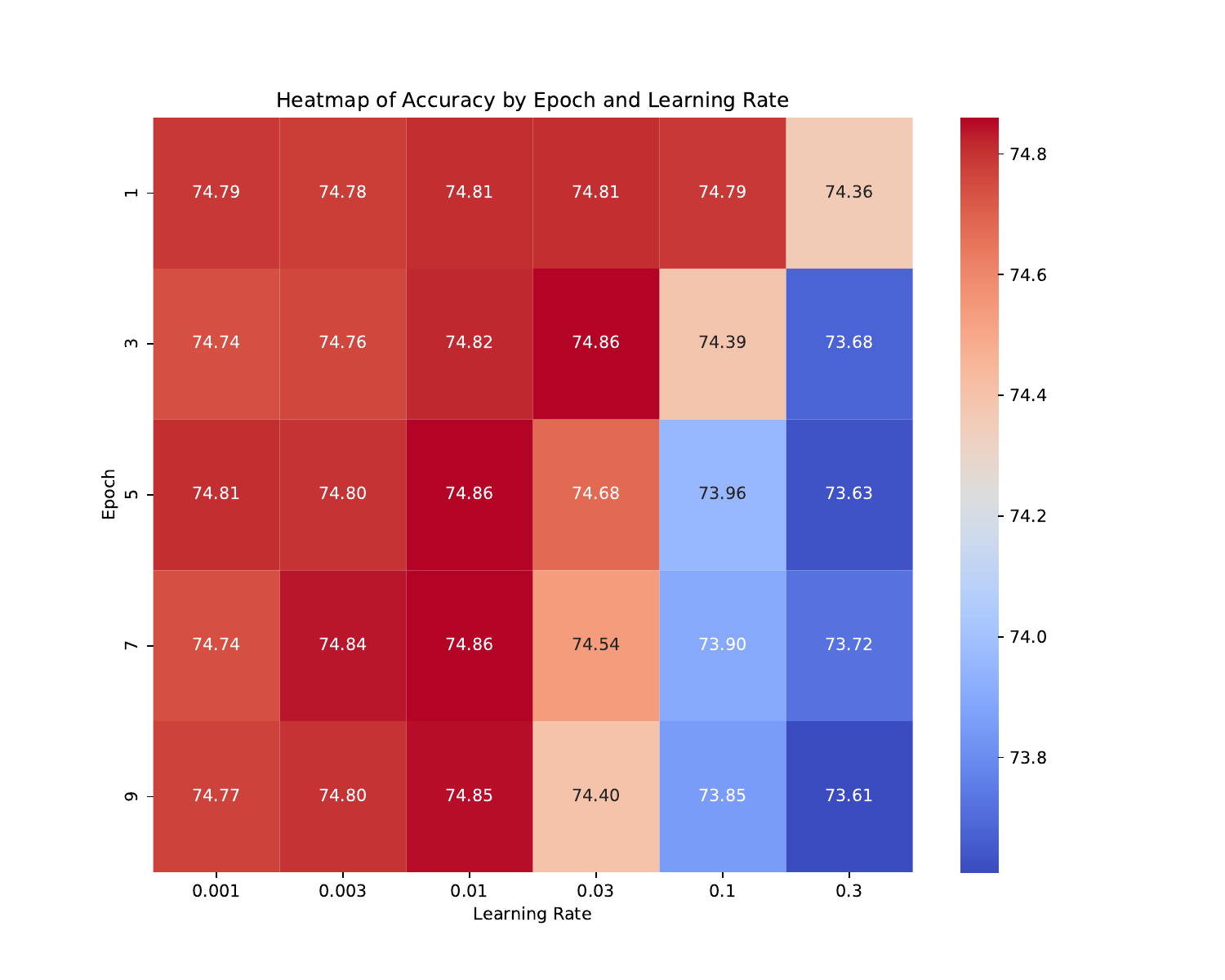}
    \caption{G-H}
  \end{subfigure}
  \hfill

  \begin{subfigure}[b]{0.19\textwidth}
    \includegraphics[width=\textwidth]{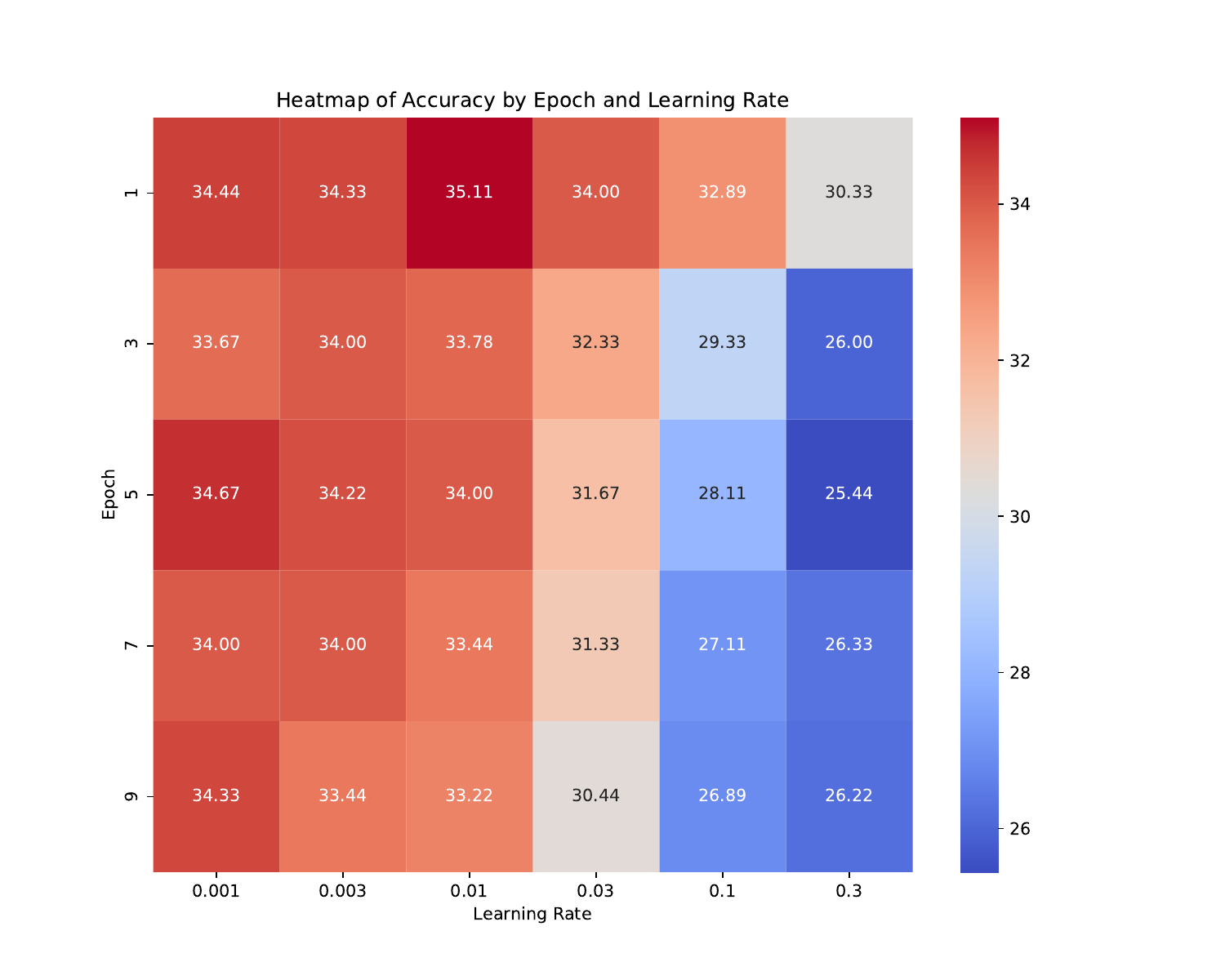}
    \caption{H-MM}
  \end{subfigure}
  \hfill
  \begin{subfigure}[b]{0.19\textwidth}
    \includegraphics[width=\textwidth]{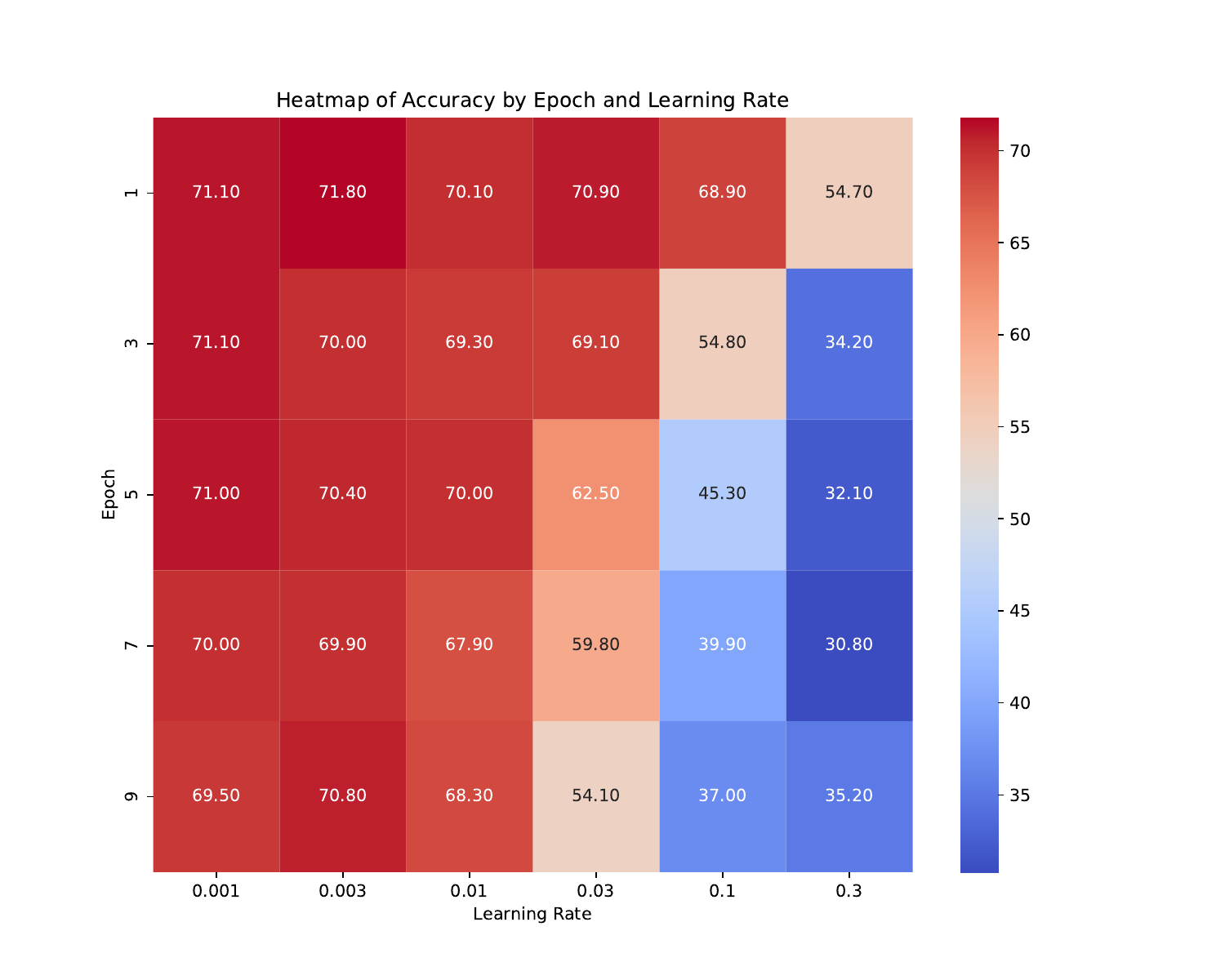}
    \caption{H-L}
  \end{subfigure}
  \hfill
  \begin{subfigure}[b]{0.19\textwidth}
    \includegraphics[width=\textwidth]{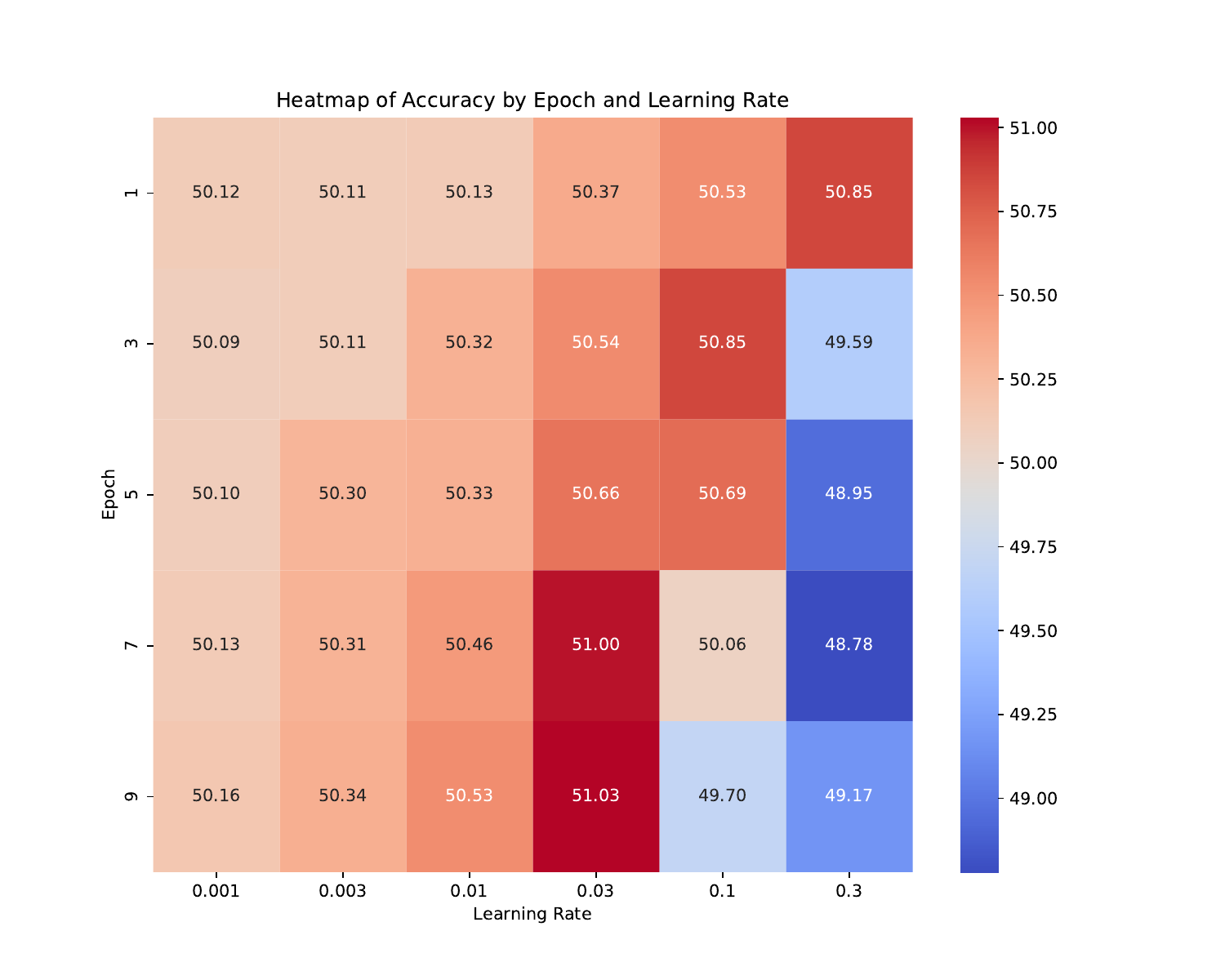}
    \caption{H-ML}
  \end{subfigure}
  \hfill
  \begin{subfigure}[b]{0.19\textwidth}
    \includegraphics[width=\textwidth]{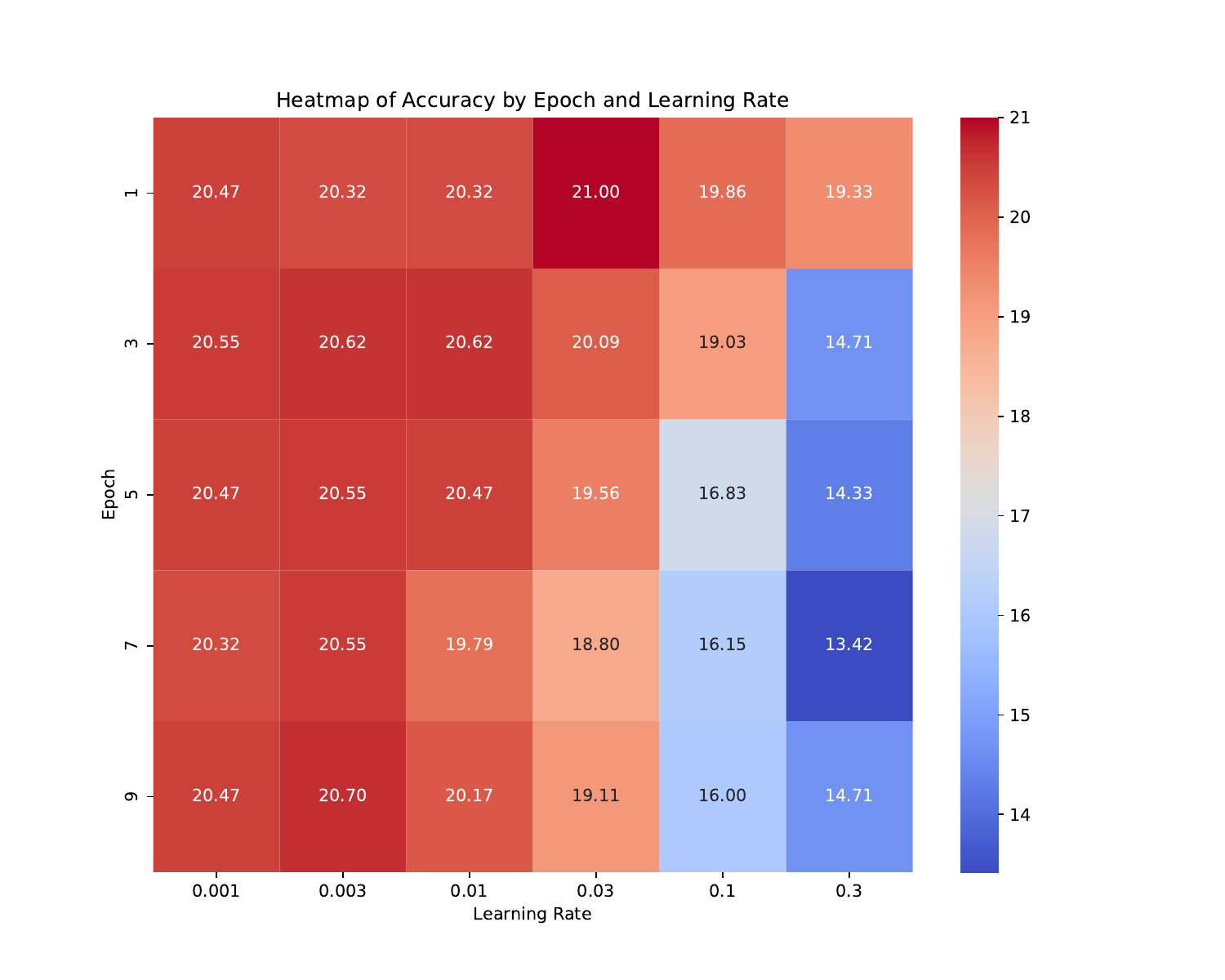}
    \caption{H-G}
  \end{subfigure}
  \hfill
  \begin{subfigure}[b]{0.19\textwidth}
    \includegraphics[width=\textwidth]{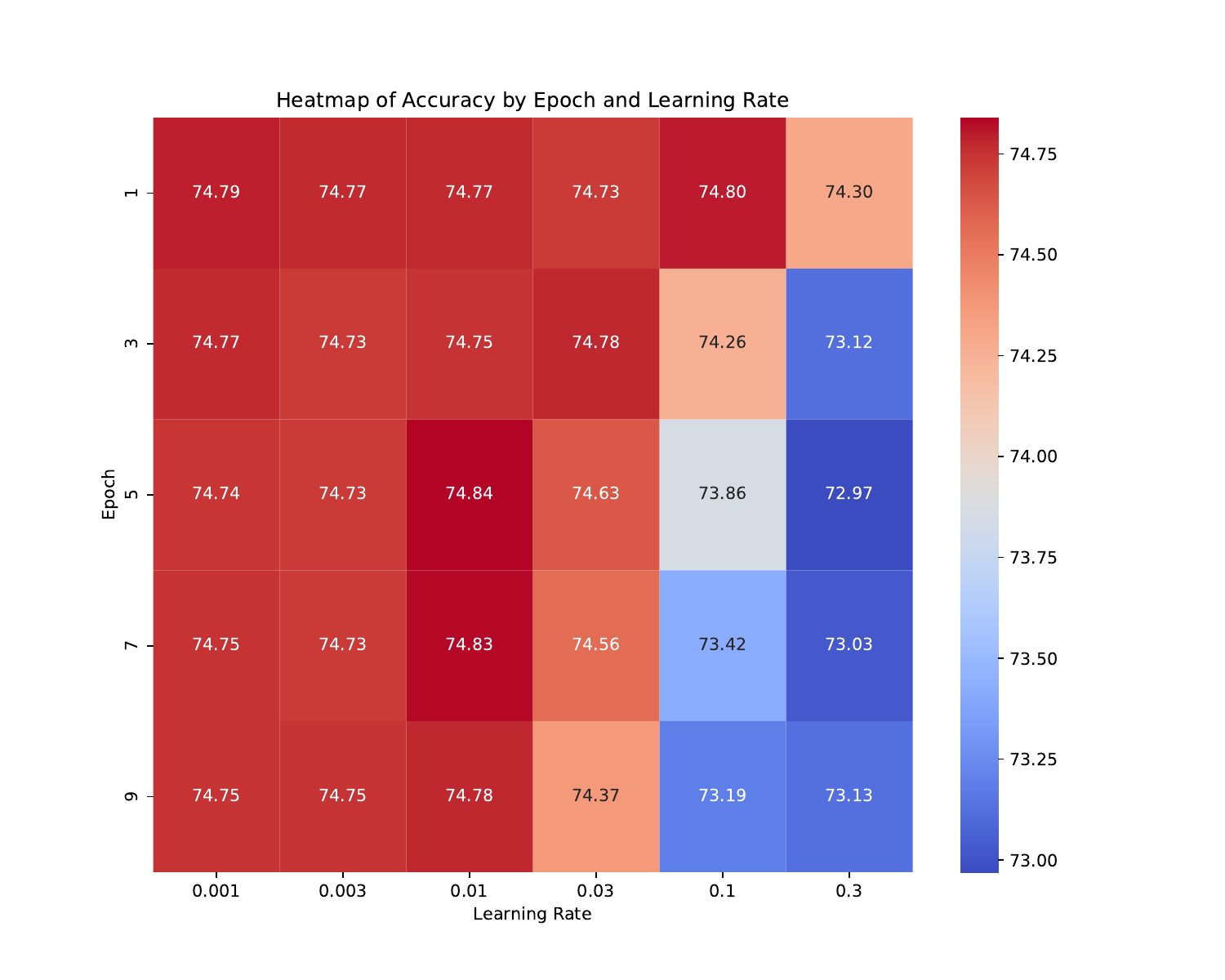}
    \caption{H-H}
  \end{subfigure}

  \caption{Complete visualization results of the first round ablation for each individual meta set.}\label{fig:1_set_ablation_supp}
\end{figure*}

We provide complete first round ablation visualizations in Fig.~\ref{fig:1_set_ablation_supp}.
It contains the complete finetuning and test set combinations, which is discussed in the \textbf{First Round} section in Sec.~\ref{sec:experiments}.

\subsection{Complete Second Round Ablation Visualizations}
\begin{figure*}[htbp]
  \centering
  \begin{subfigure}[b]{0.24\textwidth}
    \includegraphics[width=\textwidth]{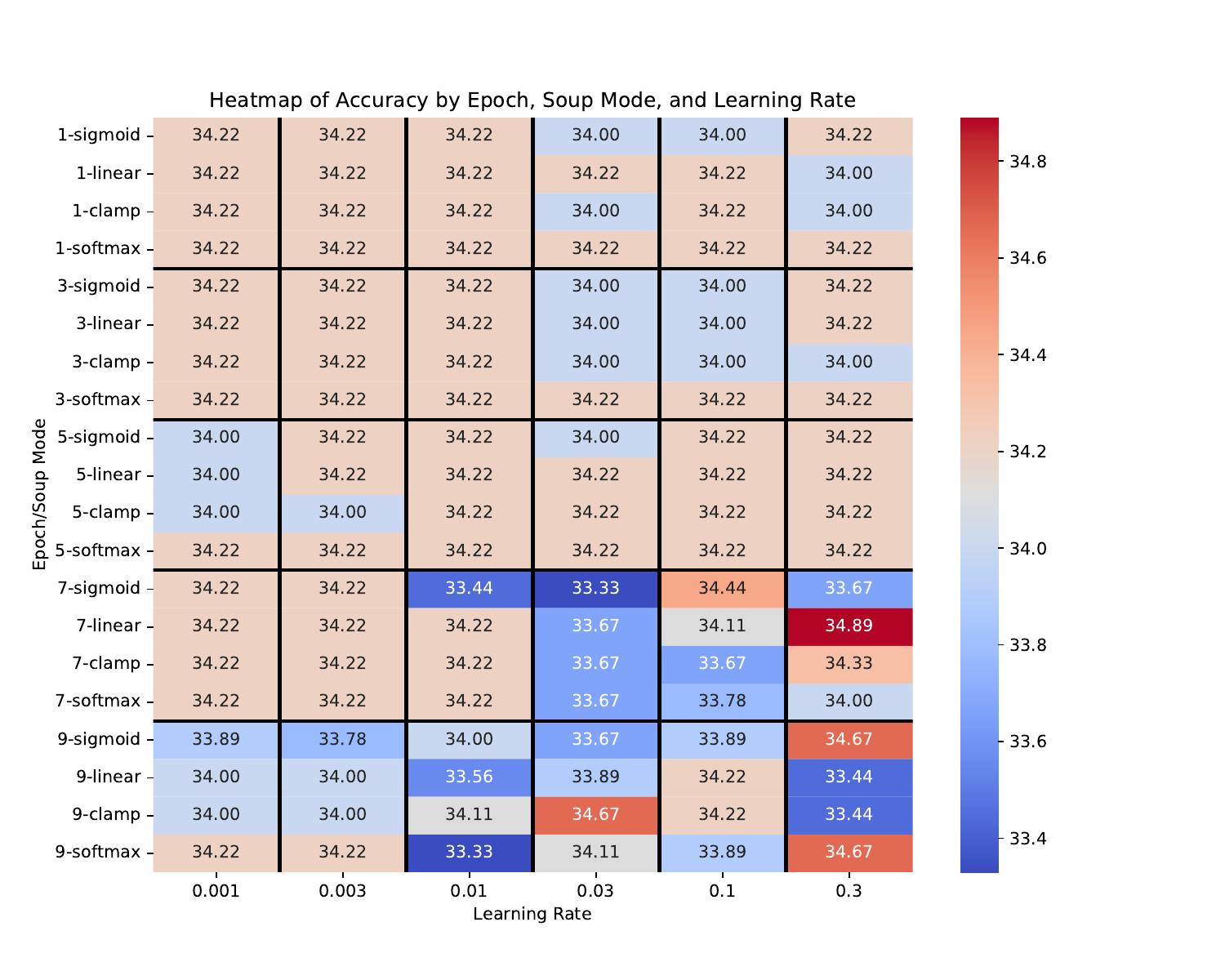}
    \caption{10:MM}
  \end{subfigure}
  \hfill 
  \begin{subfigure}[b]{0.24\textwidth}
    \includegraphics[width=\textwidth]{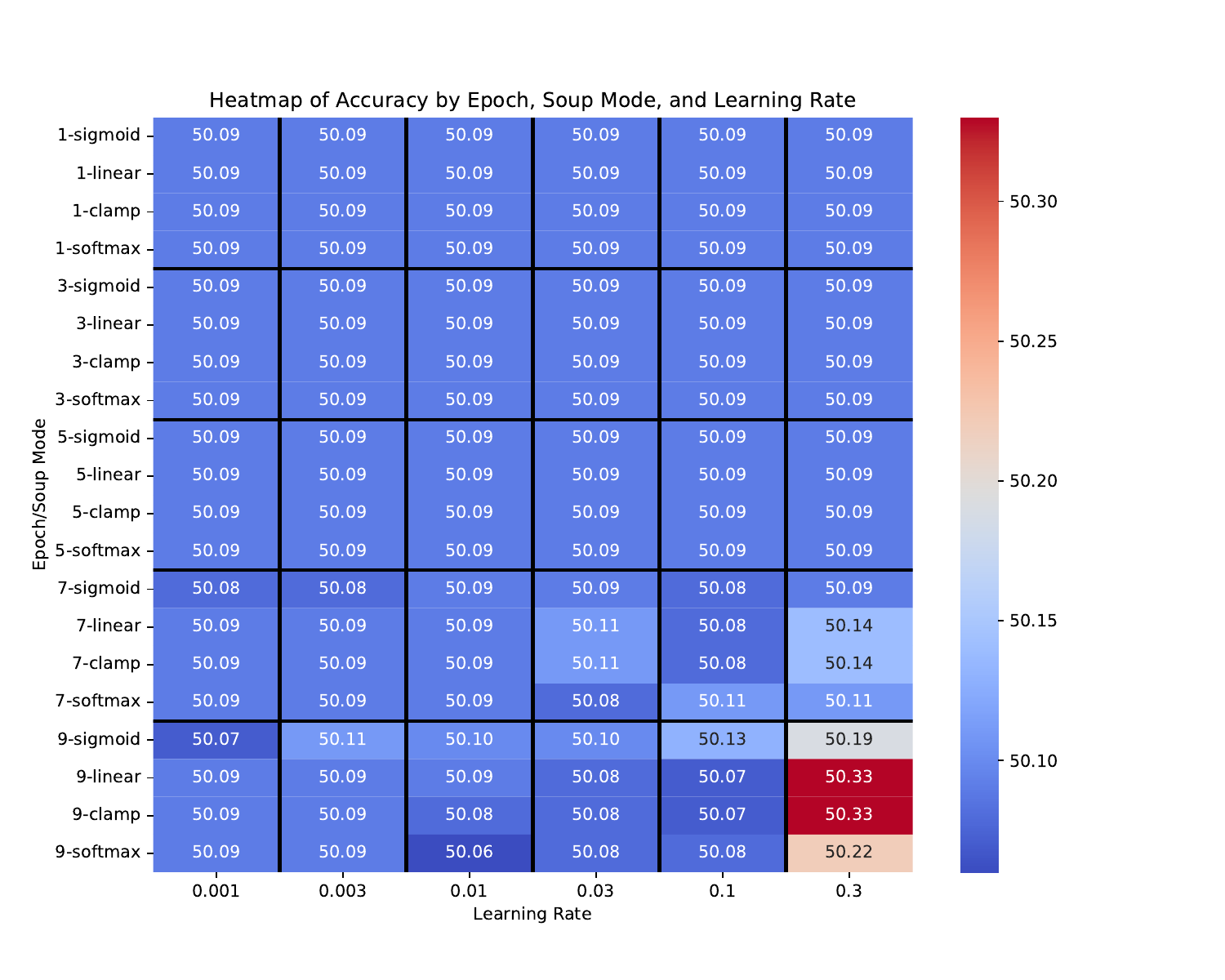}
    \caption{10:ML}
  \end{subfigure}
  \hfill 
  \begin{subfigure}[b]{0.24\textwidth}
    \includegraphics[width=\textwidth]{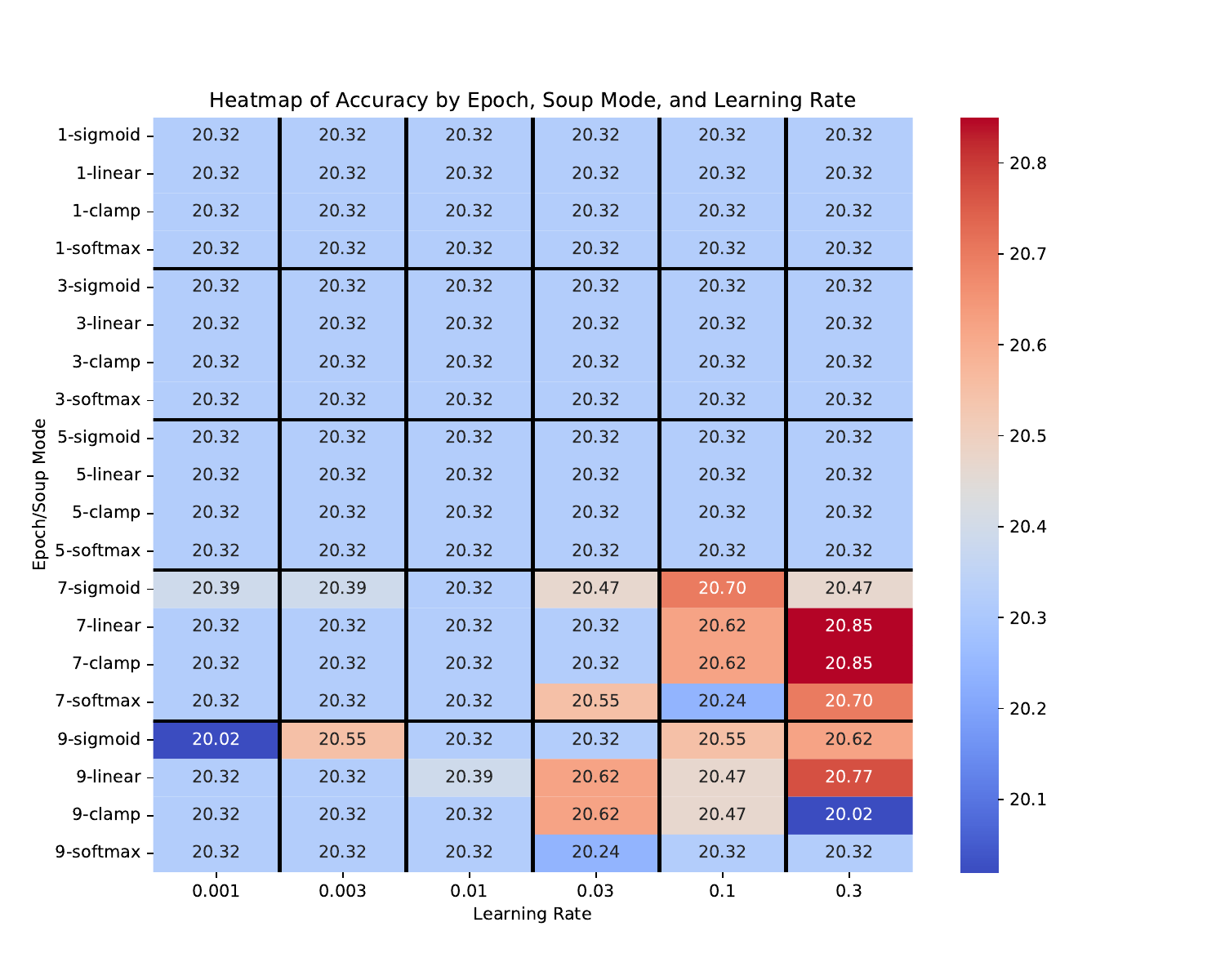}
    \caption{10:G}
  \end{subfigure}
  \hfill 
  \begin{subfigure}[b]{0.24\textwidth}
    \includegraphics[width=\textwidth]{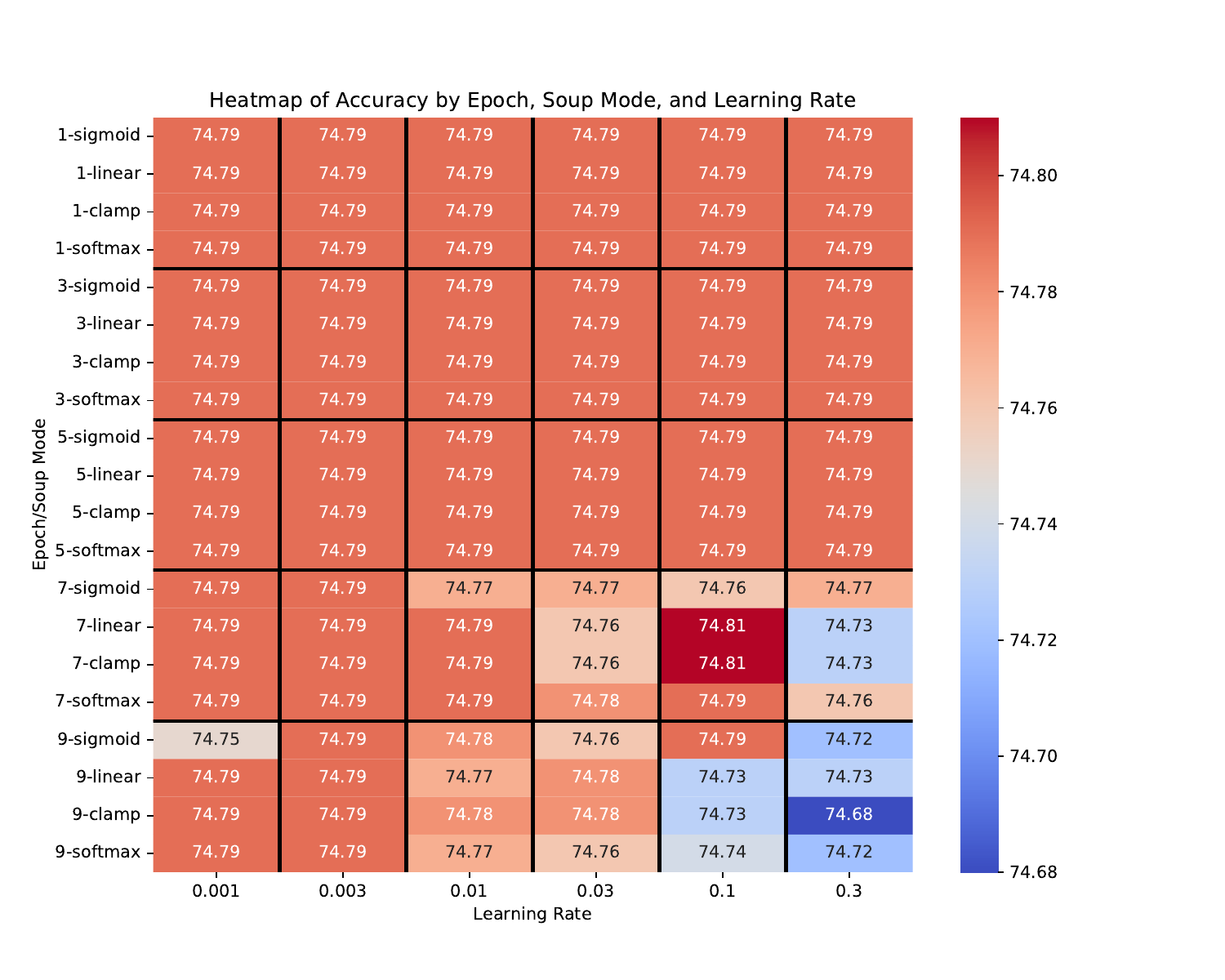}
    \caption{10:H}
  \end{subfigure}
  \hfill

  \begin{subfigure}[b]{0.24\textwidth}
    \includegraphics[width=\textwidth]{figs/llava665k_mmlu_ablation_1/50-50_mmmu.pdf}
    \caption{50:MM}
  \end{subfigure}
  \hfill 
  \begin{subfigure}[b]{0.24\textwidth}
    \includegraphics[width=\textwidth]{figs/llava665k_mmlu_ablation_1/50-50_mmlu.pdf}
    \caption{50:ML}
  \end{subfigure}
  \hfill 
  \begin{subfigure}[b]{0.24\textwidth}
    \includegraphics[width=\textwidth]{figs/llava665k_mmlu_ablation_1/50-50_gsm8k.pdf}
    \caption{50:G}
  \end{subfigure}
  \hfill 
  \begin{subfigure}[b]{0.24\textwidth}
    \includegraphics[width=\textwidth]{figs/llava665k_mmlu_ablation_1/50-50_hellaswag.pdf}
    \caption{50:H}
  \end{subfigure}
  \hfill

  \begin{subfigure}[b]{0.24\textwidth}
    \includegraphics[width=\textwidth]{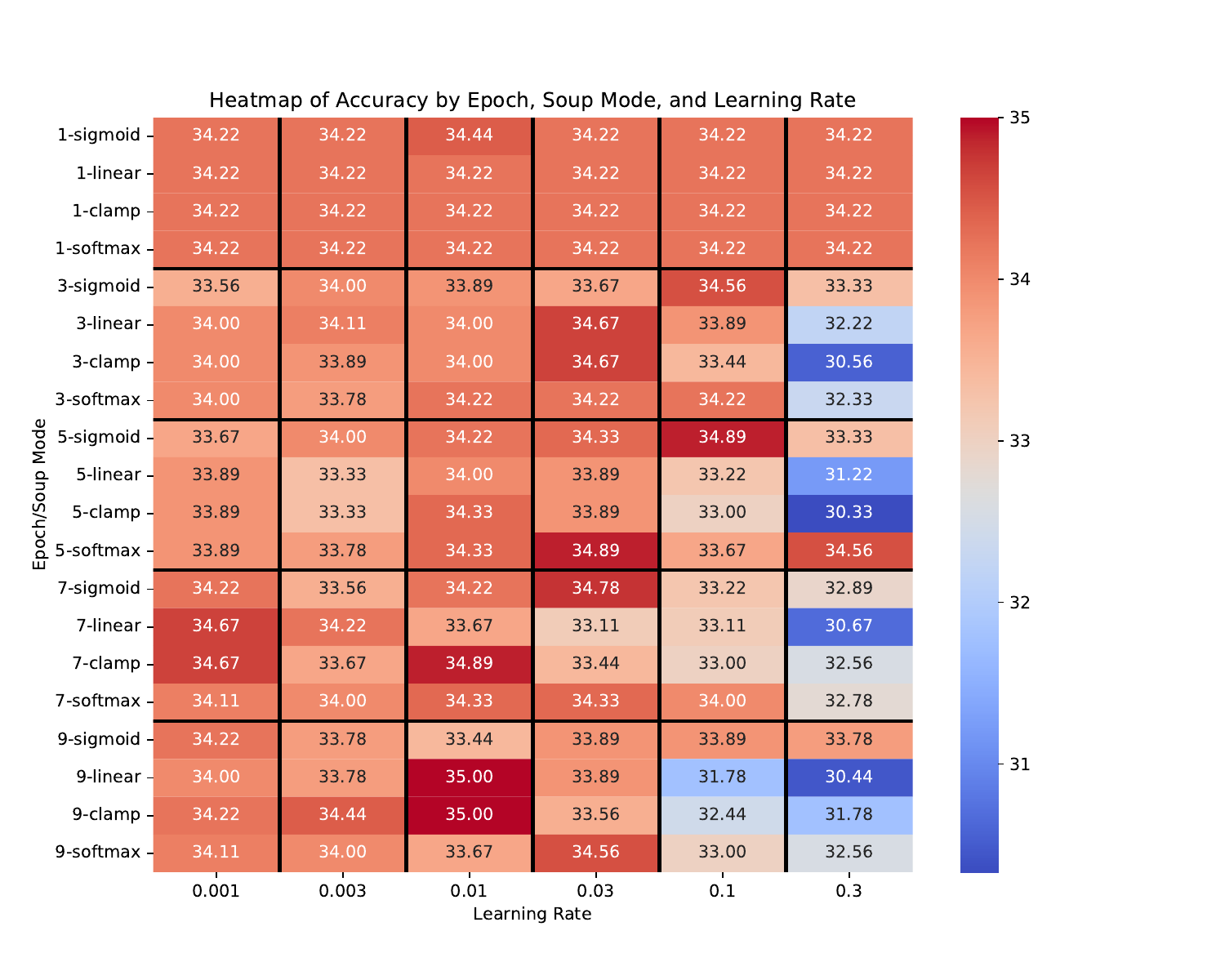}
    \caption{100:MM}
  \end{subfigure}
  \hfill
  \begin{subfigure}[b]{0.24\textwidth}
    \includegraphics[width=\textwidth]{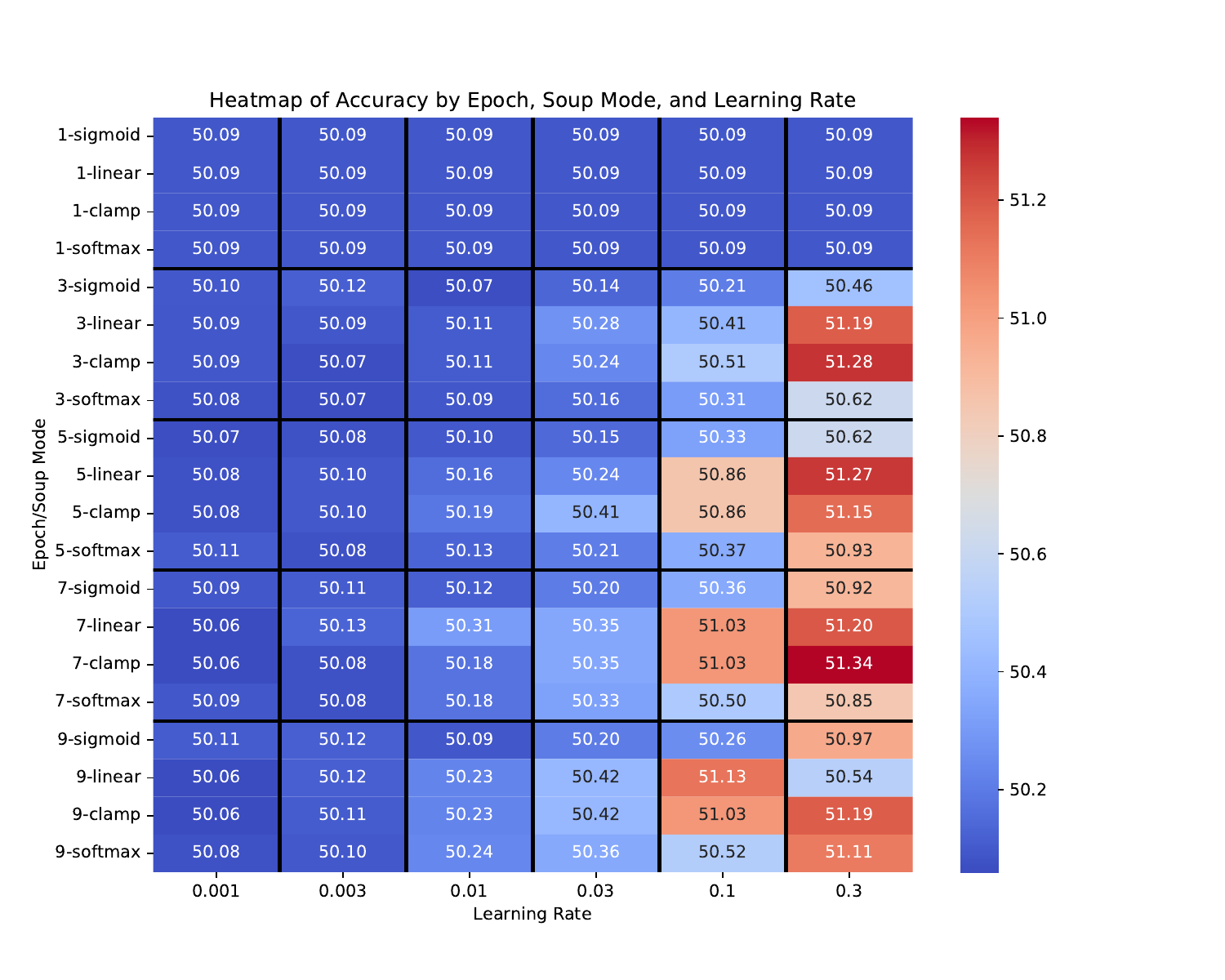}
    \caption{100:ML}
  \end{subfigure}
  \hfill
  \begin{subfigure}[b]{0.24\textwidth}
    \includegraphics[width=\textwidth]{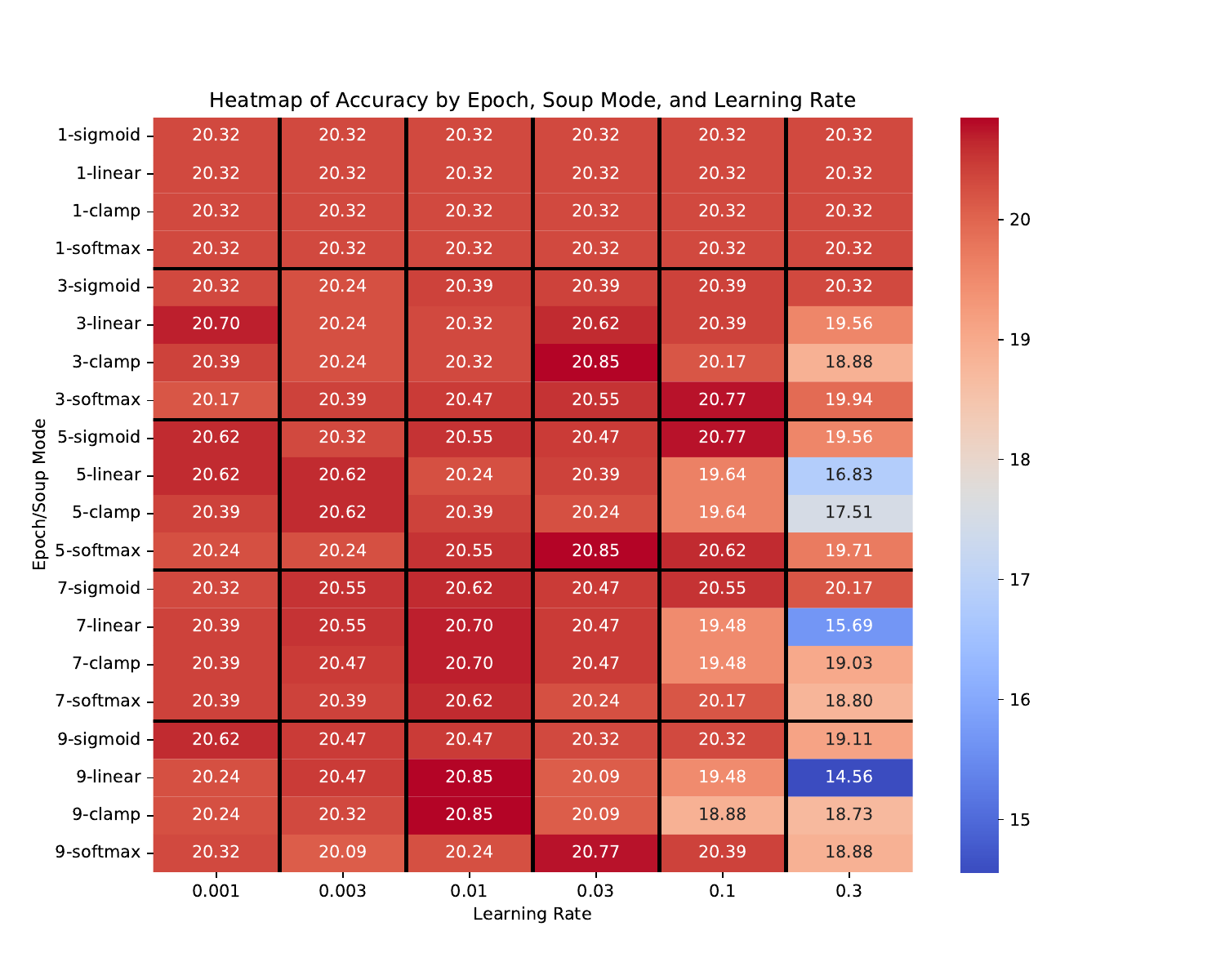}
    \caption{100:G}
  \end{subfigure}
  \hfill
  \begin{subfigure}[b]{0.24\textwidth}
    \includegraphics[width=\textwidth]{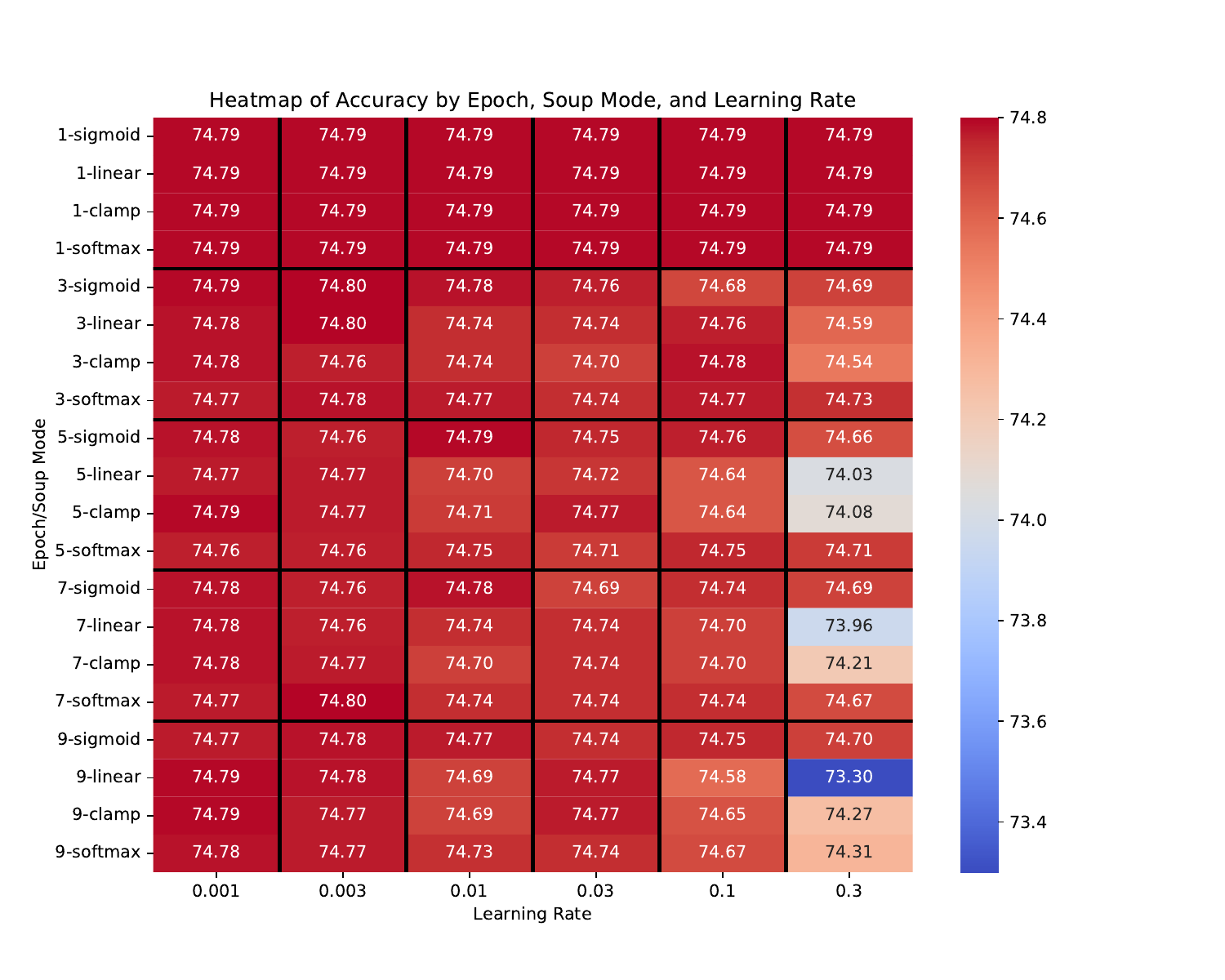}
    \caption{100:H}
  \end{subfigure}
  \hfill

  \begin{subfigure}[b]{0.24\textwidth}
    \includegraphics[width=\textwidth]{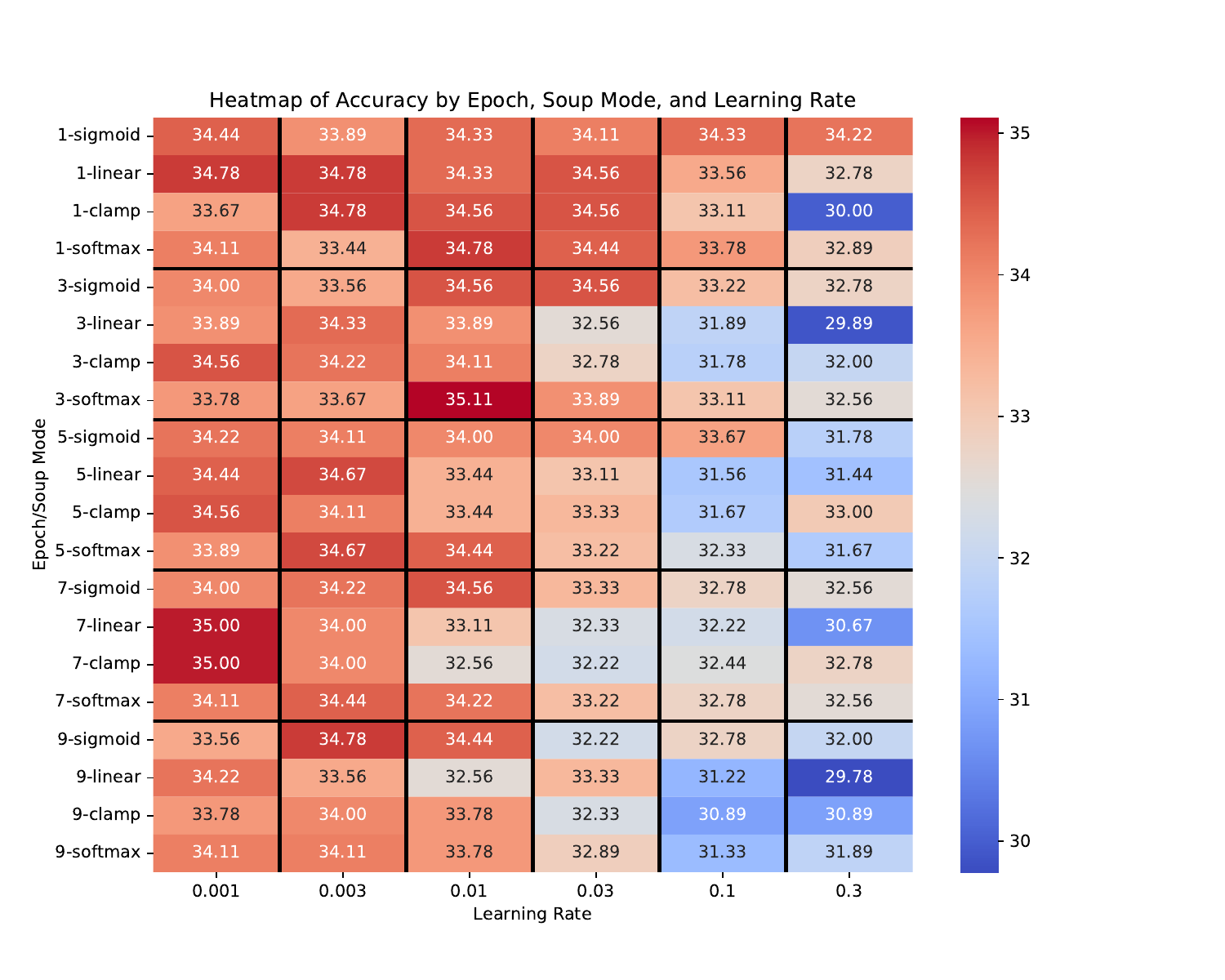}
    \caption{500:MM}
  \end{subfigure}
  \hfill
  \begin{subfigure}[b]{0.24\textwidth}
    \includegraphics[width=\textwidth]{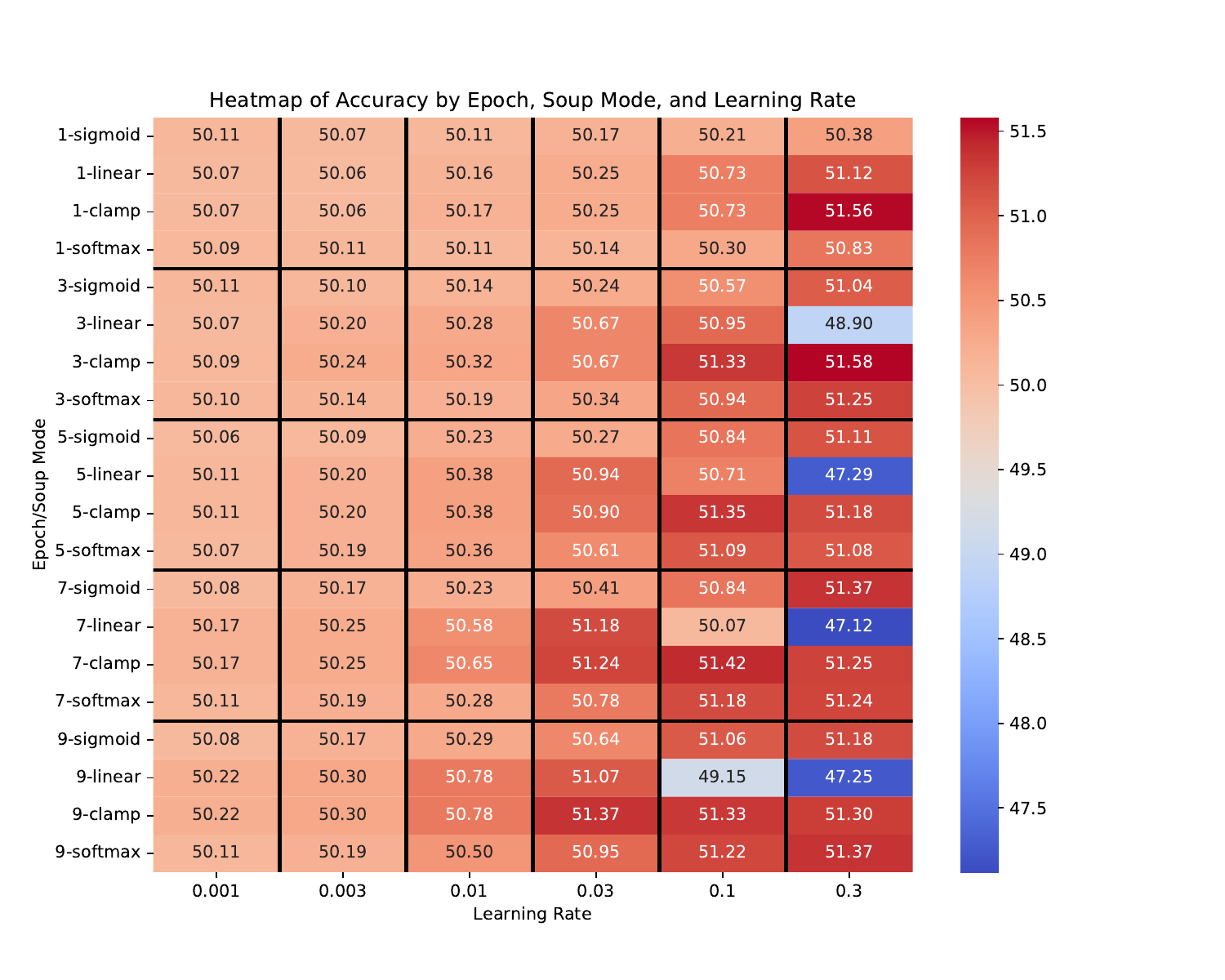}
    \caption{500:ML}
  \end{subfigure}
  \hfill
  \begin{subfigure}[b]{0.24\textwidth}
    \includegraphics[width=\textwidth]{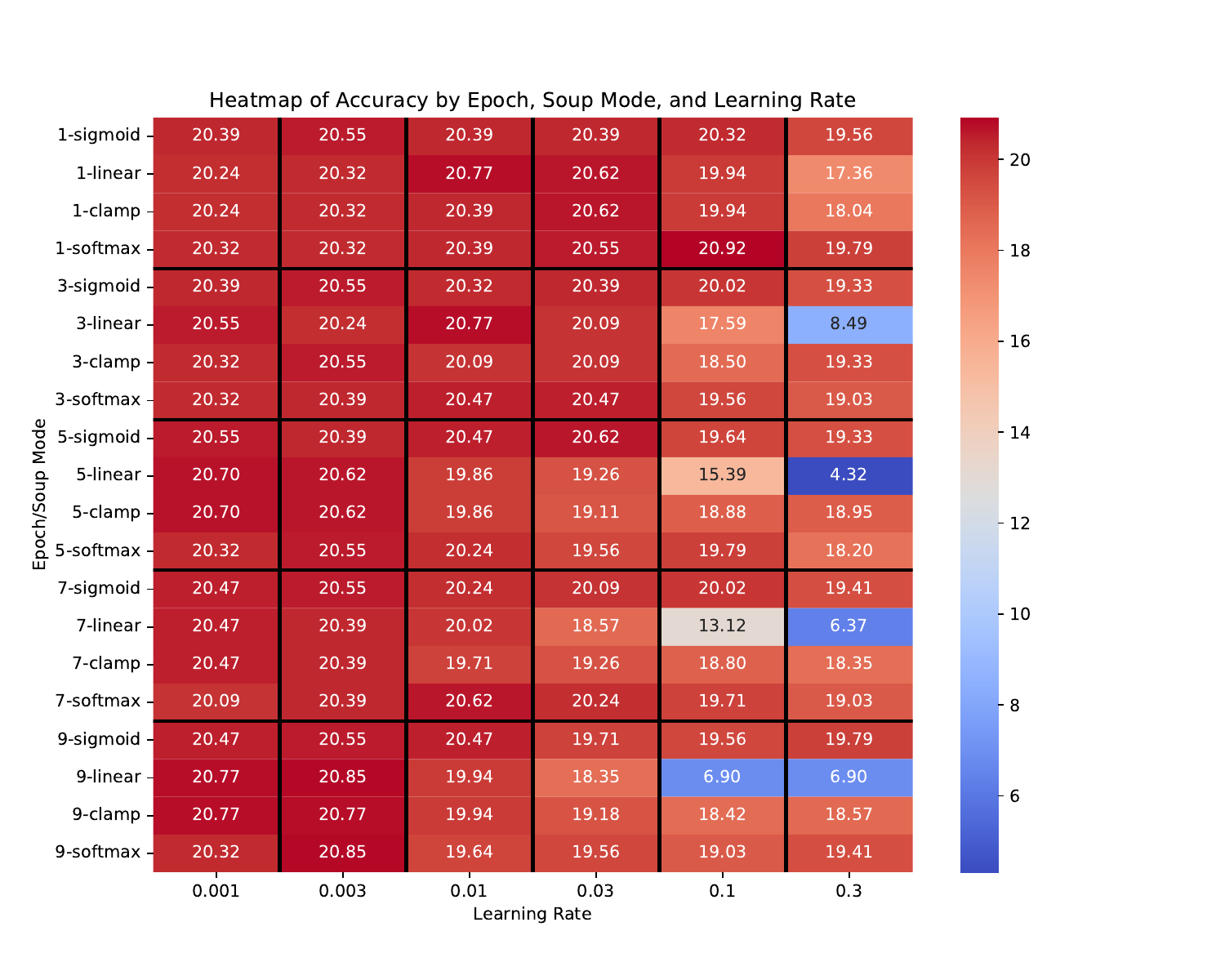}
    \caption{500:G}
  \end{subfigure}
  \hfill
  \begin{subfigure}[b]{0.24\textwidth}
    \includegraphics[width=\textwidth]{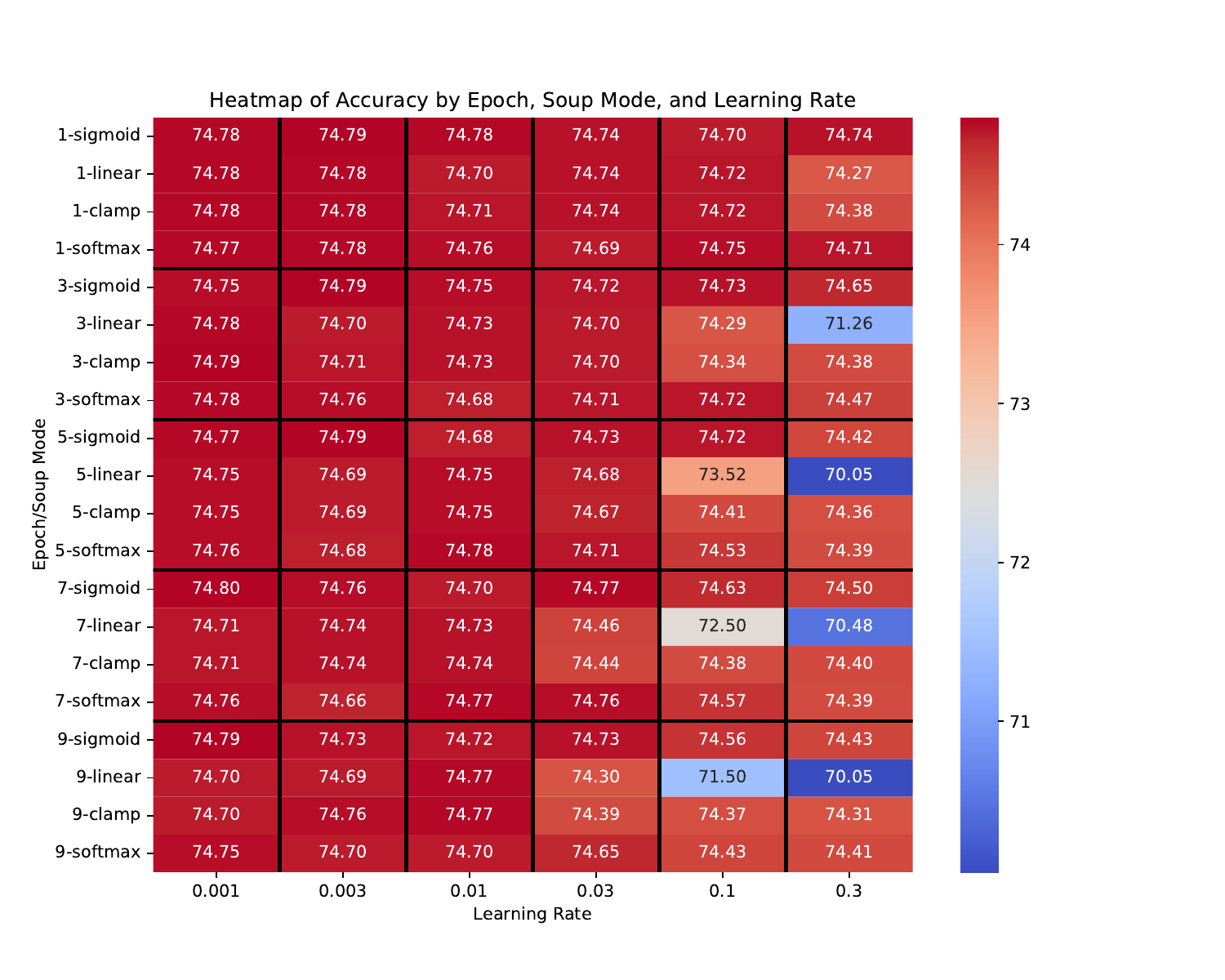}
    \caption{500:H}
  \end{subfigure}
  \hfill

  \begin{subfigure}[b]{0.24\textwidth}
    \includegraphics[width=\textwidth]{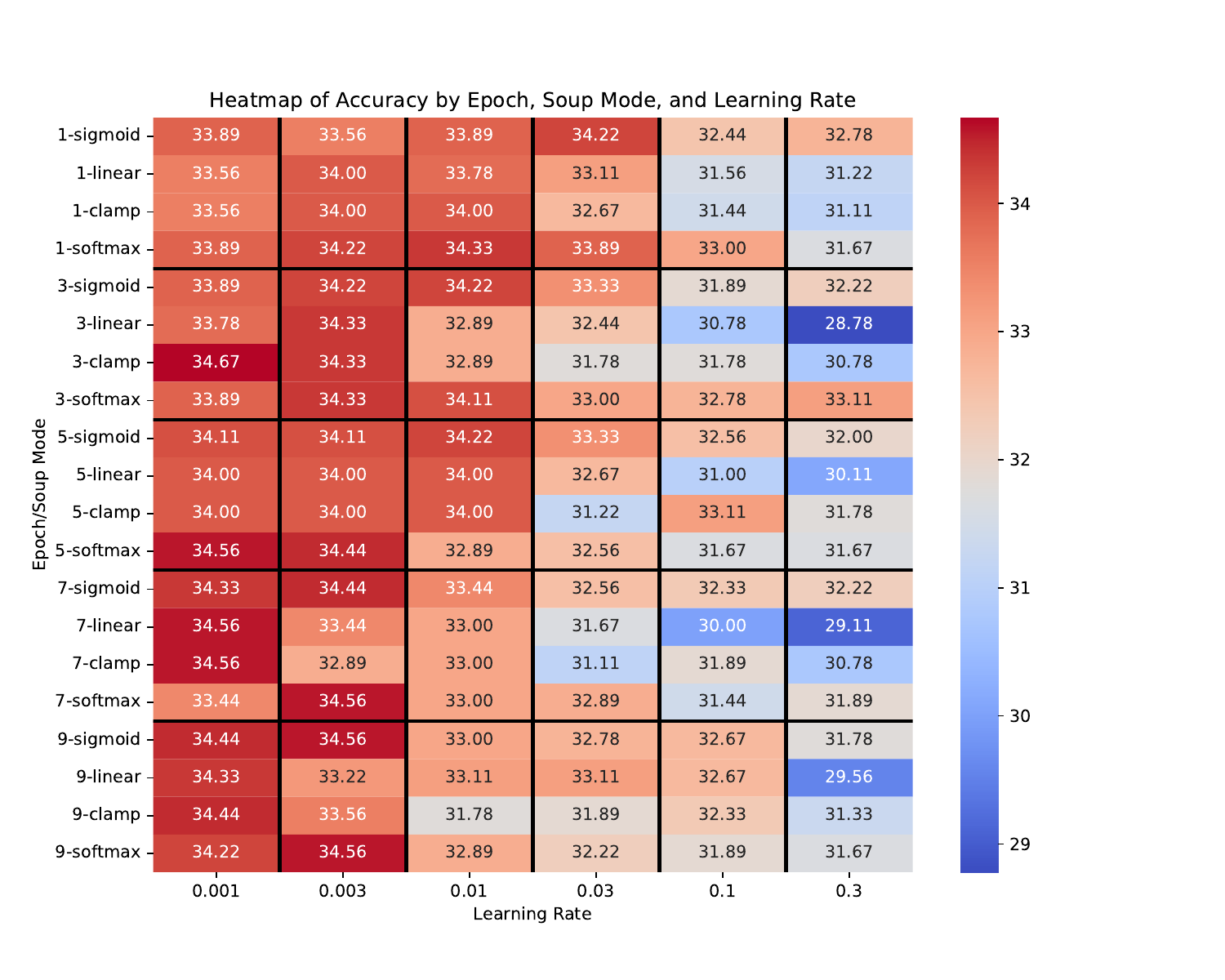}
    \caption{1000:MM}
  \end{subfigure}
  \hfill
  \begin{subfigure}[b]{0.24\textwidth}
    \includegraphics[width=\textwidth]{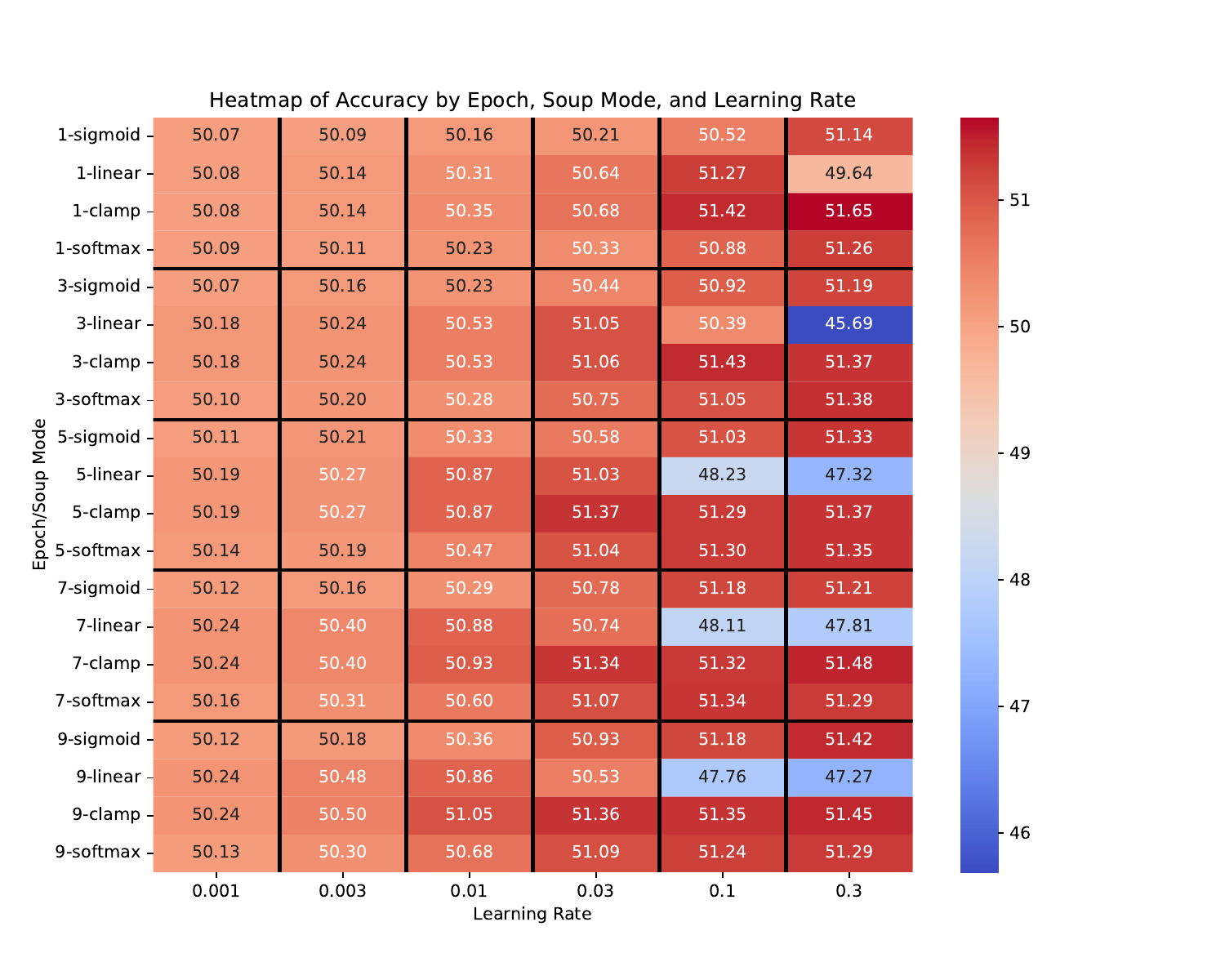}
    \caption{1000:ML}
  \end{subfigure}
  \hfill
  \begin{subfigure}[b]{0.24\textwidth}
    \includegraphics[width=\textwidth]{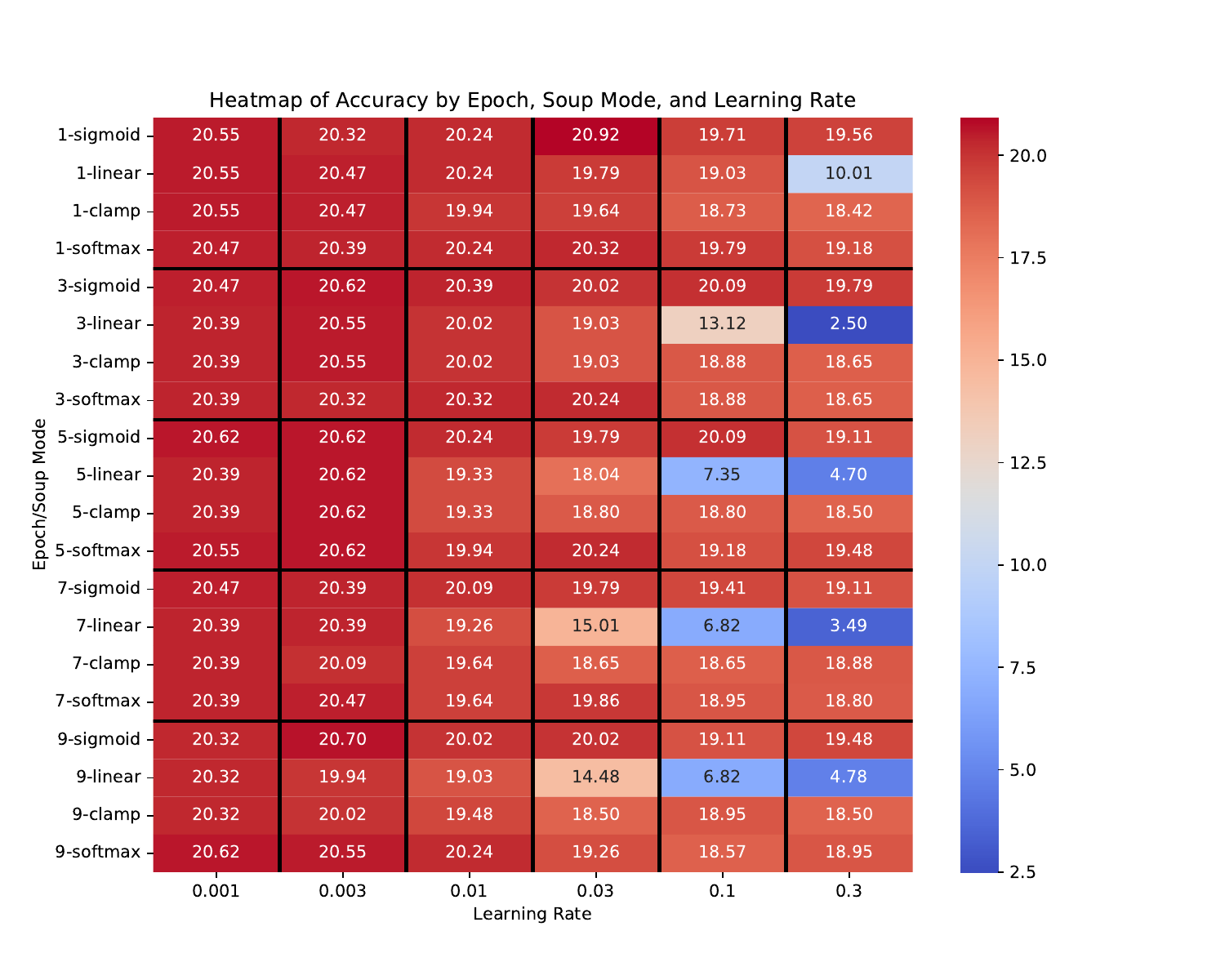}
    \caption{1000:G}
  \end{subfigure}
  \hfill
  \begin{subfigure}[b]{0.24\textwidth}
    \includegraphics[width=\textwidth]{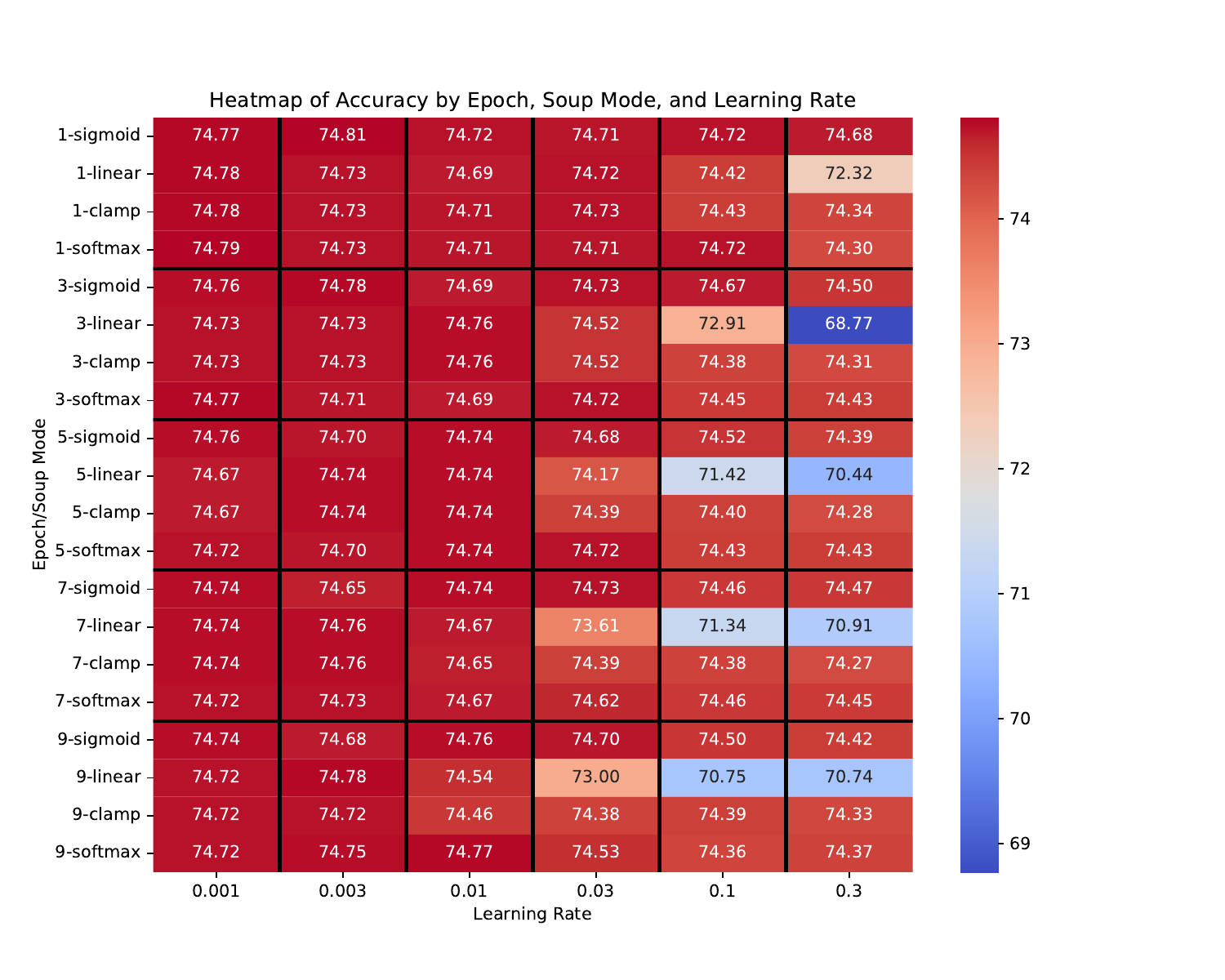}
    \caption{1000:H}
  \end{subfigure}
  \caption{Complete visualization results of the second round ablation for number of samples, epochs, learning rates, and soup activation under LLaVA665K-MMLU meta sets.}\label{fig:2_set_ablation_supp}
\end{figure*}

We provide complete second round ablation visualization in Fig.~\ref{fig:2_set_ablation_supp}.
It contains the complete number of samples settings from 10 to 1000, which is discussed in the \textbf{Second Round} section in Sec.~\ref{sec:experiments}.

\begin{figure*}[htbp]
  \centering
    \includegraphics[width=\textwidth]{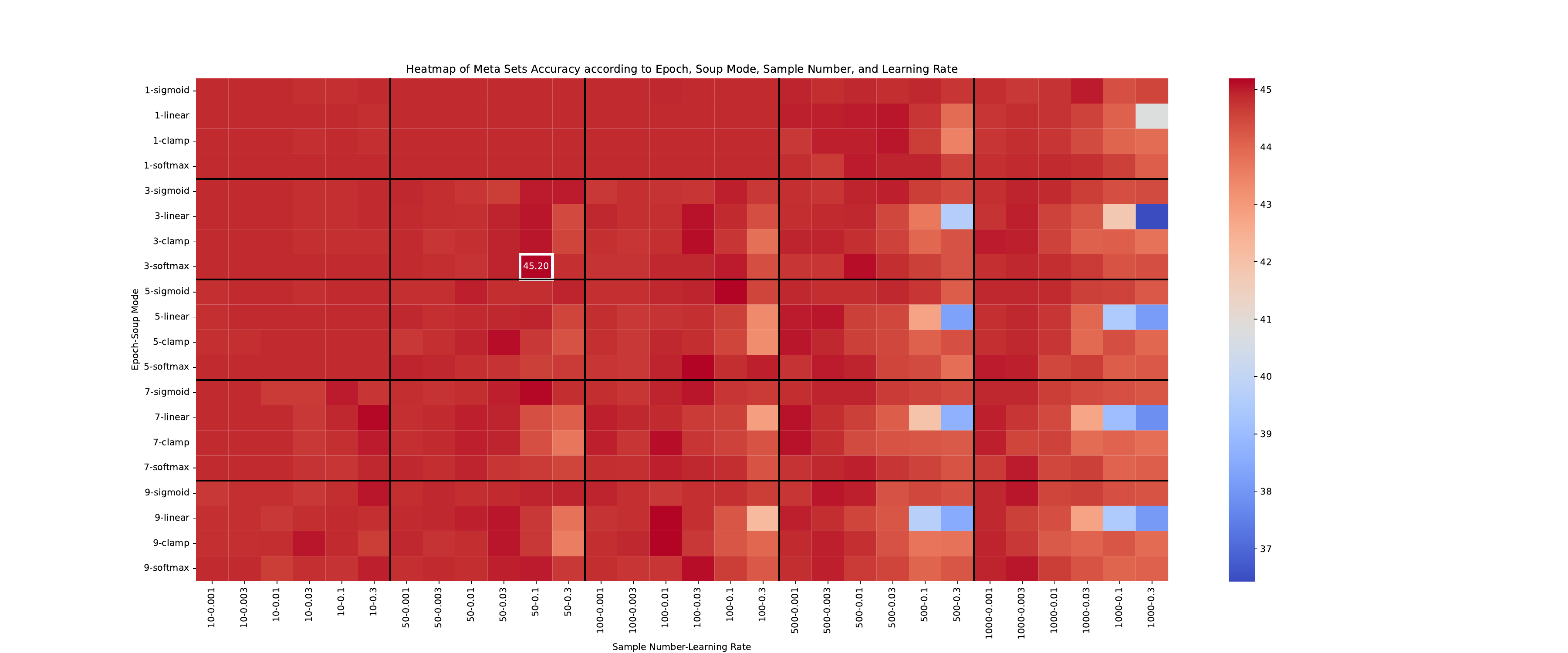}

  \caption{Heatmap visualization of the statistical summary of the second round ablation.
  The best setting with the highest performance is shown in the white box.}\label{fig:2nd_round_stats}
\end{figure*}

The statistical summary of the second round ablation is shown in Fig.~\ref{fig:2nd_round_stats}, used to choose the best hyperparameter combanitions of the second round ablation.

\subsection{Complete $\alpha$ Distribution Visualizations}\label{sec:alpha_dist_supp}
\begin{figure*}[htbp]
  \centering
  \begin{subfigure}[b]{0.95\textwidth}
    \includegraphics[width=\textwidth]{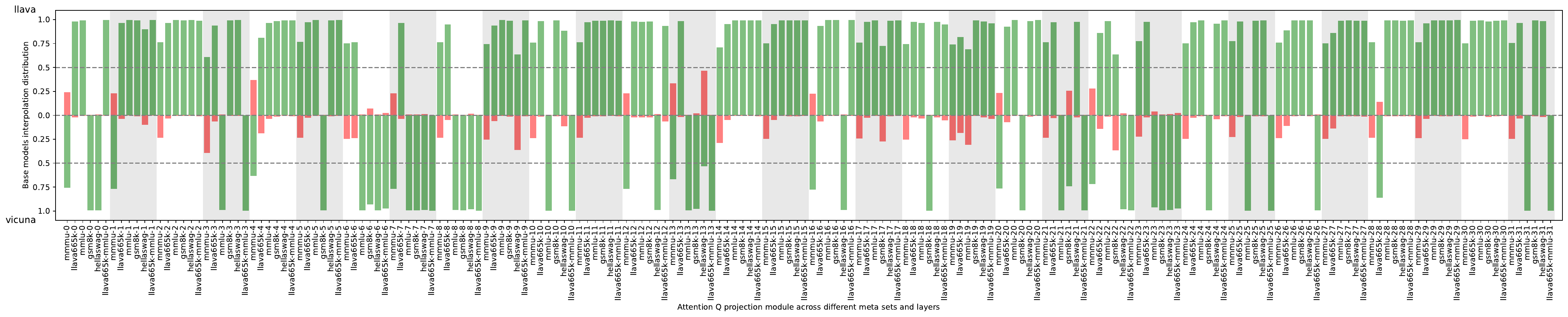}
    \caption{Attention Q mapping.}
  \end{subfigure}
  \hfill

  \begin{subfigure}[b]{0.95\textwidth}
    \includegraphics[width=\textwidth]{figs/alpha_dist/self_attn.k_proj.pdf}
    \caption{Attention K mapping.}
  \end{subfigure}
  \hfill

  \begin{subfigure}[b]{0.95\textwidth}
    \includegraphics[width=\textwidth]{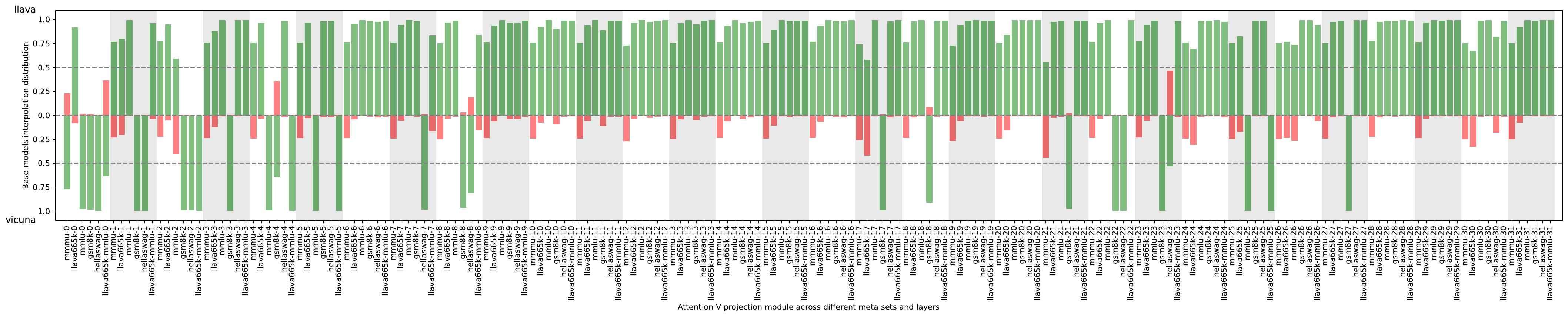}
    \caption{Attnention V mapping.}
  \end{subfigure}
  \hfill

  \begin{subfigure}[b]{0.95\textwidth}
    \includegraphics[width=\textwidth]{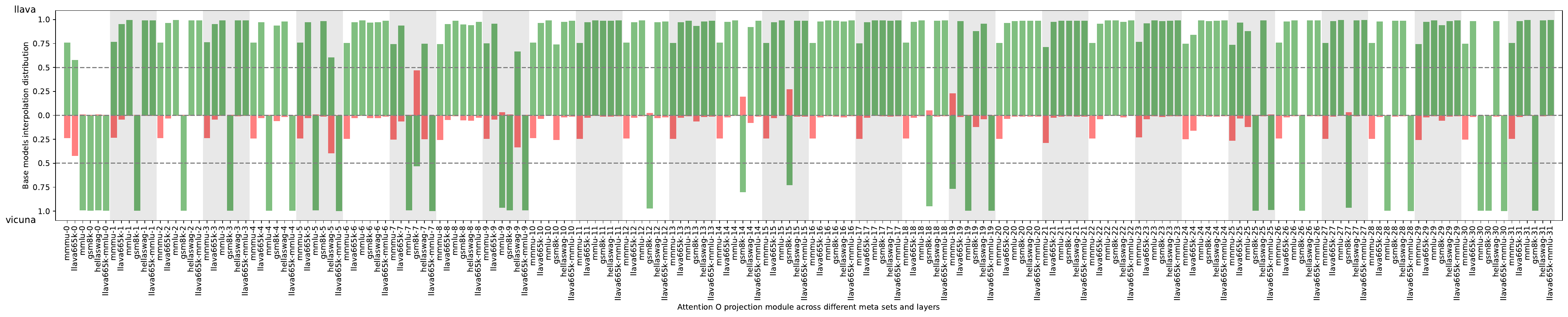}
    \caption{Attention O mapping.}
  \end{subfigure}
  \hfill

  \caption{$\alpha$ distribution visualizations for attention.}\label{fig:set_dist_supp_attn}
\end{figure*}

\begin{figure*}[htbp]
  \centering
  
  \begin{subfigure}[b]{0.95\textwidth}
    \includegraphics[width=\textwidth]{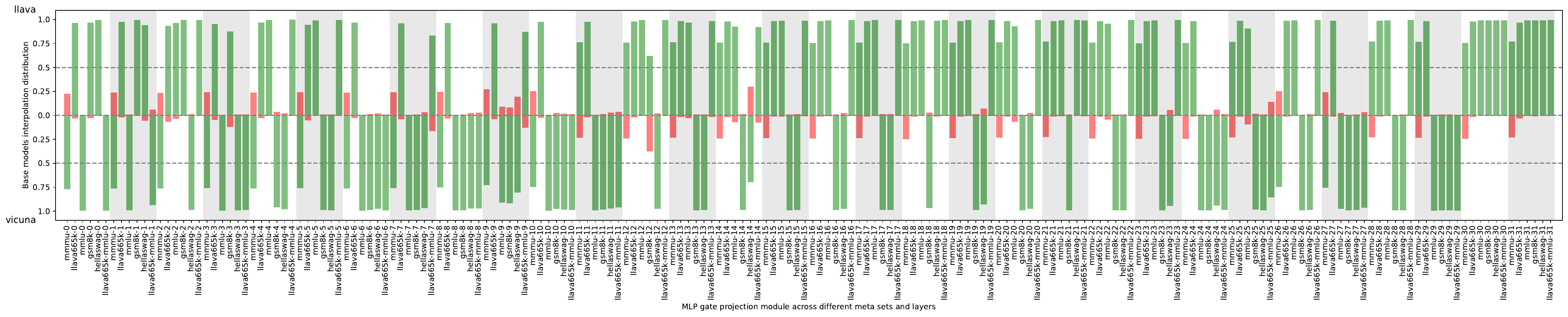}
    \caption{MLP gate mapping.}
  \end{subfigure}
  \hfill
  \begin{subfigure}[b]{0.95\textwidth}
    \includegraphics[width=\textwidth]{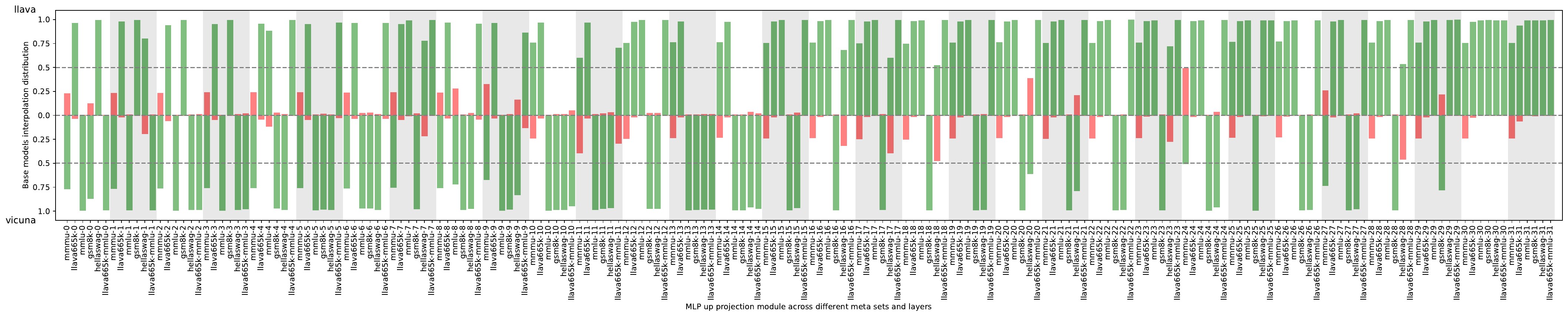}
    \caption{MLP up mapping.}
  \end{subfigure}
  \hfill
  \begin{subfigure}[b]{0.95\textwidth}
    \includegraphics[width=\textwidth]{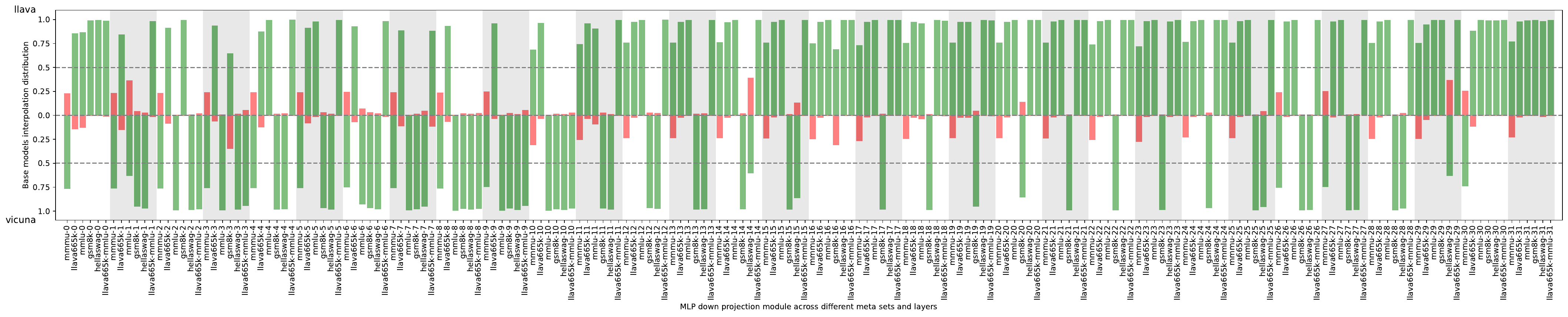}
    \caption{MLP down mapping.}
  \end{subfigure}

  \caption{$\alpha$ distribution visualizations for MLP.}\label{fig:set_dist_supp_mlp}
\end{figure*}

\begin{figure*}[htbp]
  \centering
  \begin{subfigure}[b]{0.95\textwidth}
    \includegraphics[width=\textwidth]{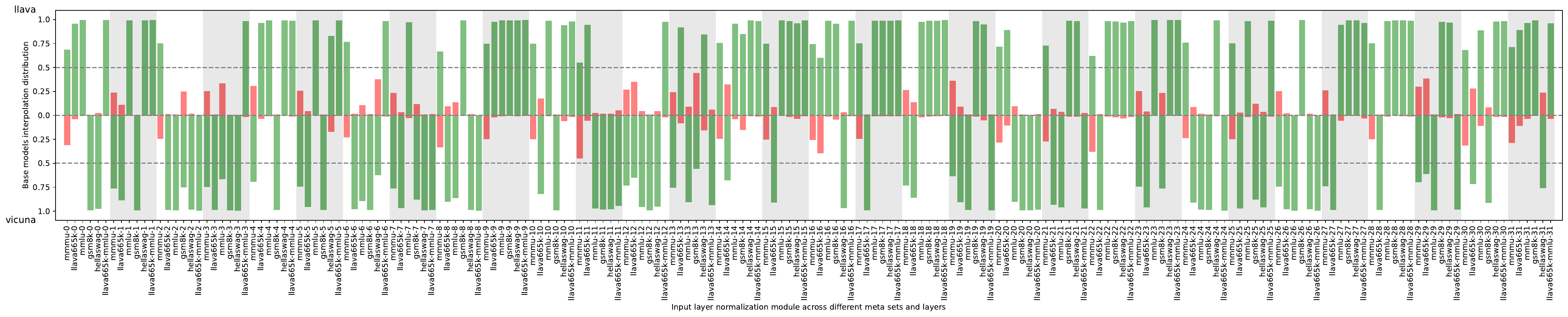}
    \caption{Input layernorm.}
  \end{subfigure}
  \hfill
  \begin{subfigure}[b]{0.95\textwidth}
    \includegraphics[width=\textwidth]{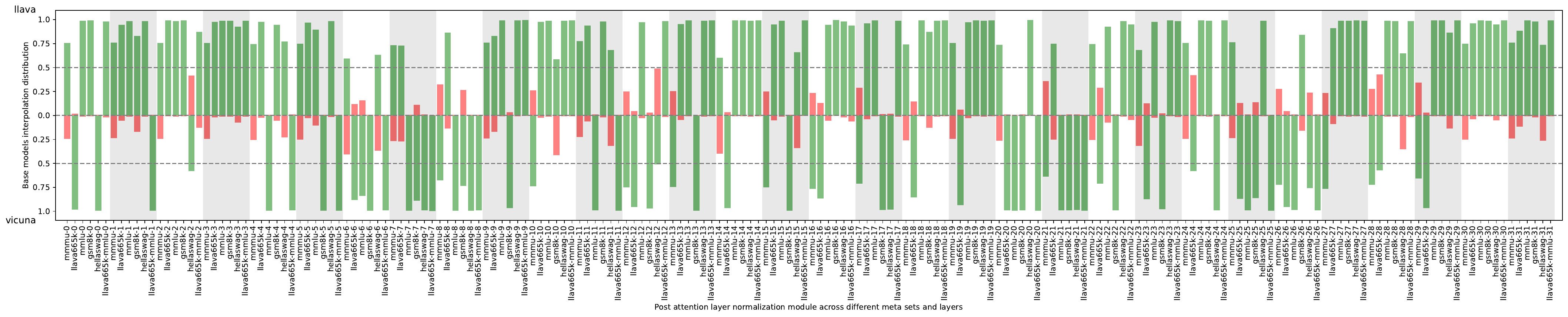}
    \caption{Post attention layernorm.}
  \end{subfigure}

  \caption{$\alpha$ distribution visualizations for layernorm.}\label{fig:set_dist_supp_norm}
\end{figure*}

\begin{figure*}[htbp]
  \centering
    \includegraphics[width=0.6\textwidth]{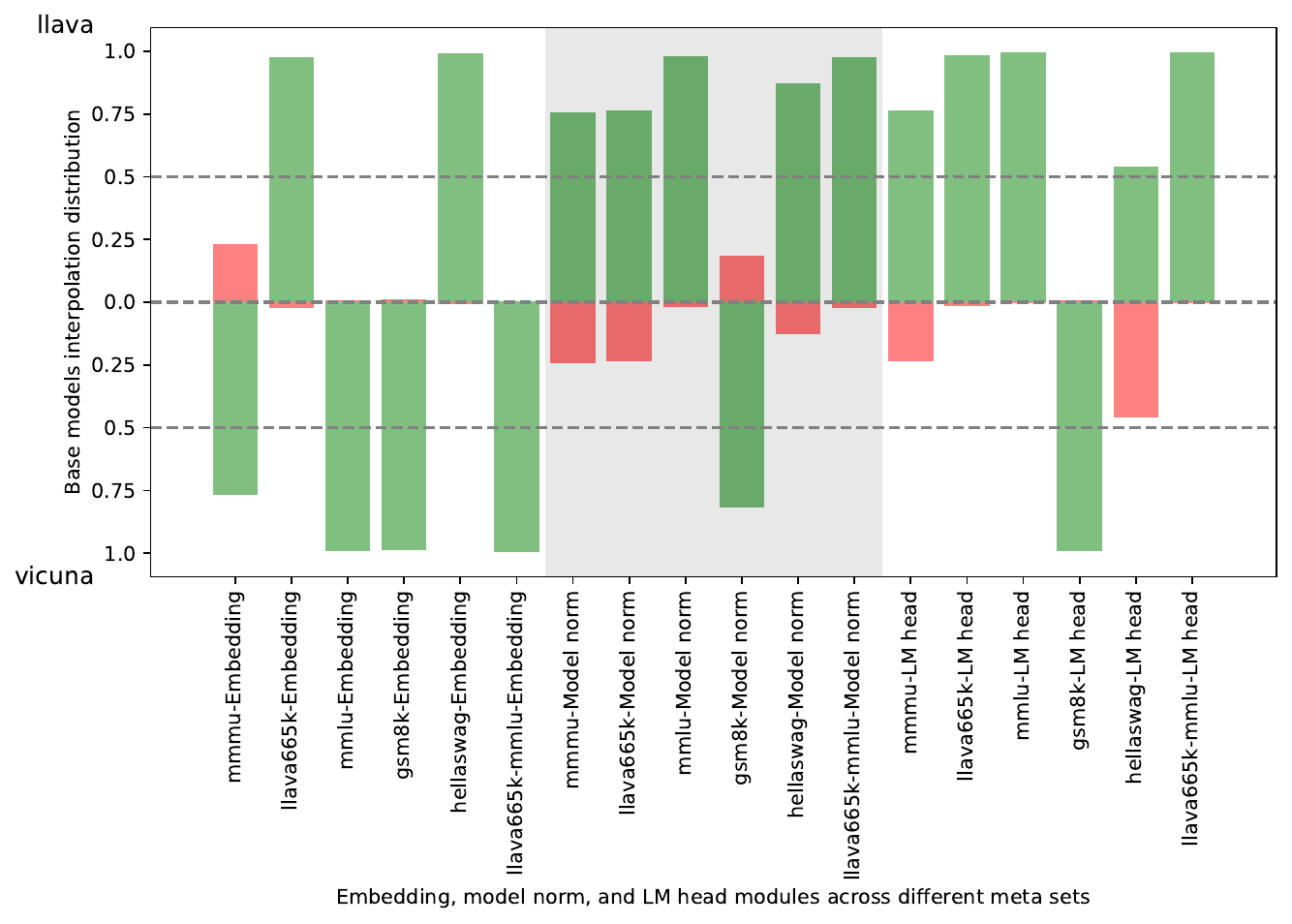}

  \caption{$\alpha$ distribution visualizations of other mappings.}\label{fig:other_alpha_dist}
\end{figure*}

We provide complete $\alpha$ distribution visualizations for different mappings in Fig.~\ref{fig:set_dist_supp_attn}, Fig.~\ref{fig:set_dist_supp_mlp}, and
Fig.~\ref{fig:set_dist_supp_norm}.
They include the mappings of attention, MLP, and normalization blocks, which are discussed in Sec~\ref{sec:soup_behavior}.
We also include visualizations of other mappings in Fig.~\ref{fig:other_alpha_dist}.

\subsection{Complete Regularized $\alpha$ Distribution Visualizations}\label{sec:reg_alpha_dist_supp}
\begin{figure*}[htbp]
  \centering
  \begin{subfigure}[b]{0.95\textwidth}
    \includegraphics[width=\textwidth]{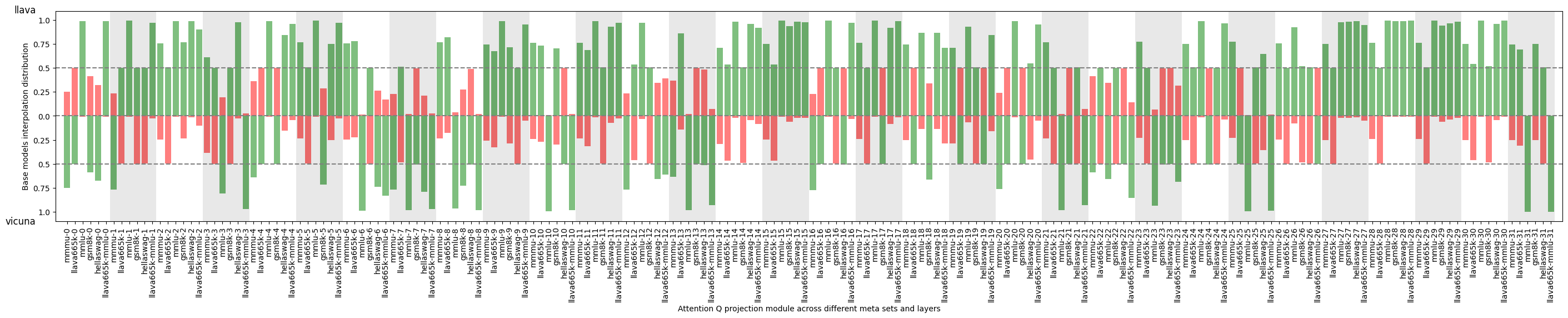}
    \caption{Attention Q mapping.}
  \end{subfigure}
  \hfill

  \begin{subfigure}[b]{0.95\textwidth}
    \includegraphics[width=\textwidth]{figs/reg_alpha_dist/self_attn.k_proj_0.0001L1.png}
    \caption{Attention K mapping.}
  \end{subfigure}
  \hfill

  \begin{subfigure}[b]{0.95\textwidth}
    \includegraphics[width=\textwidth]{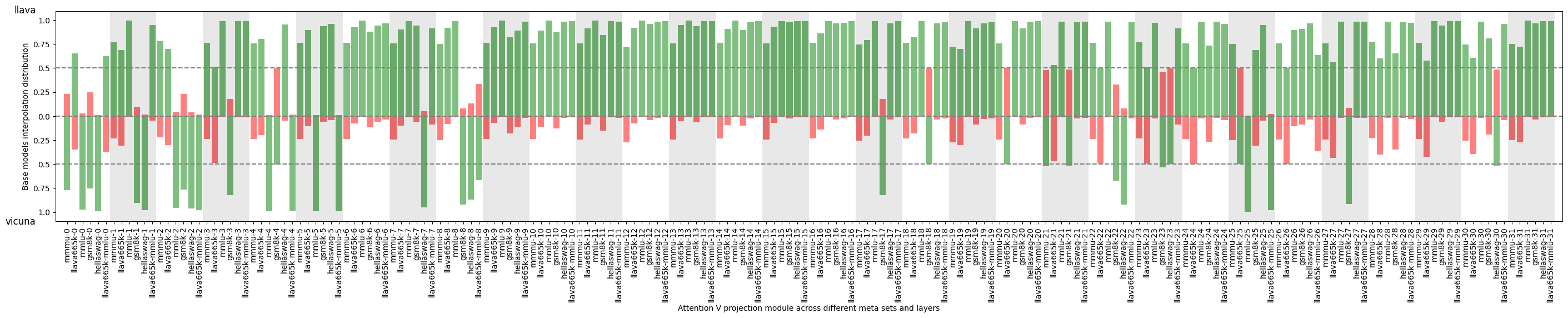}
    \caption{Attnention V mapping.}
  \end{subfigure}
  \hfill

  \begin{subfigure}[b]{0.95\textwidth}
    \includegraphics[width=\textwidth]{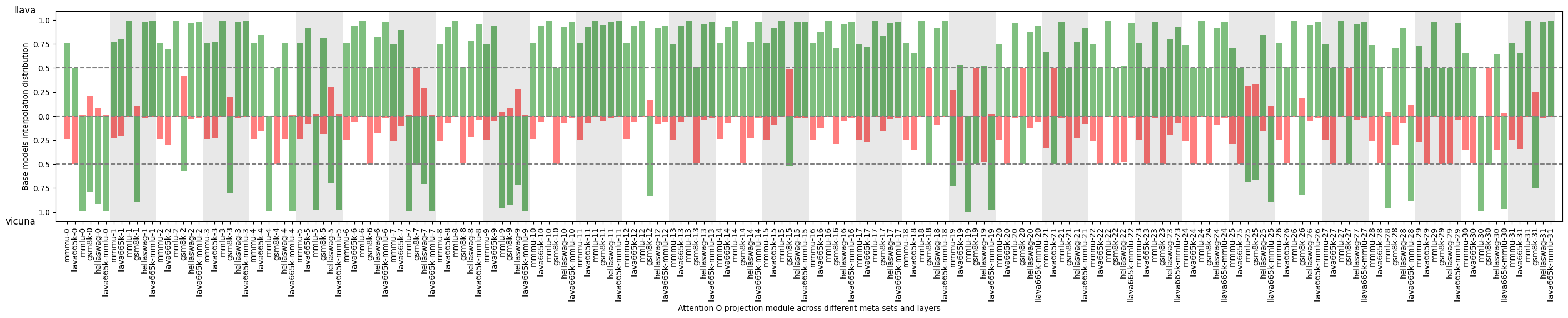}
    \caption{Attention O mapping.}
  \end{subfigure}
  \hfill

  \caption{Regularized (0.0001) $\alpha$ distribution visualizations for attention.}\label{fig:0.0001reg_dist_supp_attn}
\end{figure*}

\begin{figure*}[htbp]
  \centering
  
  \begin{subfigure}[b]{0.95\textwidth}
    \includegraphics[width=\textwidth]{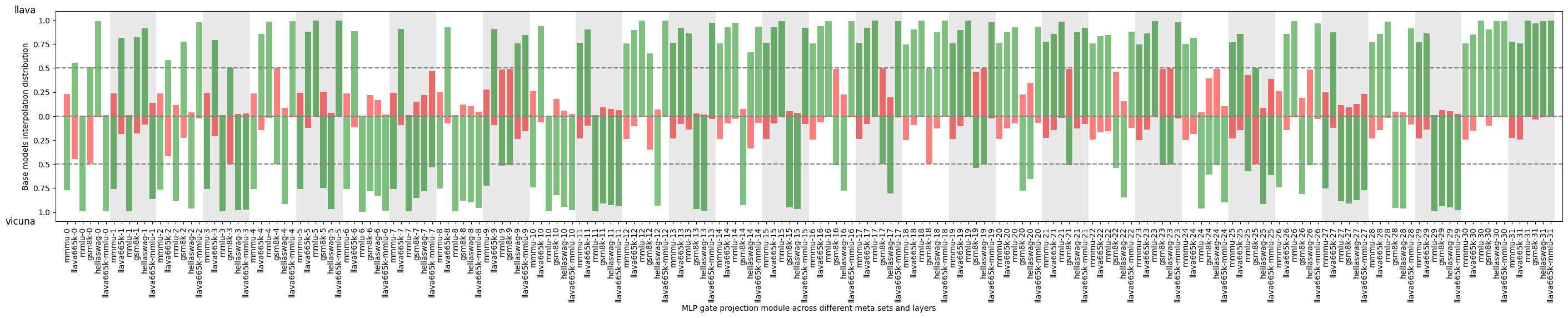}
    \caption{MLP gate mapping.}
  \end{subfigure}
  \hfill
  \begin{subfigure}[b]{0.95\textwidth}
    \includegraphics[width=\textwidth]{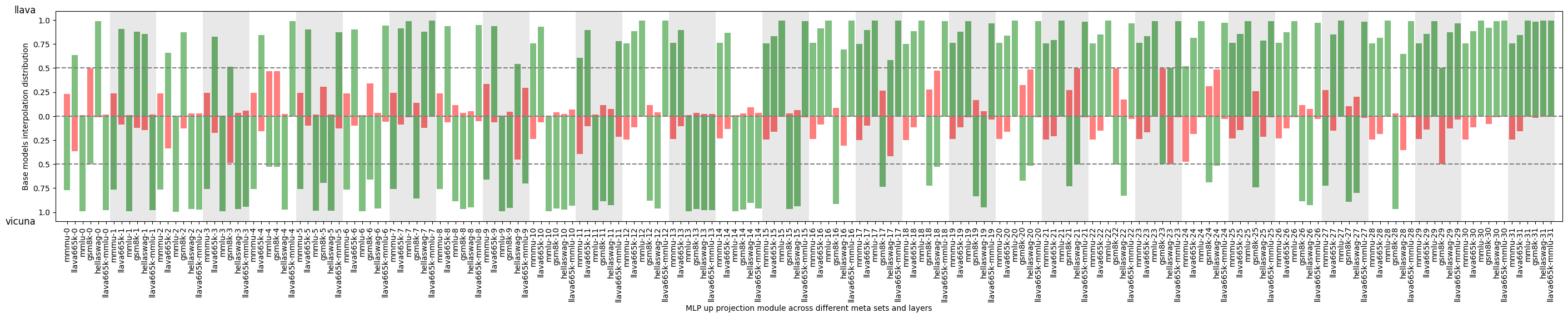}
    \caption{MLP up mapping.}
  \end{subfigure}
  \hfill
  \begin{subfigure}[b]{0.95\textwidth}
    \includegraphics[width=\textwidth]{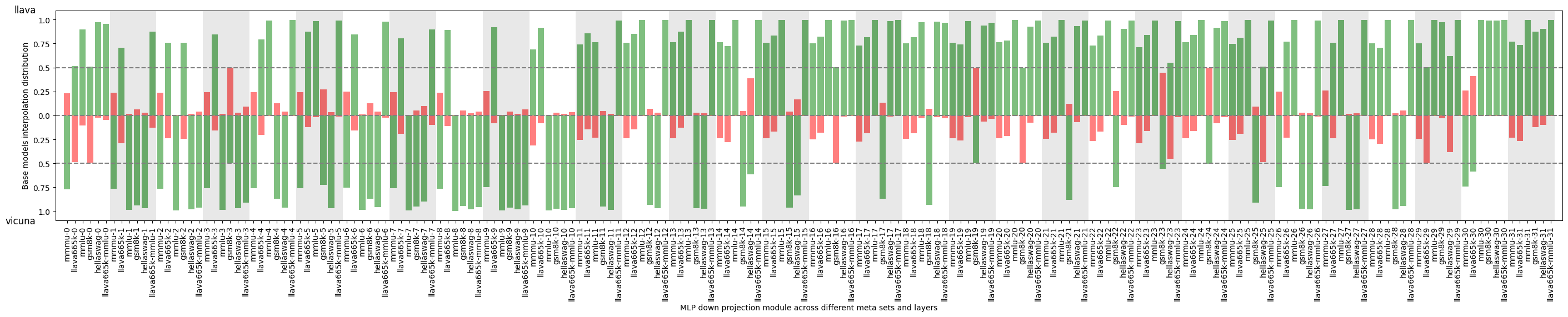}
    \caption{MLP down mapping.}
  \end{subfigure}

  \caption{Regularized (0.0001) $\alpha$ distribution visualizations for MLP.}\label{fig:0.0001reg_dist_supp_mlp}
\end{figure*}

\begin{figure*}[htbp]
  \centering
  \begin{subfigure}[b]{0.95\textwidth}
    \includegraphics[width=\textwidth]{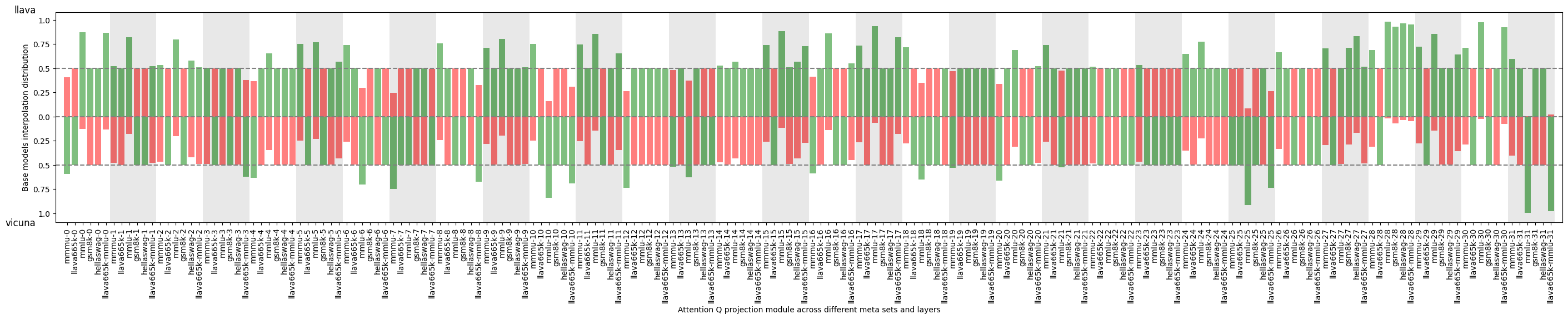}
    \caption{Attention Q mapping.}
  \end{subfigure}
  \hfill

  \begin{subfigure}[b]{0.95\textwidth}
    \includegraphics[width=\textwidth]{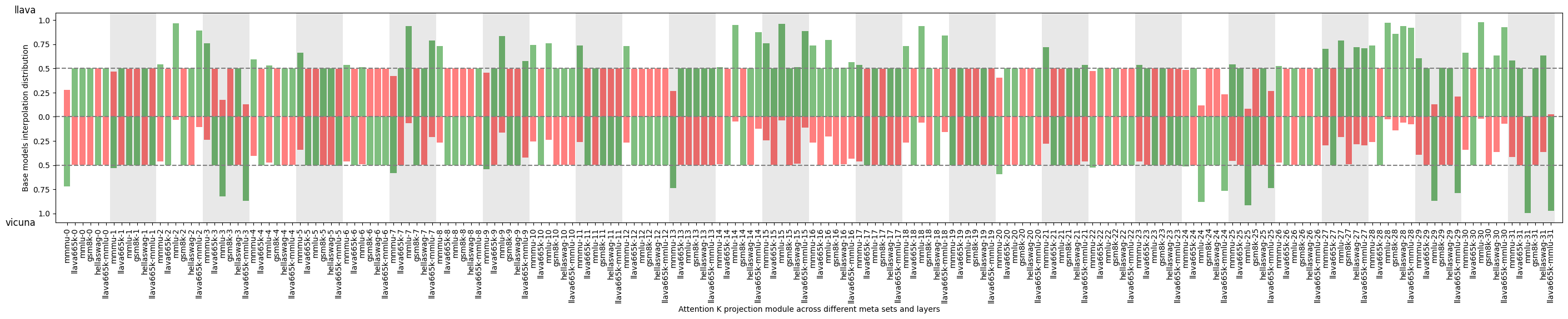}
    \caption{Attention K mapping.}
  \end{subfigure}
  \hfill

  \begin{subfigure}[b]{0.95\textwidth}
    \includegraphics[width=\textwidth]{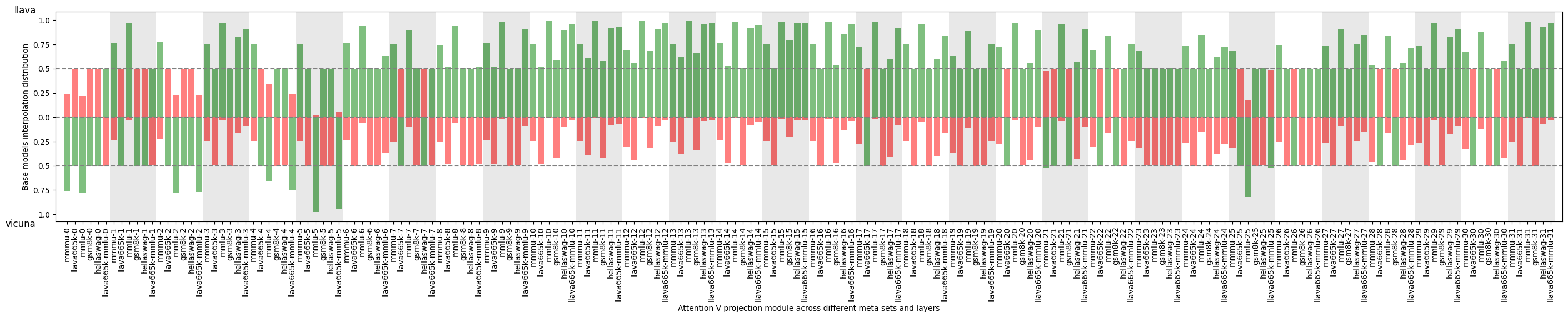}
    \caption{Attnention V mapping.}
  \end{subfigure}
  \hfill

  \begin{subfigure}[b]{0.95\textwidth}
    \includegraphics[width=\textwidth]{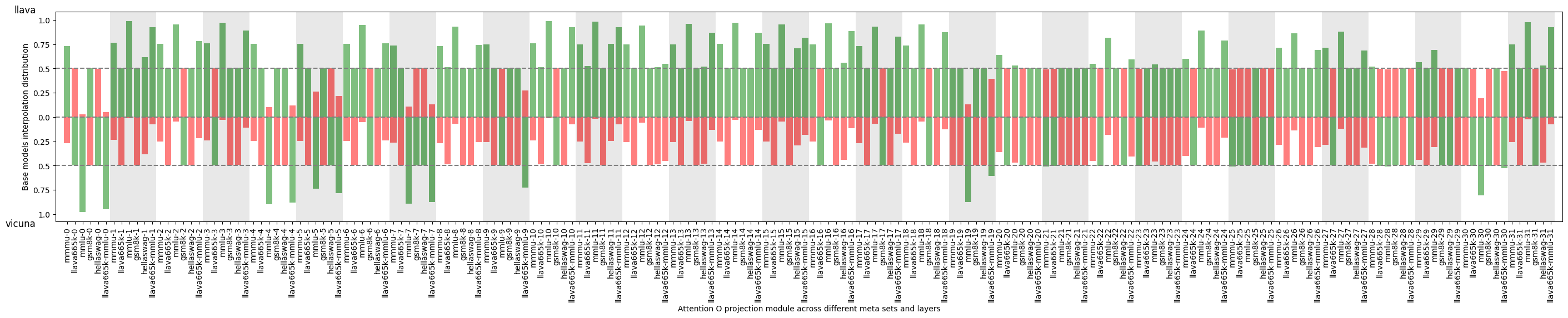}
    \caption{Attention O mapping.}
  \end{subfigure}
  \hfill

  \caption{Regularized (0.001) $\alpha$ distribution visualizations for attention.}\label{fig:0.001reg_dist_supp_attn}
\end{figure*}

\begin{figure*}[htbp]
  \centering
  
  \begin{subfigure}[b]{0.95\textwidth}
    \includegraphics[width=\textwidth]{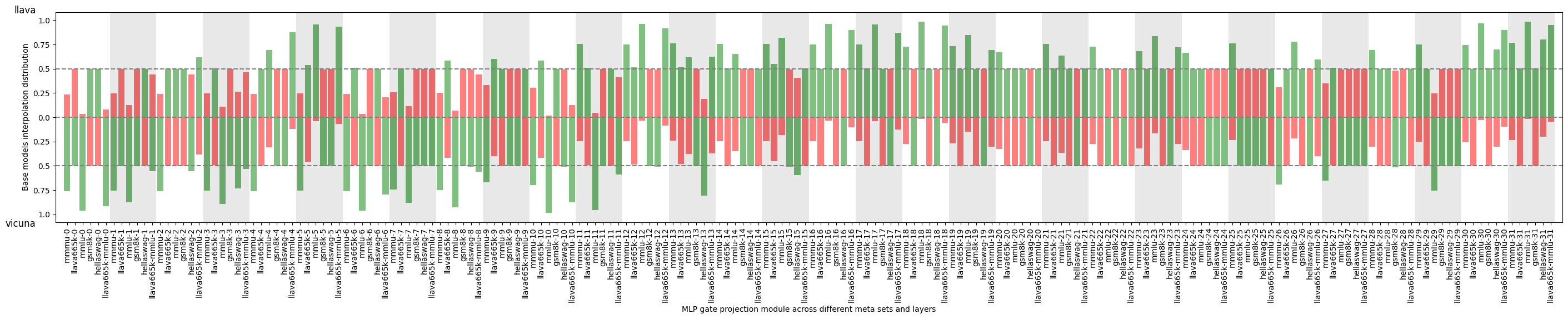}
    \caption{MLP gate mapping.}
  \end{subfigure}
  \hfill
  \begin{subfigure}[b]{0.95\textwidth}
    \includegraphics[width=\textwidth]{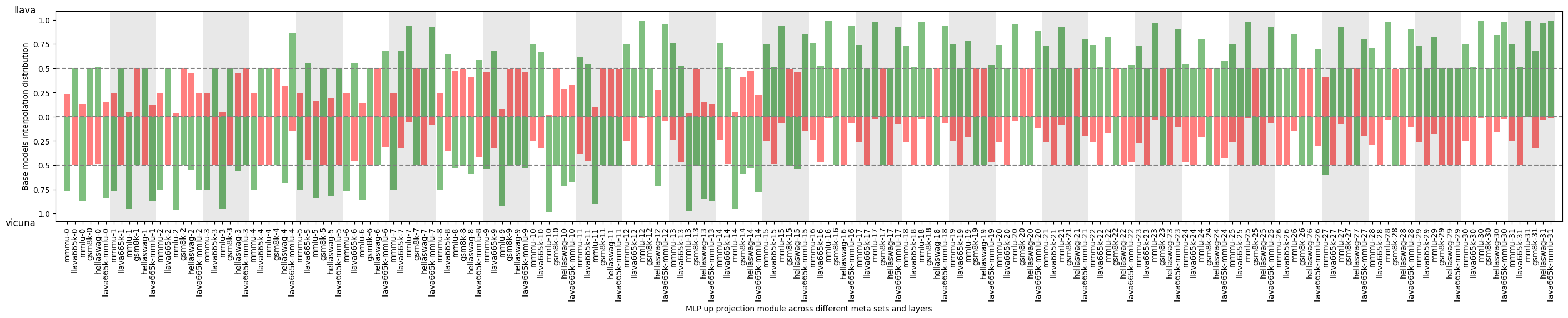}
    \caption{MLP up mapping.}
  \end{subfigure}
  \hfill
  \begin{subfigure}[b]{0.95\textwidth}
    \includegraphics[width=\textwidth]{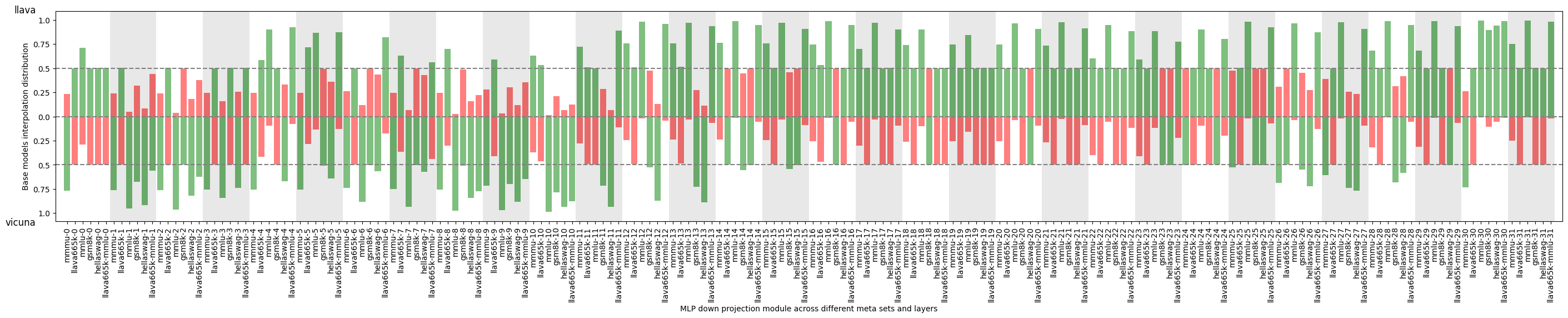}
    \caption{MLP down mapping.}
  \end{subfigure}

  \caption{Regularized (0.001) $\alpha$ distribution visualizations for MLP.}\label{fig:0.001reg_dist_supp_mlp}
\end{figure*}

We provide complete regularized $\alpha$ distribution visualizations in Fig.~\ref{fig:0.0001reg_dist_supp_attn}, Fig.~\ref{fig:0.0001reg_dist_supp_mlp}, Fig.~\ref{fig:0.001reg_dist_supp_attn}, 
Fig.~\ref{fig:0.001reg_dist_supp_mlp}.
They include 0.0001 and 0.001 regularization magnitudes for attention and MLP blocks, which are discussed in Sec.~\ref{sec:soup_behavior}.

\begin{table*}[htbp]
  \centering
    \caption{More evaluations on language and vision-language evaluation benchmarks.}\label{tab:more_eval}
    \resizebox{0.9\textwidth}{!}{
    \begin{tabular}{cccccccc}
      \toprule
      Model & Winogrande & PiQA & MathQA & BoolQA & BBH & POPE & MM-Bench \\
      \midrule
      Vicuna-7B-v1.5 & 69.46 & 77.26 & 27.14 & 80.95 & 42.79 & 80.03 & 1.98 \\
      LLaVA-7B-v1.5 & 70.64 & 77.53 & \textbf{28.11} & 81.71 & 42.14 & 85.86 & \textbf{64.69} \\
      Vanilla-Soup ($\alpha^1 = 0.5$) & 70.71 & \textbf{77.80} & 27.37 & \textbf{82.57} & 43.51 & 86.76 & 62.29 \\
      Meta-Soup & \textbf{70.72} & 77.48 & 27.27 & 82.45 & \textbf{43.96} & \textbf{86.90} & 61.86 \\
      \bottomrule
    \end{tabular}
    }
\end{table*}

\end{document}